\PassOptionsToPackage{numbers,compress,sort}{natbib}
\documentclass{article}
\usepackage[final]{aaj2026}

\usepackage[utf8]{inputenc}
\usepackage[T1]{fontenc}
\usepackage{microtype}
\usepackage{amsmath}
\usepackage{amsfonts}
\usepackage{amssymb}
\usepackage{graphicx}
\usepackage{subcaption}
\usepackage{wrapfig}
\usepackage{booktabs}
\usepackage{array}
\usepackage{mdframed}
\usepackage{nicefrac}
\usepackage{url}
\usepackage{verbatim}
\usepackage{enumitem}
\usepackage[ruled,vlined,linesnumbered]{algorithm2e}
\usepackage{xcolor}
\usepackage{listings}
\usepackage{tcolorbox}
\tcbuselibrary{listings,skins,breakable}
\usepackage{longtable}
\usepackage{pifont}
\definecolor{ggreen}{HTML}{1A7F5A}
\definecolor{gamber}{HTML}{B35C00}
\definecolor{gblue}{HTML}{004A86}
\newcommand{\yes}{\textcolor{ggreen}{\checkmark}}
\newcommand{\no}{\textcolor{gamber}{\ding{55}}}
\usepackage{hyperref}
\usepackage{bibunits}
\defaultbibliographystyle{unsrtnat}
\definecolor{headerblue}{HTML}{1A3A5C}
\definecolor{headerorange}{HTML}{C7440A}

\hypersetup{
  colorlinks = true,
  linkcolor  = headerblue,
  citecolor  = headerblue,
  urlcolor   = headerorange,
}

\lstdefinestyle{moosestyle}{
  basicstyle        = \ttfamily\footnotesize,
  backgroundcolor   = \color{gray!8},
  frame             = single,
  framerule         = 0.4pt,
  rulecolor         = \color{gray!40},
  breaklines        = true,
  keywordstyle      = \color{blue!70!black}\bfseries,
  morekeywords      = {type, grain_num, rand_seed, int_width,
                       threshold, compute_halo_maps,
                       polycrystal_ic_uo},
  tabsize           = 2,
  showstringspaces  = false,
}

\lstdefinestyle{moose_match}{
  basicstyle        = \ttfamily\scriptsize,
  backgroundcolor   = \color{green!6},
  frame             = single,
  framerule         = 0.6pt,
  rulecolor         = \color{green!50!black},
  breaklines        = true,
  tabsize           = 2,
  showstringspaces  = false,
}

\lstdefinestyle{moose_approx}{
  basicstyle        = \ttfamily\scriptsize,
  backgroundcolor   = \color{orange!8},
  frame             = single,
  framerule         = 0.6pt,
  rulecolor         = \color{orange!70!black},
  breaklines        = true,
  tabsize           = 2,
  showstringspaces  = false,
}

\usepackage{lineno}
\usepackage{setspace}
\title{AutoMOOSE: An Agentic AI for Autonomous Phase-Field Simulation}

\author{
Sukriti Manna$^{1,2,*}$ \quad
Henry Chan$^{2}$ \quad
Subramanian K.\,R.\,S. Sankaranarayanan$^{1,2,*}$ \\[0.5em]
$^{1}$Department of Mechanical and Industrial Engineering, University of Illinois Chicago, Chicago, IL 60607, USA \\
$^{2}$Center for Nanoscale Materials, Argonne National Laboratory, Lemont, IL 60439, USA \\[0.5em]
$^{*}$Correspondence: \texttt{smanna@anl.gov}, \texttt{skrssank@anl.gov}
}

\begin{document}

% ===== Unit 1: main manuscript + its own bibliography =====
% The \bibliography call inside sections/05_conclusion.tex is redefined
% to \putbib so this unit's reference list appears at that spot.
\begin{bibunit}[unsrtnat]
\renewcommand{\bibliography}[1]{\putbib[automoose]}
\renewcommand{\bibliographystyle}[1]{}

\maketitle

\begin{abstract}
Phase-field modeling links thermodynamics and kinetics to microstructural
evolution, but using multiphysics frameworks such as MOOSE requires substantial
expertise to construct valid inputs, manage simulation campaigns, diagnose
failures, and validate results. We introduce \textbf{AutoMOOSE}, an open-source
multi-agent framework that orchestrates the complete simulation lifecycle from
a single natural-language prompt. Six specialized agents --- Architect, Input
Writer, Runner, Reviewer, Visualization agent, and a physics-grounded
Skeptic --- generate, execute, analyze, and adversarially test simulations
against conservation laws, asymptotic limits, and scaling relations. We
validate AutoMOOSE in two distinct phase-field domains: non-conserved copper
grain growth governed by Allen--Cahn dynamics and conserved Fe--Cr spinodal
decomposition governed by Cahn--Hilliard dynamics. On a prospectively
specified 25-task grain-growth benchmark spanning temperature, grain count,
resolution, and model formulation, AutoMOOSE generates syntactically valid
inputs for all 25 tasks, produces completed simulations with measurable
coarsening in 19, and delivers 15 results that satisfy Burke--Turnbull
kinetics ($R^2 \geq 0.90$) and survive Skeptic falsification. The unsuccessful
cases are traced to two identifiable generator defects. An ensemble of 1000
simulations recovers the prescribed activation energy to within $1\%$ after
finite-size extrapolation ($Q_\infty = 0.228$\,eV versus $0.230$\,eV). In the
conserved domain, an agent-generated, CALPHAD-based Fe--Cr simulation limits
relative mass drift to $6.3\times10^{-6}$, dissipates free energy to a stable
plateau, and evolves toward the equilibrium tie-line. Controlled ablations
show that the full pipeline converts stochastic raw generation --- 3--7
successful tasks out of 8 across repeated trials --- into consistent
success of 8 out of 8. AutoMOOSE therefore provides a practical route from
natural-language specification to reproducible, physics-validated phase-field
simulation campaigns.
\end{abstract}


%%%============================================================
%%%  SECTION 1 — INTRODUCTION
%%%============================================================

\section{Introduction}
Microstructure governs the properties of structural and functional
materials: grain size controls yield strength and creep resistance
in metallic alloys, domain topology determines the switching
efficiency of ferroelectrics, and precipitate morphology sets the
coarsening kinetics of
superalloys~\cite{steinbach2009phase,olson1997computational,manna2023understanding}.
Because microstructure emerges from processing history, predicting
its evolution is central to materials discovery and design.
Phase-field methods provide precisely this predictive capability
--- a thermodynamically consistent framework for modeling grain
growth, spinodal decomposition, solidification, and ferroelectric
domain dynamics --- and have become a widely used approach to
mesoscale microstructure
simulation~\cite{chen2002phase,moelans2008quantitative}.
Among the platforms that have made these methods accessible at
scale, the Multiphysics Object-Oriented Simulation Environment
(MOOSE) provides a modular finite-element architecture,
scalability from workstations to leadership-class
high-performance computing (HPC) systems, and a broad, active
community of materials scientists and
engineers~\cite{gaston2009moose,tonks2012object}.

Yet the power of these frameworks comes at a steep practical
price: translating a scientific objective into an executable
phase-field model demands simultaneous expertise in
thermodynamics, kinetic equations, numerical discretization,
nonlinear solvers, and platform-specific input
syntax~\cite{wilson2017,gaston2009moose}.
Even a standard grain-growth simulation in MOOSE requires tightly
coupled specifications for meshing, order-parameter
initialization, free-energy and evolution kernels, boundary
conditions, solver controls, and adaptive time integration, often
exceeding one hundred lines of interdependent
syntax~\cite{moose_examples,moelans2008quantitative}.
An error in any one of these components can produce an
input-parsing failure, solver divergence, or --- most
insidiously --- a numerically completed simulation that violates
the intended physics.
This last failure mode connects the expertise barrier to a deeper
reproducibility problem: differences in spatial resolution, solver
tolerances, initial conditions, or undocumented parameter choices
can alter quantitative results without any visible sign of
error~\cite{milkowski2018replicability,wilkinson2016}, and the
absence of structured workflows impedes the systematic generation
of simulation datasets needed to build
processing--microstructure--property
databases~\cite{olson1997computational}.

These barriers are not unique to the mesoscale, and agentic AI
systems have recently begun to dismantle their counterparts at
other scales of materials modeling.
At the atomistic scale, large language model (LLM)-driven agents
have been applied to interatomic-potential development, automated
density-functional-theory workflows, and autonomous molecular
design~\cite{wang2025dreams,yang2026quasar,vriza2026multi};
at the continuum scale, emerging systems have explored
agent-assisted finite-element meshing and topology
optimization~\cite{qi2025feagpt,deotale2026all}.
Underpinning both is a broader body of work showing that
tool-using agents can decompose natural-language objectives,
generate executable artifacts, inspect intermediate outputs, and
revise their actions using external
feedback~\cite{yao2022react,shinn2023reflexion,schick2023toolformer,bran2023chemcrow}
--- capabilities that map directly onto the workflow friction
described above.
Mesoscale phase-field modeling, however --- despite sitting at the
intersection of thermodynamic theory, numerical methods, and
processing-scale phenomena --- has received comparatively little
attention from such systems.
Nor would input generation alone close the gap: a plausible input
file does not establish that a simulation implements the requested
physics, completes successfully, or produces physically admissible
results.
The unresolved challenge is therefore not whether an LLM can draft
a phase-field input, but whether agentic scaffolding can convert
stochastic model output into reproducible simulations whose
physical validity is quantitatively tested.

Here we introduce \textbf{AutoMOOSE}, an open-source agentic
framework built to meet exactly this challenge: it orchestrates
the full MOOSE simulation lifecycle --- from natural-language
prompt to physics-tested result --- through a six-agent pipeline
of Architect, Input Writer, Runner, Reviewer, Visualization agent,
and a physics-grounded Skeptic, running on a model-agnostic
language-model backend (Section~\ref{sec:pf}).
Along this pipeline, the agents parse user intent, generate
validated MOOSE input files, execute parallel parameter sweeps,
and screen runs for successful execution; the Skeptic then ---
critically --- adversarially tests each completed result against
conservation laws, asymptotic limits, and scaling relations,
classifying it as credible only if it survives falsification.
This scientific core is wrapped in an extensible engineering
shell: a modular plugin architecture separates physics-specific
logic from the orchestration layer, so new phase-field
formulations can be added without modifying shared
infrastructure, while a Model Context Protocol (MCP)
server~\cite{anthropic2024claude} exposes the complete workflow as
ten structured tools for integration with MCP-compatible clients
and composability with external optimization pipelines.
Finally, every run produces a self-documenting directory recording
inputs, outputs, execution metadata, and validation decisions ---
structured provenance that supports reproducibility and alignment
with FAIR (Findable, Accessible, Interoperable, Reusable) data
principles~\cite{wilkinson2016}.

To test whether this architecture delivers on the challenge, we
validate AutoMOOSE in two phase-field domains with complementary
physics: non-conserved copper grain growth governed by
Allen--Cahn dynamics and conserved Fe--Cr spinodal decomposition
governed by Cahn--Hilliard dynamics.
In the first domain, on a prospectively specified 25-task
grain-growth benchmark with machine-checkable success gates,
AutoMOOSE generates syntactically valid inputs for all 25 tasks,
produces completed simulations with measurable coarsening in 19,
and delivers 15 fully credible results --- those satisfying all
pre-specified physical-validity criteria and surviving Skeptic
falsification (Section~\ref{sec:results}) --- with the
unsuccessful cases traced to two identifiable generator defects.
Scaling beyond the benchmark, an ensemble campaign of 1000
simulations recovers the prescribed activation energy to within
$1\%$ after finite-size extrapolation.
In the second domain, an agent-generated Cahn--Hilliard simulation
exhibits a relative mass drift of $6.3\times10^{-6}$, monotonic
free-energy dissipation, and evolution toward the equilibrium
tie-line.
Finally, a controlled ablation isolates the contribution of the
scaffolding itself: raw single-model generation succeeds on only
3--7 of 8 tasks across repeated trials, whereas the full pipeline
succeeds on all 8 in every trial --- direct evidence that the
agentic layer, not the underlying model alone, is what renders the
workflow reproducible.

The remainder of this paper is organized as follows.
Section~\ref{sec:pf} describes the AutoMOOSE software design,
including the phase-field models, the six-agent pipeline, the
plugin architecture, and the benchmark specification.
Section~\ref{sec:results} presents validation results across both
physics domains, the ensemble campaign, the ablation study, and
Skeptic falsification as methodology.
Sections~\ref{sec:discussion} and~\ref{sec:conclusion} discuss
limitations, future directions, and an outlook on agentic AI for
computational materials science.

%%%============================================================
%%%  SECTION 2 — SOFTWARE DESIGN
%%%============================================================

\section{Computational Methods and Software Design}
\label{sec:methods}

%─────────────────────────────────────────────
\subsection{Phase-Field Model for Grain Growth}
\label{sec:pf}
%─────────────────────────────────────────────

Phase-field models represent discrete crystallographic
orientations as continuous order parameters, so that grain
boundary migration emerges from free-energy minimization rather
than explicit interface tracking --- circumventing the topological
challenges of sharp-interface methods while capturing
curvature-driven coarsening, Zener pinning, solute
drag~\cite{manna2023understanding,chakrabarti2018zener}, and the
many-grain topology changes of microstructure evolution.
Among these formulations, AutoMOOSE is demonstrated on the
multiphase Allen--Cahn model of Moelans
\textit{et al.}~\cite{moelans2008quantitative}, which serves as
the representative physics throughout
Sections~\ref{sec:overview}--\ref{sec:viz}.
This first demonstration targets non-conserved dynamics, in which
the order parameters evolve to minimize free energy without a
conservation constraint; the extension to conserved
Cahn--Hilliard dynamics --- Fe--Cr spinodal decomposition, where
composition is a conserved field --- follows in
Section~\ref{sec:spinodal}.
We establish the model here because the agent pipeline translates
each physical quantity introduced below directly into a MOOSE
input construct, and the closing consistency check of this
section becomes the pipeline's end-to-end validation target.

In the Moelans formulation, each grain $i$ is represented by an
order parameter $\eta_i$ ($i = 1, \ldots, N_{\mathrm{op}}$) equal
to unity inside the grain and zero elsewhere, with boundaries
identified by the overlap of neighboring order parameters.
Each $\eta_i$ evolves by the Allen--Cahn equation,
\begin{equation}
    \frac{\partial \eta_i}{\partial t}
    \;=\; -L(T)\,\frac{\delta F}{\delta \eta_i},
    \label{eq:allen_cahn}
\end{equation}
where $L(T)$ is the temperature-dependent kinetic coefficient and
$F$ the total free energy,
\begin{equation}
    F \;=\; \int_V \!\left[\, f_{\mathrm{loc}}(\boldsymbol{\eta})
    \;+\; \frac{\kappa}{2} \sum_{i=1}^{N_{\mathrm{op}}}
    |\nabla\eta_i|^2 \right] dV,
    \label{eq:free_energy}
\end{equation}
with the local free energy density in the Moelans
form~\cite{moelans2008quantitative},
\begin{equation}
    f_{\mathrm{loc}} \;=\; \mu \left(
    \sum_{i} \left(\frac{\eta_i^4}{4} - \frac{\eta_i^2}{2}\right)
    + \gamma \sum_{i}\sum_{j>i} \eta_i^2\,\eta_j^2
    + \frac{1}{4} \right),
    \label{eq:floc}
\end{equation}
where $\mu$ is the free energy weight and $\gamma = 1.5$ enforces
symmetric interfacial profiles.
Together, these terms encode the physics of coarsening: the
double wells stabilize bulk grain regions
($\eta_i \in \{0,1\}$), the cross-coupling term penalizes overlap
between neighboring order parameters, and the ratio of gradient to
local energy sets the diffuse interface width $w_{\mathrm{GB}}$,
which the computational mesh must resolve
($h \leq w_{\mathrm{GB}}/4$).

The model coefficients $\{L, \mu, \kappa\}$ appearing in these
equations are not directly measurable; assuming isotropic grain
boundary properties, they map onto three quantities that are ---
grain boundary energy $\sigma$, interface width $w_{\mathrm{GB}}$,
and mobility $M_{\mathrm{GB}}$ ---
through~\cite{moelans2008quantitative}
\begin{equation}
    L \;=\; \frac{4\,M_{\mathrm{GB}}}{3\,w_{\mathrm{GB}}}, \qquad
    \mu \;=\; \frac{6\,\sigma}{w_{\mathrm{GB}}}, \qquad
    \kappa \;=\; \frac{3}{4}\,\sigma\,w_{\mathrm{GB}}.
    \label{eq:model_params}
\end{equation}
This mapping is central to AutoMOOSE: the user specifies
$\{\sigma, w_{\mathrm{GB}}, M_{\mathrm{GB}}\}$, and the Input
Writer agent ($f_2$) computes $\{L, \mu, \kappa\}$ analytically
via Eq.~\eqref{eq:model_params}, eliminating a common source of
error in hand-authored inputs (Section~\ref{sec:inputwriter}).
Temperature enters the model through the mobility, which carries
an Arrhenius dependence,
\begin{equation}
    M_{\mathrm{GB}}(T) \;=\; M_0\,
    \exp\!\left(-\frac{Q}{k_{\mathrm{B}}\,T}\right),
    \label{eq:mob_arrhenius}
\end{equation}
with pre-exponential mobility $M_0$, activation energy $Q$, and
Boltzmann constant $k_{\mathrm{B}}$; through
Eq.~\eqref{eq:model_params}, $L(T)$ inherits the same dependence.

This Arrhenius input has a directly observable consequence in the
coarsening kinetics, quantified through the grain count $N(t)$
tracked by the MOOSE \texttt{GrainTracker} algorithm.
In two dimensions, the grain coarsening law~\cite{burke1952} gives
\begin{equation}
    \frac{1}{N(t)} - \frac{1}{N_0} \;=\; \tilde{k}(T)\,t,
    \label{eq:Nt}
\end{equation}
where $N_0 = N(0)$ and the rate constant $\tilde{k}(T)$ is
extracted from the slope of $N^{-1}(t)$ against $t$.
Because $\tilde{k}(T) \propto M_{\mathrm{GB}}(T)$, it follows the
same Arrhenius form,
\begin{equation}
    \tilde{k}(T) \;=\; \tilde{k}_0\,
    \exp\!\left(-\frac{Q}{k_{\mathrm{B}}\,T}\right),
    \label{eq:arrhenius}
\end{equation}
so the activation energy governing the microscopic mobility in
Eq.~\eqref{eq:mob_arrhenius} reappears as the slope of a
macroscopic Arrhenius plot.
Since $Q$ is a user-specified input, the value recovered by
fitting Eq.~\eqref{eq:arrhenius} to a simulated temperature sweep
must match it --- the closed-loop consistency check anticipated
above, exercised in Section~\ref{sec:arrhenius}.

\begin{figure}
  \centering
  \includegraphics[width=\linewidth]{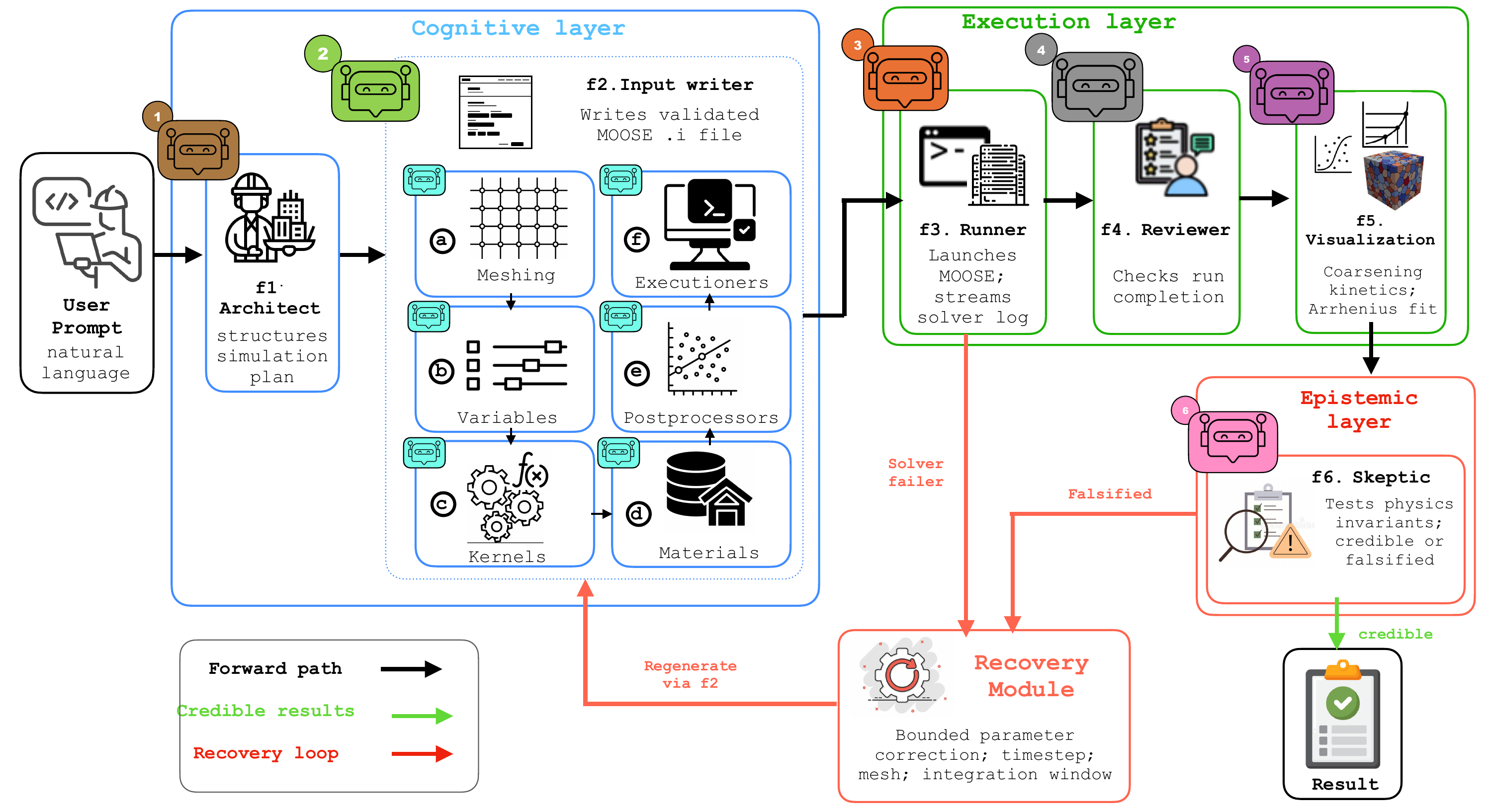}
  \caption{%
    \textbf{AutoMOOSE agentic pipeline.}
    Six model-agnostic language-model agents ($f_1$--$f_6$),
    organized into a cognitive, an execution, and an epistemic
    layer, transform a natural-language prompt into a completed
    and physics-tested MOOSE phase-field simulation.
    In the \emph{cognitive layer}, the \textbf{Architect}~($f_1$)
    parses the user prompt into the structured simulation plan
    $\mathcal{P}$ (Eq.~\eqref{eq:simplan}), encoding the physics
    model, mesh geometry, solver tolerances, and sweep parameters;
    the \textbf{Input Writer}~($f_2$), a compound agent, then
    coordinates six sub-agents in strict dependency order ---
    (\textit{a})~Meshing, (\textit{b})~Variables,
    (\textit{c})~Kernels, (\textit{d})~Materials,
    (\textit{e})~Postprocessors, and (\textit{f})~Executioner ---
    to render a validated MOOSE \texttt{.i} input file via the
    plugin registry (Section~\ref{sec:plugin}).
    In the \emph{execution layer}, the \textbf{Runner}~($f_3$)
    launches MOOSE and streams the solver log; the
    \textbf{Reviewer}~($f_4$) screens each run for completion,
    distinguishing genuine solver breakdown from transient
    adaptive-timestep events (Section~\ref{sec:reviewer}); and the
    \textbf{Visualization} agent~($f_5$) extracts the coarsening
    kinetics and performs the Arrhenius analysis from CSV
    postprocessor output (Section~\ref{sec:viz}).
    In the \emph{epistemic layer}, the \textbf{Skeptic}~($f_6$)
    adversarially tests each completed run against falsifiable
    physics invariants and classifies it as credible or falsified
    (Section~\ref{sec:skeptic}); only credible runs enter
    campaign-level analysis.
    Recovery is triggered on two paths: a run that fails to
    complete is diagnosed from its solver-log signatures, and a
    completed run falsified for a correctable cause is diagnosed
    from the Skeptic's verdict; either routes to the recovery
    module, which applies a bounded, logged correction (timestep,
    mesh, or integration window; at most three attempts) and
    re-enters the pipeline at the Input Writer
    (Section~\ref{sec:failure_recovery}).
    Black arrows: forward path; green arrows: credible results;
    red arrows: recovery loop.
  }
  \label{fig:pipeline}
\end{figure}

\subsection{Overview of Agentic Workflow}
\label{sec:overview}
%─────────────────────────────────────────────

Building on the phase-field model established in
Section~\ref{sec:pf}, AutoMOOSE realizes the simulation workflow
through a modular, agent-based orchestration framework that
operates entirely at the workflow layer --- interpreting user
intent, constructing MOOSE inputs that encode
Eqs.~\eqref{eq:allen_cahn}--\eqref{eq:model_params}, managing
execution, monitoring for failures, and extracting quantitative
results --- without requiring the user to write a single line of
MOOSE syntax.
Six specialized large language model agents implement this
pipeline (Fig.~\ref{fig:pipeline}), each carrying a physics-aware
system prompt reproduced in full in Supplementary
Section~\ref{sec:SI_prompts}.
The agents themselves are model-agnostic: they are routed through
a provider-independent client, and the results reported here use
Claude Opus~4.8 and Claude Sonnet~4.6, with a self-hosted
open-weights model also verified end-to-end
(Table~\ref{tab:backends})~\cite{anthropic2024claude}.

\begin{table}[h]
\centering\small
\caption{AutoMOOSE runs on interchangeable backends. The same
agent pipeline drives a grain-growth task end-to-end under both a
closed-weights frontier model and a self-hosted open-weights
model; switching backends requires only a configuration change,
so the framework does not depend on any single provider's
availability.}
\label{tab:backends}
\begin{tabular}{llll}
\toprule
Backend & Weights & Hosting & Pipeline (end-to-end)\\
\midrule
Claude Sonnet~4.6 & Closed & Provider API & completes\\
Qwen2.5-32B-Instruct & Open & Self-hosted (Perlmutter, vLLM,
4$\times$A100) & completes\\
\bottomrule
\end{tabular}
\end{table}

What the agents exchange, and in what order, admits a compact
formal description: the pipeline is a sequential composition of
agent functions,
\begin{equation}
    \mathcal{S} \;=\; f_6 \circ f_5 \circ f_4 \circ f_3 \circ f_2
    \circ f_1(\mathcal{U}),
    \label{eq:pipeline}
\end{equation}
where $\mathcal{U}$ denotes the natural-language user intent and
each agent $f_i : \mathcal{P}_{i-1} \to \mathcal{P}_i$
progressively enriches a shared simulation plan,
\begin{equation}
    \mathcal{P} \;=\; \bigl(\,\Omega,\; h,\; \mathcal{M},\;
    \mathcal{B},\; \boldsymbol{\theta}_{\mathrm{solver}},\;
    \boldsymbol{\theta}_{\mathrm{run}},\; \mathcal{O}\,\bigr),
    \label{eq:simplan}
\end{equation}
encoding the domain $\Omega$, mesh spacing $h$, physical model
$\mathcal{M}$ (which stores the measurable set $\{\sigma,
w_{\mathrm{GB}}, M_{\mathrm{GB}}\}$ of Section~\ref{sec:pf}),
boundary conditions $\mathcal{B}$, solver tolerances
$\boldsymbol{\theta}_{\mathrm{solver}}$, runtime parameters
$\boldsymbol{\theta}_{\mathrm{run}}$, and requested outputs
$\mathcal{O}$.
The plan is thus the single shared state object of the pipeline:
each $f_i$ reads the fields it requires, writes the fields it
produces, and hands the enriched plan to the next stage, so no
inter-agent messaging is needed beyond this structured handoff
(managed by the FastAPI backend~\cite{fastapi2024}).

Reading Eq.~\eqref{eq:pipeline} left to right, the six agents
group naturally into three functional layers, indicated by the
colored frames in Fig.~\ref{fig:pipeline}.
The \emph{cognitive layer} (Architect~$f_1$, Input Writer~$f_2$)
interprets user intent and constructs a validated MOOSE input
file encoding Eqs.~\eqref{eq:allen_cahn}--\eqref{eq:model_params}.
The \emph{execution layer} (Runner~$f_3$, Reviewer~$f_4$,
Visualization~$f_5$) manages simulation execution, screens each
run for successful completion, and extracts the kinetic
observables $N(t)$ and $\tilde{k}(T)$ of
Eqs.~\eqref{eq:Nt}--\eqref{eq:arrhenius}.
The \emph{epistemic layer} (Skeptic~$f_6$) adversarially tests
each completed run against the physical laws it must satisfy,
classifying it as credible only if it survives falsification
(Section~\ref{sec:skeptic}).
This separation of concerns serves the pipeline at both ends:
input-construction errors are caught in the cognitive layer
before they propagate into costly simulation runs, and the
credibility verdict is kept independent of the machinery that
produced the result.

These layers come together when a task traverses the pipeline
(Fig.~\ref{fig:pipeline}).
The Architect ($f_1$) parses the user prompt into the simulation
plan $\mathcal{P}$, from which the Input Writer ($f_2$) --- a
compound agent coordinating six sub-agents in strict dependency
order --- renders a validated MOOSE \texttt{.i} input file.
The Runner ($f_3$) launches MOOSE, streams the solver log, and
records whether the run reaches its target integration time; the
Reviewer ($f_4$), a completion-aware execution screen, then
determines whether the run executed successfully, distinguishing
genuine solver breakdown from transient adaptive-timestep events
that recover and complete.
For every completed run, the Visualization agent ($f_5$) extracts
the coarsening kinetics and performs the Arrhenius analysis, and
the Skeptic ($f_6$) independently subjects the result to
falsification against its physical invariants.
Only runs that both complete execution and survive falsification
enter campaign-level analysis; this verification arc operates
without user intervention and is detailed in
Sections~\ref{sec:reviewer} and~\ref{sec:skeptic}.
With the workflow now in view end to end, the following
subsections examine each agent in turn
(Sections~\ref{sec:architect}--\ref{sec:viz}), followed by the
plugin architecture that makes the framework physics-agnostic
(Section~\ref{sec:plugin}) and the Model Context Protocol
interface that enables headless operation
(Section~\ref{sec:mcp}).

%─────────────────────────────────────────────
\subsection{Architect Agent}
\label{sec:architect}
%─────────────────────────────────────────────

The Architect agent $f_1$ is the entry point of the pipeline,
receiving the raw natural-language user intent $\mathcal{U}$ and
producing a fully resolved simulation plan $\mathcal{P}$
(Eq.~\eqref{eq:simplan}) as a structured JSON object.
It is called once per user request and its output is passed
directly to the Input Writer $f_2$.

$f_1$ is responsible for five parsing tasks: identifying the
physics formulation (\texttt{GBEvolution} or
\texttt{LinearizedInterface}), the spatial dimension
(2D QUAD4 or 3D HEX8), the boundary conditions (periodic or
Dirichlet), the sweep intent (detecting whether the user specifies
multiple values for any sweepable parameter such as $T$,
\texttt{grain\_num}, or $Q$, and extracting the corresponding
sweep range $\Theta$), and the solver strategy (adaptive
timestepping and AMR on or off).
These five items map directly onto the components of $\mathcal{P}$:
task~1 populates $\mathcal{M}$; tasks~2--3 populate $\Omega$ and
$\mathcal{B}$; task~4 populates $\boldsymbol{\theta}_\mathrm{r}$
with $\Theta$; task~5 populates $\boldsymbol{\theta}_\mathrm{s}$.

A strict JSON-only output constraint is enforced in the system
prompt: the FastAPI backend parses the response with
\texttt{json.loads()} and raises a structured error on any prose,
preventing silent propagation of ambiguous plans to $f_2$.
When the requested physics is not supported by any registered
plugin, $f_1$ returns
\texttt{\{"formulation":"stub","error":"unsupported\_physics"\}},
which the backend surfaces to the user with a list of available
plugins rather than attempting input generation.
The complete system prompt for $f_1$ is reproduced verbatim in
Supplementary Section~S1.1.
% ============================================================
% §2.4  Input Writer Agent
% ============================================================

%%% ========================================================
\subsection{Input Writer Agent}
\label{sec:inputwriter}

The Input Writer agent $f_2$ receives the simulation plan
$\mathcal{P}$ from the Architect and produces a complete,
syntax-validated MOOSE \texttt{.i} input file.
Rather than delegating this task to a single monolithic prompt,
$f_2$ is implemented as a compound agent (Fig.~\ref{fig:pipeline})
that coordinates six sequential sub-agents, each responsible for
one logical block group of the input file.
This decomposition mirrors the physical dependency structure of
MOOSE~\cite{gaston2009moose}, which forms a directed acyclic graph (DAG):
each sub-agent executes only after
all blocks it depends on are complete, following the topological
order
\begin{equation}
\begin{aligned}
  \texttt{[Mesh]} &\prec \texttt{[GlobalParams]} \prec
  \texttt{[Variables]} \prec \texttt{[Kernels]} \\
  &\prec \texttt{[Materials]} \prec \texttt{[Postprocessors]}
  \prec \texttt{[Executioner]},
\end{aligned}
\label{eq:dag_order}
\end{equation}
where $\prec$ denotes topological precedence.

Three dependencies in this ordering are worth making explicit.
First, \texttt{op\_num} declared in \texttt{[GlobalParams]}
propagates into every downstream block referencing $\eta_i$, so
it must be fixed before kernels, materials, or postprocessors are
written.
Second, \texttt{GrainTracker} in \texttt{[UserObjects]} requires
both variable declarations and material definitions, placing it
last among the physics blocks.
Third, the solver tolerances $\epsilon_{\mathrm{nl}}$ and
$\epsilon_{\mathrm{l}}$ written into \texttt{[Executioner]} are
drawn from $\boldsymbol{\theta}_{\mathrm{solver}} \subset
\mathcal{P}$ --- the same values the Reviewer agent uses to
classify convergence failures at runtime
(Section~\ref{sec:reviewer}) --- ensuring consistency between the
input file and the error-diagnosis logic.
Table~\ref{tab:subagents} summarises each sub-agent, its owned
blocks, key parameters, and upstream dependencies.

\begin{table}[h]
\centering
\caption{%
  \textbf{Input Writer sub-agents and MOOSE block responsibilities.}
  The six sub-agents of $f_2$ generate the MOOSE \texttt{.i} input
  file in the topological order of Eq.~\eqref{eq:dag_order}.
  The \textit{Depends on} column lists upstream fields that must be
  available before each sub-agent executes.
}
\label{tab:subagents}
\small
\renewcommand{\arraystretch}{1.3}
\begin{tabular}{p{2.4cm} p{3.8cm} p{3.8cm} p{3.0cm}}
\toprule
\textbf{Sub-agent}
  & \textbf{MOOSE blocks}
  & \textbf{Key parameters set}
  & \textbf{Depends on} \\
\midrule
\textit{(a)}~Mesh
  & \texttt{[Mesh]}
  & $n_x$/$n_y$/$n_z$, $L_x$/$L_y$/$L_z$,
    \texttt{elem\_type} (QUAD4 or HEX8),
    \texttt{parallel\_type};
    enforces $h \leq w_{\mathrm{GB}}/4$
  & $\Omega$, $h$, \texttt{dim}
    from $\mathcal{P}$ \\
\textit{(b)}~Variables
  & \texttt{[GlobalParams]},
    \texttt{[Variables]},
    \texttt{[ICs]},
    \texttt{[AuxVariables]}
  & \texttt{op\_num},
    $\eta_i$ fields,
    Voronoi or Random IC;
    \texttt{bound\_value}
    (LinearizedInterface)
  & $n_x$, $n_y$
    from \textit{(a)} \\
\textit{(c)}~Kernels
  & \texttt{[Kernels]},
    \texttt{[AuxKernels]}
  & \texttt{GrainGrowth} or
    \texttt{GrainGrowthLinearizedInterface}
    module~\cite{tonks2012object};
    \texttt{BndsCalcAux},
    \texttt{FeatureFloodCountAux}
  & \texttt{op\_num},
    $\eta_i$
    from \textit{(b)} \\
\textit{(d)}~Materials
  & \texttt{[Materials]},
    \texttt{[UserObjects]}
  & $L$, $\mu$, $\kappa$ via
    Eq.~\eqref{eq:model_params}~\cite{moelans2008quantitative}
    (\texttt{GBEvolution}) or
    parsed $L$/$\kappa$/$\mu$
    (\texttt{LinearizedInterface});
    \texttt{PolycrystalVoronoi},
    \texttt{GrainTracker}
  & $\{\sigma, w_{\mathrm{GB}},
    M_0, Q\}$ from $\mathcal{M}$;
    mesh from \textit{(a)} \\
\textit{(e)}~Postprocessors
  & \texttt{[Postprocessors]}
  & \texttt{TimestepSize},
    \texttt{NumDOFs},
    \texttt{NumElements}
  & Variables from \textit{(b)},
    UserObjects from \textit{(d)} \\
\textit{(f)}~Executioner
  & \texttt{[Executioner]},
    \texttt{[BCs]},
    \texttt{[Adaptivity]},
    \texttt{[Outputs]}
  & \texttt{PJFNK}~\cite{knoll2004jacobian},
    \texttt{bdf2},
    \texttt{IterationAdaptiveDT},
    $\epsilon_{\mathrm{nl}}$,
    $\epsilon_{\mathrm{l}}$,
    periodic BCs, AMR, CSV
  & $\boldsymbol{\theta}_{\mathrm{solver}}$,
    $\boldsymbol{\theta}_{\mathrm{run}}$
    from $\mathcal{P}$ \\
\bottomrule
\end{tabular}
\end{table}

Beyond parameter transcription, three sub-agents encode physics
directly into the input file.
The Mesh sub-agent~\textit{(a)} selects \texttt{QUAD4} elements
for 2D and \texttt{HEX8} for 3D, and enforces the
interface-resolution constraint $h \leq w_{\mathrm{GB}}/4$
derived from Eq.~\eqref{eq:free_energy}, raising a configuration
error before any file is written if the mesh would under-resolve
the diffuse interface.
The Materials sub-agent~\textit{(d)} branches on formulation:
for \texttt{GBEvolution} it computes $\{L, \mu, \kappa\}$
analytically via Eq.~\eqref{eq:model_params} from the
user-specified set $\{\sigma, w_{\mathrm{GB}}, M_0, Q, T\}$~\cite{schoenfelder1997};
for \texttt{LinearizedInterface} it derives $L$ and $\kappa$
from the same physically grounded copper parameters
$\{M_\mathrm{GB}, \sigma, w_{\mathrm{GB}}\}$ and adds a
\texttt{bound\_value} constraint that stabilizes the linearized
free energy at coarse mesh resolution. (An earlier version of this
branch emitted placeholder material constants rather than the copper
parameters used by the \texttt{GBEvolution} path; this defect, surfaced
by the benchmark and falsified by the Skeptic, is corrected as described
in Section~\ref{sec:benchmark}.)
The Postprocessors sub-agent~\textit{(e)} registers
\texttt{GrainTracker} exclusively in \texttt{[UserObjects]}
rather than \texttt{[Postprocessors]} --- a distinction that
proved critical during failure recovery
(Section~\ref{sec:failure_recovery}) --- and configures
\texttt{TimestepSize}, \texttt{NumDOFs}, and \texttt{NumElements}
as the primary CSV observables from which the Visualization agent
extracts $N(t)$ (Eq.~\eqref{eq:Nt}).

Two additional design choices promote reproducibility and
flexibility. The initial condition type is independently selectable:
\texttt{Voronoi} initializes grains via \texttt{PolycrystalVoronoi}
with a fixed random seed, ensuring bit-for-bit reproducibility
across runs; \texttt{Random} uses \texttt{PolycrystalRandomIC}
with \texttt{random\_type = discrete} for more natural initial
microstructures without a fixed grain count.
For Voronoi ICs, the coloring algorithm is selectable between
\texttt{jp} (Jones--Plassmann~\cite{jones1993parallel}, recommended for
$N_\mathrm{grains} > \texttt{op\_num}$) and \texttt{bt} (backtracking, requires $N_\mathrm{grains} =
\texttt{op\_num}$).

Once all six sub-agents complete their blocks, the assembled
\texttt{.i} file undergoes a final syntax validation pass checking
block completeness, parameter types, and cross-reference
consistency; only a file that passes this check is forwarded to
the Runner agent.
In the grain growth benchmark, agent-generated input files
achieved 9/10 exact block matches against an expert-authored
reference at $T = 450$\,K, with a relative error in the kinetic
rate constant of $\Delta\tilde{k}/\tilde{k} = 0.30$\%
(Section~\ref{sec:input_fidelity}).

%─────────────────────────────────────────────
\subsection{Runner Agent}
\label{sec:runner}
%─────────────────────────────────────────────

Once the Input Writer has produced a validated \texttt{.i} file,
the Runner agent $f_3$ takes sole responsibility for simulation
execution, output organization, and provenance recording
(Fig.~\ref{fig:pipeline}).
It is the only agent that interacts directly with the MOOSE
executable and the only one that writes files to disk.

\begin{wrapfigure}{r}{0.45\textwidth}
  \centering
  \includegraphics[width=0.45\textwidth]{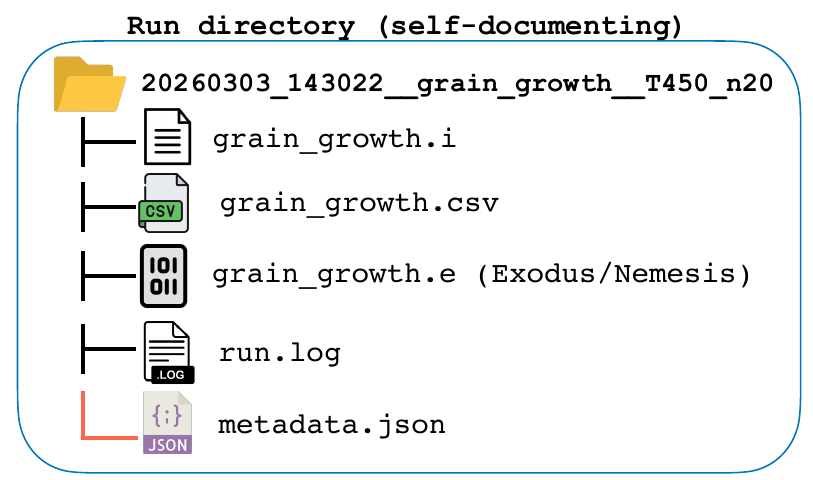}
  \caption{\textbf{AutoMOOSE run directory structure.}
    Each run directory is timestamped and self-contained, comprising:
    \texttt{grain\_growth.i} (complete MOOSE input file),
    \texttt{grain\_growth.csv} (tabulated grain count time series
    $N(t)$, Eq.~\eqref{eq:Nt}),
    \texttt{run.log} (full solver stdout),
    \texttt{metadata.json} (structured provenance record encoding
    all simulation parameters, executable path, hostname, MPI
    configuration, and wall-clock duration), and
    \texttt{record.json} (run status and parsed kinetics metrics).
    Any run can be exactly reproduced by executing the MOOSE command
    (Eq.~\eqref{eq:mpi_cmd}) from within this directory, satisfying
    FAIR data principles by construction~\cite{wilkinson2016}.}
  \label{fig:rundirectory}
\end{wrapfigure}

\subsubsection{Execution and parallelism}
For each simulation case, $f_3$ launches MOOSE as a managed
subprocess,
\begin{equation}
    \texttt{mpiexec -n}\; N_{\mathrm{MPI}}\;
    \texttt{phase\_field-opt -i sim.i},
    \label{eq:mpi_cmd}
\end{equation}
where $N_{\mathrm{MPI}} \geq 1$ is drawn from
$\boldsymbol{\theta}_{\mathrm{run}} \subset \mathcal{P}$.
For temperature sweep requests, $f_3$ expands the parameter range
$\Theta = \{T_i\}_{i=1}^{n}$ populated by the Architect into $n$
independent simulation instances $\{\mathcal{P}_i\}_{i=1}^{n}$,
each assigned a unique run identifier, and dispatches them as
concurrent threads.
The total sweep wall-clock time is therefore
\begin{equation}
    T_{\mathrm{sweep}} \;=\; \max_{i \in \{1,\ldots,n\}} T_i,
    \label{eq:sweep_time}
\end{equation}
rather than the serial sum $\sum_i T_i$, yielding an approximately
$n$-fold reduction in elapsed time relative to sequential execution.

\subsubsection{Run directory and provenance}
Before launching each case, $f_3$ creates a timestamped run
directory and copies the validated \texttt{.i} file into it,
ensuring that every run is self-contained from the moment it
starts --- even a mid-run crash leaves a recoverable directory.
On completion, the directory contains five files
(Fig.~\ref{fig:rundirectory}):
\texttt{sim.i} (full MOOSE input),
\texttt{output.csv} (grain count time series $N(t)$,
  Eq.~\eqref{eq:Nt}),
\texttt{run.log} (full solver stdout),
\texttt{metadata.json} (provenance record encoding all fields of
  $\mathcal{P}_i$, the MOOSE executable path, hostname, MPI rank
  count, random seed, and wall-clock duration $T_i$), and
\texttt{record.json} (run status and parsed metrics, updated
  incrementally during execution).
Any run can be exactly reproduced by executing
Eq.~\eqref{eq:mpi_cmd} from within its directory, without
reference to any external state --- directly satisfying the
reproducibility and FAIR data requirements discussed in
Section~\ref{sec:overview}~\cite{wilkinson2016}.

\subsubsection{Live monitoring and exit routing}
While MOOSE executes, $f_3$ streams solver stdout to the live log
panel in real time, enabling the user to monitor convergence
without polling.
On exit, $f_3$ inspects the process exit code and routes control
accordingly: a non-zero exit code triggers handoff to the Reviewer
agent (Section~\ref{sec:reviewer}), while a zero exit code passes
\texttt{output.csv} directly to the Visualization agent
(Section~\ref{sec:viz}).
This binary routing implements the failure and success paths
shown in Fig.~\ref{fig:pipeline}, and is the junction at which
the pipeline's completion-aware verification screen is engaged.

%─────────────────────────────────────────────

\subsection{Reviewer Agent}
\label{sec:reviewer}
%─────────────────────────────────────────────

When Runner~($f_3$) reports the outcome of a MOOSE invocation, control
passes to the Reviewer agent $f_4$, a completion-aware execution screen.
Its role is operational, not epistemic: it determines whether the run
\emph{executed successfully}, distinguishing a genuine solver breakdown
from the transient adaptive-timestep events --- linear-solve cutbacks
that recover and complete --- that are a normal part of robust nonlinear
solves. A run that reaches its requested integration time is admitted as
executed; a run that aborts or fails to advance in time is flagged as a
failed execution. Judgement about whether an executed run is
\emph{physically credible} is deferred entirely to the Skeptic
(Section~\ref{sec:skeptic}); the Reviewer only asks whether there is a
completed result to assess.

\subsubsection{Execution screening}
$f_4$ evaluates the solver state against the tolerances stored in
$\boldsymbol{\theta}_{\mathrm{solver}} \subset \mathcal{P}$, which
were written into \texttt{[Executioner]} by sub-agent~\textit{(f)}
of the Input Writer (Section~\ref{sec:inputwriter}).
The two primary convergence criteria are the nonlinear residual at
Newton iteration $k$,
\begin{equation}
    \bigl\|\mathbf{R}^{(k)}\bigr\|_2 \;<\; \epsilon_{\mathrm{nl}},
    \label{eq:nl_residual}
\end{equation}
and the linear (Krylov) residual within each Newton step,
\begin{equation}
    \bigl\|\mathbf{r}^{(k)}\bigr\|_2 \;<\; \epsilon_{\mathrm{l}},
    \label{eq:l_residual}
\end{equation}
where $\mathbf{R}^{(k)}$ is the global finite-element residual
assembled from the Allen--Cahn weak form
(Eq.~\eqref{eq:allen_cahn}) and $\mathbf{r}^{(k)}$ is the Krylov
linear system residual.
A persistently large $\|\mathbf{R}^{(k)}\|_2$ indicates that the
current order parameter field $\{\eta_i\}$ has not minimized the
free energy $F$ (Eq.~\eqref{eq:free_energy}) at the prescribed
timestep, directly connecting the numerical failure criterion to
the phase-field physics of Section~\ref{sec:pf}.
Crucially, $f_4$ reads the completion state from the MOOSE log rather
than from the exit code alone: a log reporting \texttt{Finished
Executing} denotes a valid completion even when transient
\texttt{DIVERGED\_ITS} warnings appear earlier from adaptive-timestep
cutbacks, which prevents normal solver behaviour from being misread as
fatal failure.

\subsubsection{Closed-loop recovery module}
The pipeline is paired with a closed-loop recovery module
(\texttt{recovery.py}) that acts on failures with a recognized,
correctable signature.
The module's diagnosis draws on two sources: a run that fails to
complete is classified directly from real solver-log signatures
(non-finite residuals, convergence-failure messages), while a run
that completes but is falsified by the Skeptic is classified from
the physics verdict itself, mapping each violated invariant to a
candidate cause (Section~\ref{sec:skeptic}).
Each diagnosis triggers a bounded parameter-level correction; for
solver divergence, the primary response is initial-timestep
cutback,
\begin{equation}
    \Delta t_0^{(k+1)} \;=\; \alpha\,\Delta t_0^{(k)},
    \qquad \alpha = 0.5,
    \label{eq:dt_cutback}
\end{equation}
floored at a minimum step, together with a stepwise tightening of
the adaptive-cutback factor.
The physical rationale is that a smaller $\Delta t$ reduces the
per-step increment $\Delta\eta_i$, keeping the Newton iterate
within the basin of convergence of the free-energy landscape
defined by $f_{\mathrm{loc}}$ (Eq.~\eqref{eq:floc}).
Other diagnoses map to their own bounded edits --- extending the
integration window when the asymptotic regime is not reached,
refining the mesh when non-physical nucleation is flagged --- and
every correction is capped, logged, and limited to three attempts
per task.
The corrected input is regenerated and re-executed through the
standard pipeline with no shortcut: a corrected run is classified
as credible only if it independently completes execution and
survives Skeptic falsification, so recovery cannot promote a run
past any gate of Section~\ref{sec:overview}.
Two safeguards bound the module's action.
First, its completion classifier is defensive and validated
against real run logs: a run whose log reports completion, or that
reached its target integration time, is never ``corrected'' on
the basis of a stale \texttt{status: running} record in its
metadata.
Second, corrections that do not apply in a given configuration ---
for example, a spatial-resolution parameter that is meaningless in
a two-dimensional run --- are suppressed.
Failures with no recognized signature, or tasks whose correction
budget is exhausted, fall outside this bounded envelope and are
flagged to the user rather than silently reported.
We report the recovery module as implemented and
component-validated; its quantitative behaviour on real failure
logs is described in Section~\ref{sec:failure_recovery}.

%─────────────────────────────────────────────
\subsection{Visualization Agent}
\label{sec:viz}
%─────────────────────────────────────────────

With convergence confirmed by a zero exit code from Runner~($f_3$),
the Visualization agent $f_5$ completes the pipeline by extracting
quantitative observables, fitting the grain growth kinetics model
established in Section~\ref{sec:pf}, and returning both numerical
results and a natural-language interpretation to the user ---
closing the loop from natural-language prompt to
publication-ready analysis.

\subsubsection{Observable extraction and kinetics fitting}
$f_5$ calls the plugin's \texttt{parse\_results()} contract
function to extract the grain count time series $N(t)$ from the
\texttt{output.csv} file produced by \texttt{GrainTracker}
(sub-agent~\textit{(e)}, Section~\ref{sec:inputwriter}).
$f_5$ then fits $N(t)$ to the grain coarsening law
(Eq.~\eqref{eq:Nt}) by linear regression of $N^{-1}(t)$
against $t$~\cite{virtanen2020scipy}, extracting the macroscopic
rate constant $\tilde{k}(T)$ for each temperature $T_i \in \Theta$.
Fit quality is quantified by the coefficient of determination,
\begin{equation}
    R^2 \;=\; 1 \;-\;
    \frac{\sum_{i}\bigl(N_i - \hat{N}_i\bigr)^2}
         {\sum_{i}\bigl(N_i - \bar{N}\bigr)^2},
    \label{eq:r2}
\end{equation}
where $N_i$ are the \texttt{GrainTracker} grain counts, $\hat{N}_i$
are the model predictions from Eq.~\eqref{eq:Nt}, and $\bar{N}$
is the temporal mean.

\subsubsection{Arrhenius regression and consistency check}
Once $\{\tilde{k}(T_i)\}_{i=1}^{n}$ are obtained across the
temperature sweep, $f_5$ performs Arrhenius regression by fitting
Eq.~\eqref{eq:arrhenius} to recover the activation energy
$Q_{\mathrm{fit}}$ and pre-exponential factor $\tilde{k}_0$.
As established in Section~\ref{sec:pf}, $Q_{\mathrm{fit}}$ must
equal the activation energy $Q$ originally specified by the user
in $\mathcal{M} \subset \mathcal{P}$, providing a closed-loop
consistency check of the full pipeline
(Eq.~\eqref{eq:pipeline}).
In the grain growth benchmark, $f_5$ recovered
$Q_{\mathrm{fit}} = 0.296$\,eV against the specified value of
$Q = 0.23$\,eV, with $R^2 = 0.90$--$0.95$ at $T \geq 600$\,K;
the suppressed kinetics at $T = 300$\,K are consistent with the
Arrhenius reduction of $L(T)$ at low temperature
(Eq.~\eqref{eq:mob_arrhenius}), in agreement with the
Burke--Turnbull framework~\cite{burke1952recrystallization}.

\subsubsection{Narrated output}
Beyond numerical metrics, $f_5$ generates a structured
natural-language interpretation of the results, grounding the
kinetics findings in the phase-field model of
Section~\ref{sec:pf} and identifying physically notable
features --- such as the suppression of coarsening kinetics at
low temperature.
This narrated output is reproduced verbatim in
Section~\ref{sec:narrated}.

%─────────────────────────────────────────────
\subsection{Skeptic Agent}
\label{sec:skeptic}

The Skeptic agent $f_6$ is the framework's verification stage. It addresses a
failure mode that the Reviewer ($f_4$) cannot: a simulation may run to
completion and produce well-formed output, yet still be \emph{physically}
wrong. Whereas $f_4$ asks an operational question --- did the run execute? ---
$f_6$ asks an epistemic one: should the result be believed? This reframes the
final stage of the pipeline from parsing to \emph{falsification}, and is a
core methodological contribution of the agentic design: the system does not
merely generate and execute a simulation, it adversarially tests its own output
against the physical laws the result must obey.

Given a completed run, $f_6$ applies a battery of physics-grounded
falsification tests, each a hypothesis that the result must satisfy if it is
physically valid. A test that fails \emph{falsifies} the run; a run that
survives all tests is reported as credible. Crucially, each test is paired with
a diagnosis that maps the failure to a likely physical or numerical cause,
providing the reasoning needed for remediation. The battery is
physics-specific: the dispatcher routes each run to the invariants appropriate
to its governing equations.

\subsubsection{Grain-growth invariants}
For the non-conserved Allen--Cahn dynamics of grain growth, the per-run battery
comprises four invariants. \textbf{Numerical integrity ($T_5$):} the run must
reach its target integration time without fatal solver breakdown; reaching the
requested \texttt{end\_time} is treated as decisive evidence of numerical
validity, and transient adaptive-timestep events (linear-solve cutbacks) that
recover and complete are not failures, whereas an application abort or a solve
that never advances in time is. \textbf{Monotonicity ($T_1$):} the grain count
must be non-increasing over time; a sustained increase indicates non-physical
spontaneous nucleation or a grain-tracker artifact. \textbf{Asymptotic
consistency ($T_2$):} the initial grain count must match the requested number
of seeds, confirming correct initial-condition generation. \textbf{Parabolic
scaling ($T_3$):} the microstructure must obey the Burke--Turnbull coarsening
law, $N^{-1}(t)$ linear in $t$ with a positive rate constant and coefficient of
determination $R^2 \geq 0.90$ (Eq.~\eqref{eq:Nt}). A fifth, cross-run invariant
tests Arrhenius consistency over a temperature sweep ($T_4$): the rate constants
$\tilde{k}(T)$ must increase monotonically with temperature and yield a linear
$\ln\tilde{k}$ versus $1/T$ relation ($R^2 \geq 0.90$), from which the activation
energy is recovered (Eq.~\eqref{eq:arrhenius}).

\subsubsection{Conserved-dynamics invariants}
For the conserved Cahn--Hilliard dynamics of spinodal decomposition, the
invariants are \emph{exact} physical laws, which makes them stronger
falsification tests than the kinetics-based grain-growth checks.
\textbf{Mass conservation ($S_1$):} the integral of the conserved composition
must remain constant to within a numerical tolerance (relative drift
$\leq 10^{-5}$), as required by the Cahn--Hilliard conservation law.
\textbf{Free-energy dissipation ($S_2$):} the total free energy must show a net
decrease consistent with the gradient-flow structure of the dynamics; transient
step-wise increases are admitted only if they remain within numerical round-off,
and a sustained increase beyond tolerance falsifies the run.
\textbf{Phase separation ($S_3$):} the composition field must separate toward the
two-phase equilibrium tie-line, with the extreme compositions approaching the
equilibrium phase concentrations.

\subsubsection{Verdict, diagnosis, and recovery}
A run is reported as credible only if it survives every applicable invariant;
otherwise it is flagged as falsified, annotated with the specific invariants
violated and a diagnosis localizing the likely cause (for example, an
unconverged or non-conservative solve for a mass-conservation failure, or an
integration window too short for the asymptotic regime for a parabolic-scaling
failure). The Skeptic itself only detects and falsifies; it never repairs.
Building on its diagnoses, however, we implement a closed-loop recovery module
(\texttt{recovery.py}) that consumes the Skeptic's verdict and, for the subset of
failures attributable to time-step divergence, maps the diagnosis to a bounded
parameter-level correction --- a reduced initial time step and a tightened
adaptive cutback --- and regenerates and re-executes the simulation through the
standard pipeline. The module is completion-aware and validated against real run
logs: completed runs are never ``corrected,'' divergences that make no time
advance are recognized and routed to correction, and a corrected run is reported
valid only if it independently passes the Skeptic. Failures rooted in
formulation or spatial resolution rather than time-stepping fall outside this
envelope and are flagged rather than forced. By flagging runs that lie outside
the validated envelope --- rather than silently reporting their metrics --- the
Skeptic ensures that only physically credible results enter the downstream
analysis.

\subsection{Plugin Architecture}

\label{sec:plugin}
%─────────────────────────────────────────────

A core design principle of AutoMOOSE is that the six-agent
orchestration pipeline is entirely physics-agnostic.
The connection between the agents and the underlying MOOSE physics
is mediated by a \textit{plugin contract} --- a minimal
two-function interface that any physics module must implement ---
allowing new simulation types to be integrated without modifying
the shared agent infrastructure.

\subsubsection{Plugin contract}
Each plugin implements exactly two functions.
\texttt{generate\_input(**params)~$\to$~str} receives the fields
of $\mathcal{M} \subset \mathcal{P}$ and returns a complete MOOSE
\texttt{.i} input file as a string; it is called by the Input
Writer (Section~\ref{sec:inputwriter}).
\texttt{parse\_results(csv\_data:~dict)~$\to$~dict} receives the
parsed CSV content as a dictionary of time-series arrays and
returns a structured dictionary of observables --- grain count
trajectory $N(t)$, rate constant $\tilde{k}$, goodness-of-fit
$R^2$, peak DOF count, and wall-clock statistics --- and is called
by the Visualization agent (Section~\ref{sec:viz}).
This two-function contract is the complete boundary between the
physics-specific and physics-agnostic layers: the agent pipeline
reads from and writes to $\mathcal{P}$ without direct knowledge of
any MOOSE block or physical model, as all physics-specific logic is
encapsulated within the plugin.

\subsubsection{Plugin metadata}
Each plugin additionally exposes a \texttt{PLUGIN} metadata
dictionary that the framework uses for UI population, agent
configuration, and sweep orchestration.
The metadata registers: a human-readable label; the
\texttt{executable\_key} pointing to the correct MOOSE binary;
a \texttt{params} dictionary of configurable parameters with
default values; a \texttt{presets} library of named parameter
configurations; a \texttt{sweepable} list of parameters available
for parallel sweep dispatch; a \texttt{result\_keys} list
identifying the CSV columns to track; and a
\texttt{system\_prompt} string that specializes the Architect
agent's physics knowledge for this formulation.
This self-describing design means the UI, the Architect's
intent-parsing logic, and the sweep dispatcher are all populated
automatically from plugin metadata, with no hardcoded assumptions
about any specific physics module.

\subsubsection{Grain growth plugin}
The grain growth plugin is the first fully-implemented plugin in
AutoMOOSE.
It supports two Allen--Cahn formulations
(Section~\ref{sec:pf}), two IC types, 2D and 3D geometries,
and seven named presets spanning from a lightweight 2D test
(10 grains, $12\times12$ mesh) to a production 3D HPC
configuration (6000 grains, $180^3$ mesh, 32 MPI ranks, checkpoint
output, and a \texttt{Terminator} that halts execution when the
grain count falls below a user-specified threshold).
\texttt{GBEvolution} uses the multi-order-parameter Allen--Cahn
model~\cite{moelans2008quantitative} with $\{L, \mu, \kappa\}$
computed analytically from $\{\sigma, w_{\mathrm{GB}}, M_0, Q, T\}$
via Eq.~\eqref{eq:model_params}.
\texttt{LinearizedInterface} employs a linearized
formulation~\cite{kim1999} that produces sharper interfaces at
coarser meshes, preferred for large-scale 3D runs.
For 3D geometries, the plugin automatically switches to
\texttt{HEX8} elements, \texttt{parallel\_type = distributed},
and Nemesis-format parallel output with \texttt{PerfGraphOutput}
for HPC profiling.

\subsubsection{Plugin registry and extensibility}
Plugins are loaded dynamically at runtime from the
\texttt{plugins/} directory by \texttt{plugin\_registry.py}, which
scans for Python modules implementing the two-function contract
and registers them by name.
A new physics module requires only two functions, a \texttt{PLUGIN}
metadata dictionary, and a physics-specific system prompt; the
orchestration layer, agent pipeline, MCP server, and UI require no
changes.
Beyond the grain-growth plugin, a second plugin --- Spinodal
Decomposition (conserved Cahn--Hilliard~\cite{cahn1958free}) --- is
fully implemented and validated end-to-end on a realistic Fe--Cr alloy
(Sections~\ref{sec:spinodal} and~\ref{sec:spinodal_results}),
demonstrating that the framework generalizes across the non-conserved
and conserved dynamics. Two further plugins --- Ferroelectric Switching
(Landau--Ginzburg--Devonshire~\cite{wang2019understanding}) and
Solidification (Allen--Cahn dendritic~\cite{karma1998quantitative}) ---
are currently registered as stubs and will be fully implemented in
subsequent releases.

\begin{figure}[t]
  \centering
  \includegraphics[width=0.8\linewidth]{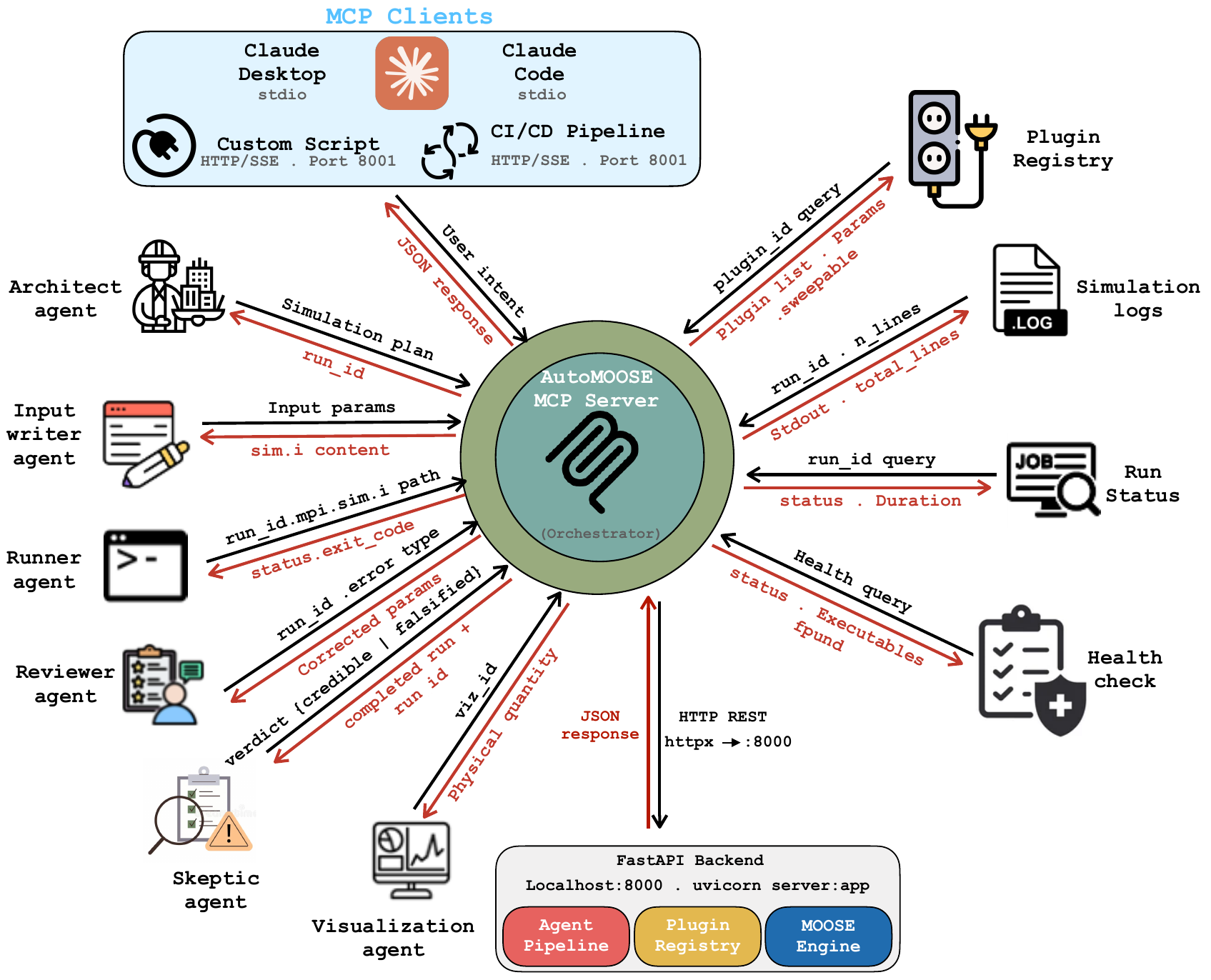}
  \caption{%
    \textbf{Model Context Protocol server architecture of
    AutoMOOSE.}
    The MCP server (port~8001, center) acts as the central
    orchestration hub between external LLM clients and the
    internal agent pipeline $\mathcal{S}$
    (Eq.~\eqref{eq:pipeline}).
    \textit{Top:} four supported client entry points ---
    Claude Desktop, Claude Code, custom scripts, and CI/CD
    pipelines --- all communicating via HTTP and Server-Sent Events (SSE) on port~8001.
    \textit{Left:} the AutoMOOSE agent pipeline
    forms the core intelligence layer; black arrows denote
    control flow (requests and commands) and red arrows denote
    data flow (JSON responses, \texttt{.i} file content, run
    identifiers, and verification verdicts).
    Execution outcomes return through the verification stages
    (the Reviewer's completion screen and the Skeptic's
    falsification); on a recoverable divergence, the closed-loop
    recovery module re-enters the pipeline with a bounded
    correction (Section~\ref{sec:failure_recovery}).
    \textit{Right:} four backend resource endpoints ---
    Plugin Registry, Simulation Logs, Run Status, and Health
    Check --- each mapped to one or more of the ten registered
    MCP tools (Table~\ref{tab:mcp_tools}).
    \textit{Bottom:} the FastAPI backend (port~8000) comprises
    three internal modules --- Agent Pipeline, Plugin Registry,
    and MOOSE Engine --- accessed by the MCP server over HTTP
    REST.
  }
  \label{fig:mcp}
\end{figure}
% ============================================================
% §2.9  Model Context Protocol Interface
% ============================================================
\subsection{Spinodal Decomposition: Model and Fe--Cr Validation}
\label{sec:spinodal}

%%% ========================================================
%%% (spinodal model subsection — validation outcome reported in Results sec:spinodal_results)
%%% ========================================================

To demonstrate that the framework generalizes beyond non-conserved dynamics, a
second plugin (\texttt{plugins/spinodal/plugin.py}) implements spinodal decomposition via the conserved
Cahn--Hilliard equation. Unlike grain growth, the order parameter here is a
conserved composition field $c$, which imposes an exact mass-conservation law
and changes the governing dynamics from Allen--Cahn to Cahn--Hilliard. The
plugin uses the split form, evolving the composition $c$ together with a
chemical potential $w$,
\begin{equation}
    \frac{\partial c}{\partial t} = \nabla\!\cdot\!\bigl(M\,\nabla w\bigr),
    \qquad
    w = \frac{\partial f_{\mathrm{loc}}}{\partial c}
        - \kappa_c \nabla^2 c,
    \label{eq:cahn_hilliard}
\end{equation}
where $M(c)$ is the interdiffusion mobility, $f_{\mathrm{loc}}(c)$ the local
free-energy density, and $\kappa_c$ the gradient-energy coefficient.

The plugin is validated on a realistic alloy system: Fe--Cr spinodal
decomposition at 500\,$^\circ$C, the phase separation underlying the
475--500\,$^\circ$C embrittlement of ferritic steels, with equilibrium
compositions of 23.6 and 82.3\,mol\%\,Cr. The local free energy and mobility use
the CALPHAD (CALculation of PHAse Diagrams) free-energy parameterization for Fe--Cr, with the local free energy
expressed as
\begin{equation}
\begin{aligned}
    f_{\mathrm{loc}}(c) = {}& A c + B(1-c) + C\,c\ln c + D(1-c)\ln(1-c) \\
    & + E\,c(1-c) + F\,c(1-c)(2c-1) + G\,c(1-c)(2c-1)^2,
\end{aligned}
    \label{eq:fecr_floc}
\end{equation}
with coefficients $A$--$G$ and unit-conversion factors as given in the generated
input (Supplementary Section~\ref{sec:SI_spinodal}), and a
concentration-dependent mobility $M(c)$ of the same fitted form. The coefficients
place the initial composition $c_0 = 0.4677$ inside the spinodal (linearly
unstable, $f_{\mathrm{loc}}'' < 0$) region, so the alloy decomposes spontaneously.

From a single natural-language specification, the agent generates the complete
split Cahn--Hilliard input, run on a $25\times25$\,nm$^2$ periodic domain
($100\times100$ \texttt{QUAD4} elements) with an initial condition of
$c_0 = 0.4677$ perturbed by uniform random noise of amplitude $0.02$ to seed the
instability, and the total free energy (bulk plus gradient) accumulated through a
\texttt{TotalFreeEnergy} auxiliary field. The run is scored by the Skeptic against
the conserved-dynamics invariants $S_1$--$S_3$ of
Section~\ref{sec:skeptic}; the validation outcome is reported in
Section~\ref{sec:spinodal_results}.
%%% ========================================================

\subsection{Model Context Protocol Interface}

\label{sec:mcp}

Building on the agent pipeline described in
Sections~\ref{sec:architect}--\ref{sec:plugin}, AutoMOOSE
additionally exposes the full workflow $\mathcal{S}$
(Eq.~\eqref{eq:pipeline}) as a Model Context Protocol (MCP)
server layer (Fig.~\ref{fig:mcp}), enabling headless operation
from Claude Desktop, Claude Code, or any MCP-compatible client
without requiring the browser-based
interface~\cite{anthropic2024claude}.
From the client's perspective, triggering a simulation is
equivalent to invoking Eq.~\eqref{eq:pipeline} remotely: the
client specifies user intent $\mathcal{U}$ as a tool argument,
and the MCP server coordinates $f_1$ through $f_6$ sequentially,
returning the completed plan $\mathcal{P}_i$ and extracted
metrics as a structured JSON response.

\subsubsection{Server architecture}
The MCP server acts as the central orchestration hub, coordinating
agents, plugins, and the execution backend through a unified
JSON-based data flow (Fig.~\ref{fig:mcp}).
It is implemented in Python using Starlette/uvicorn and listens
on port~8001, while the FastAPI~\cite{fastapi2024} simulation
backend runs independently on port~8000.
This decoupled topology allows the MCP layer to be replaced or
extended without modifying either the agent logic or the MOOSE
execution environment.
Two transport modes are supported: \texttt{stdio} transport, in
which the server process is spawned as a child of the
orchestrator and communicates over standard input/output streams
(used by Claude Desktop), and SSE transport, which exposes a
persistent HTTP endpoint for remote clients, CI/CD pipelines, and
custom scripts.
Both modes expose an identical surface of ten tools enumerated in
Table~\ref{tab:mcp_tools}.

\subsubsection{Data flow and tool schemas}
All communication through the MCP server follows a consistent
two-channel pattern (Fig.~\ref{fig:mcp}): black arrows denote
control flow --- requests, commands, and queries --- while red
arrows denote data flow --- JSON responses, \texttt{.i} file
content, run identifiers, and corrected parameters.
Each tool is registered with a JSON Schema describing its input
parameters and return type, which the LLM agent receives at
context initialization via the MCP \texttt{initialize} handshake.
Tool schemas are the sole mechanism by which agents discover
available capabilities: no hard-coded tool names appear in agent
prompts, so adding a new tool to the registry automatically makes
it available to any agent whose system prompt grants permission
to call it.
Tools are divided into two permission classes
(Table~\ref{tab:mcp_tools}): read-only tools ($\dagger$) available
to all agents for querying state, and execution tools ($\ddagger$)
that launch or modify simulation processes, accessible only to
the Runner agent $f_3$ under the orchestrator's permission policy.

\subsubsection{Backend resource endpoints}
The right side of Fig.~\ref{fig:mcp} shows four backend resource
endpoints that the MCP server queries on behalf of agents.
The \textit{Plugin Registry} stores reusable simulation modules
and is queried via plugin identifier and parameter lists,
underpinning the \texttt{generate\_input} and
\texttt{get\_results} tools that map directly onto the two-function
plugin contract (Section~\ref{sec:plugin}).
\textit{Simulation Logs} provide solver stdout and execution
traces, accessed via \texttt{get\_log\_tail} to enable the
Reviewer agent $f_4$ to inspect MOOSE diagnostics without direct
filesystem access.
\textit{Run Status} tracks job state and wall-clock duration,
queried by \texttt{get\_run\_status} using a unique
\texttt{run\_id} assigned at dispatch.
\textit{Health Check} validates system executables and
environment readiness, exposed via \texttt{health\_check} before
any simulation is launched.

\subsubsection{Verification arc}
A key architectural feature visible in Fig.~\ref{fig:mcp} is the
verification arc that follows execution: from Runner~($f_3$) to the
Reviewer~($f_4$) completion screen, and then to the Skeptic~($f_6$)
falsification stage. When $f_3$ reports a run outcome, the MCP server
routes the log to $f_4$, which determines whether the run executed to
completion; every completed run is then passed to $f_6$, which tests it
against the physics invariants it must satisfy. Only runs that both
execute and survive falsification are reported to the client, and
divergence-class failures may additionally be routed to the bounded
recovery module (Section~\ref{sec:reviewer}). This verification arc,
rather than a simple launch-and-return, is what distinguishes AutoMOOSE
from a simulation launcher.

\subsubsection{Headless operation and composability}
The MCP interface enables two usage patterns not supported by the
chat frontend.
First, \textit{headless operation}: an external agent or script
can drive the complete AutoMOOSE pipeline --- constructing a
parameter sweep, dispatching runs, polling status, and retrieving
results --- entirely through structured tool calls, with no human
in the loop and no browser session required.
This is the natural mode for integration with automated materials
discovery pipelines, active learning loops, or high-throughput
screening workflows.
Second, \textit{tool composability}: because MCP tools are
first-class objects in the Claude API tool-use protocol, an
orchestrating agent can interleave AutoMOOSE tool calls with
calls to other MCP servers --- for example, querying a
crystallographic database, retrieving CALPHAD thermodynamic data,
or writing results to a laboratory notebook --- within a single
reasoning context.

\begin{table}[ht]
\centering
\caption{%
  \textbf{MCP tools exposed by the AutoMOOSE server.}
  Tools marked ($\dagger$) are available to all agents;
  tools marked ($\ddagger$) execute or modify simulation state
  and are accessible only to the Runner agent $f_3$ under the
  orchestrator's permission policy.
  Full argument schemas are given in Table~S3
  (Supplementary Information).
}
\label{tab:mcp_tools}
\small
\renewcommand{\arraystretch}{1.3}
\begin{tabular}{@{}lll@{}}
\toprule
\textbf{Tool} & \textbf{Class} & \textbf{Description} \\
\midrule
\texttt{health\_check}    & $\dagger$
  & Verify server and backend availability \\
\texttt{list\_plugins}    & $\dagger$
  & Enumerate registered plugins and their status \\
\texttt{generate\_input}  & $\dagger$
  & Render a MOOSE \texttt{.i} file via the plugin registry \\
\texttt{run\_simulation}  & $\ddagger$
  & Launch a single MOOSE job; return run identifier \\
\texttt{run\_sweep}       & $\ddagger$
  & Fan out a parametric sweep; return run identifiers \\
\texttt{get\_run\_status} & $\dagger$
  & Query the terminal or running state of a job \\
\texttt{get\_results}     & $\dagger$
  & Retrieve parsed postprocessor output and narrative \\
\texttt{list\_runs}       & $\dagger$
  & List all runs with metadata summary \\
\texttt{get\_log\_tail}   & $\dagger$
  & Stream the tail of a MOOSE solver log \\
\texttt{stop\_run}        & $\ddagger$
  & Send a termination signal to a running job \\
\bottomrule
\end{tabular}
\end{table}

% ============================================================
% §2.10  Parameter Sweep Automation
% ============================================================
\subsection{Parameter Sweep Automation}
\label{sec:sweep}

The MCP interface described in Section~\ref{sec:mcp} exposes
sweep execution as a first-class capability through the
\texttt{run\_sweep} tool, but the underlying sweep mechanism
operates at the level of the simulation plan.
The framework supports automated parameter studies by expanding
$\mathcal{P}$ (Eq.~\eqref{eq:simplan}) into a family of
independent cases $\{\mathcal{P}_i\}_{i=1}^{n}$ indexed over a
user-defined parameter range $\Theta = \{\theta_i\}_{i=1}^{n}$.
For the temperature sweep demonstrated here,
$\theta_i = T_i \in \{300, 450, 600, 750\}$\,K, and the
Architect agent populates $\boldsymbol{\theta}_{\mathrm{run}}$
in each $\mathcal{P}_i$ with the corresponding temperature.

The physical significance of this parameterization flows directly
through the model: each $T_i$ propagates through
Eq.~\eqref{eq:mob_arrhenius} to set the Allen--Cahn mobility
$L(T_i)$, which in turn determines the macroscopic rate constant
$\tilde{k}(T_i)$ extracted by the Visualization agent $f_5$ via
Eq.~\eqref{eq:Nt}.
The total sweep wall-clock time is bounded by
Eq.~\eqref{eq:sweep_time}, since all cases are dispatched
concurrently.
Once complete, aggregated results
$\{\tilde{k}(T_i),\,R^2_i\}_{i=1}^{n}$ are compiled
automatically and passed to the Arrhenius fit of
Eq.~\eqref{eq:arrhenius}, recovering the activation energy
$Q_{\mathrm{fit}}$ and closing the quantitative loop from
user-specified grain boundary parameters through the full
pipeline.
Each case retains a complete \texttt{metadata.json} provenance
record encoding all fields of $\mathcal{P}_i$, the hostname,
MOOSE executable path, wall-clock duration, and output file
locations, ensuring full reproducibility from the run directory
alone~\cite{wilkinson2016}.

% ============================================================
% §2.11  User Interface and Demonstration Workflow
% ============================================================
\subsection{Benchmark Design and Execution}
\label{sec:benchmark}

%%% ========================================================
%%% (benchmark design subsection)
%%% ========================================================

The framework is evaluated on a benchmark of 25 end-to-end grain-growth tasks
spanning six regimes: a core temperature and grain-count sweep, spatial
resolution and domain variations, the two Allen--Cahn formulations
(\texttt{GBEvolution} and \texttt{LinearizedInterface}), robustness across
initial-condition seeds, three-dimensional cases, and high-temperature and
high-density stress tests. Each task is specified by a natural-language prompt
and carried through the full pipeline --- generation, execution, and
verification --- with no human intervention. The complete prompt set, listed
verbatim, and the per-task gate outcomes are given in
Supplementary Section~\ref{sec:SI_evalset}.

Each task is scored against five pre-registered, machine-checkable gates, with
no human judgement: (G1) the agent produces a parseable MOOSE input; (G2) the
simulation runs to completion without solver divergence; (G3) the microstructure
coarsens (final grain count below initial); (G4) parabolic Burke--Turnbull
kinetics are recovered ($R^2 \geq 0.90$, Eq.~\eqref{eq:Nt}); and (G5) the
fitted rate constant is physically valid ($\tilde{k} > 0$). We report two tiers:
a \emph{valid-and-runs} rate (G1--G3) and a stricter \emph{full} rate
(G1--G5), the latter additionally requiring quantitative kinetics, which depend
on production-length integration to develop the asymptotic regime. Each task is,
in addition, independently falsification-tested by the Skeptic
(Section~\ref{sec:skeptic}), so that a task counts as fully credible only if it
both clears the gates and survives falsification.

The grain-growth physics uses the MOOSE \texttt{GBEvolution} module with copper
parameters from Sch\"onfelder \textit{et al.}~\cite{schoenfelder1997}:
$\sigma = 0.708$\,J\,m$^{-2}$, $w_{\mathrm{GB}} = 14$\,nm,
$M_0 = 2.5\times10^{-6}$\,m$^4$\,J$^{-1}$\,s$^{-1}$, $Q = 0.23$\,eV. Solver
settings are PJFNK~\cite{knoll2004jacobian} with BDF2 time integration, ASM
preconditioning, $\epsilon_{\mathrm{nl}} = 10^{-8}$,
$\epsilon_{\mathrm{l}} = 10^{-4}$, and adaptive timestepping
(\texttt{IterationAdaptiveDT}). The full suite is executed at production
integration length on NERSC Perlmutter, with each of the 25 tasks assigned a
dedicated compute node within a single 12-hour batch allocation.

\subsubsection{Benchmark-driven generator corrections}
Executing the suite surfaced two systematic input-generation defects, each
falsified by the Skeptic rather than reported as a metric. First, the
\texttt{LinearizedInterface} formulation emitted placeholder material constants
rather than the copper \texttt{GBEvolution} parameters used elsewhere, producing
an interface mobility orders of magnitude too large; the stiff
variational-inequality solve then diverged on the first nonlinear step with no
time advance. Second, the three-dimensional cases diverged under the default
time-stepping (linear solves failing with \texttt{DIVERGED\_ITS}), the step size
and solver settings inherited from the two-dimensional configuration being too
aggressive for the stiffer three-dimensional system. Both were diagnosed from
the solver error signatures together with the Skeptic's verdicts and corrected
in the generator --- the linearized-interface block now uses the physically
grounded copper parameters, and the three-dimensional path uses a smaller initial
time step with a stronger preconditioner --- and verified by re-execution. The
benchmark results reported below reflect the original configuration in which
these cases failed; the corrections are forward-looking improvements committed to
the public repository, an instance of the benchmark functioning as a diagnostic
that drives a falsification-to-fix loop rather than merely a pass/fail score.
%%% ========================================================

\subsection{User Interface and Demonstration Workflow}
\label{sec:ui}

To make the agent pipeline accessible to domain scientists
without programming expertise, AutoMOOSE provides a browser-based
chat interface that exposes the full workflow through six
interactive panels.
Figure~\ref{fig:pipeline} illustrates this interface using the
four-temperature grain growth sweep
($T \in \{300, 450, 600, 750\}$\,K) as a representative
demonstration.

The workflow begins in the \textit{Chat panel}
(Fig.~\ref{fig:pipeline}a), where the user submits a single
natural-language prompt specifying the sweep temperatures,
physical model, domain geometry, and analysis goal --- without
any MOOSE syntax knowledge.
The Architect agent $f_1$ parses this prompt and auto-populates
the \textit{Configure panel} (Fig.~\ref{fig:pipeline}b) with the
corresponding simulation plan $\mathcal{P}$
(Eq.~\eqref{eq:simplan}), mapping user intent onto physical
parameters $\{\sigma, w_{\mathrm{GB}}, M_0, Q\}$ and numerical
settings $\{\boldsymbol{\theta}_{\mathrm{solver}},
\boldsymbol{\theta}_{\mathrm{run}}\}$; the user may inspect or
override any field before launching.

With the plan confirmed, the Runner agent $f_3$ dispatches four
independent MOOSE processes concurrently via the sweep
orchestrator, visible as parallel run cards in the
\textit{Run Sidebar} (Fig.~\ref{fig:pipeline}c), achieving total
wall-clock time $T_{\mathrm{sweep}} = \max_i T_i$
(Eq.~\eqref{eq:sweep_time}) rather than the serial sum
$\sum_i T_i$.
Solver progress is streamed in real time to the
\textit{Live Log panel}, where the Reviewer agent $f_4$ monitors
nonlinear residual convergence
$\|\mathbf{R}^{(k)}\|_2 < \epsilon_{\mathrm{nl}}$
(Eq.~\eqref{eq:nl_residual}) and triggers timestep cutback
$\Delta t^{(k+1)} = \alpha\,\Delta t^{(k)}$
(Eq.~\eqref{eq:dt_cutback}) on convergence failure.
The agent-generated \texttt{.i} input file is simultaneously
accessible in the \textit{Input File viewer}
(Fig.~\ref{fig:pipeline}f), assembled by $f_2$ following the
block order of Eq.~\eqref{eq:dag_order} with physical parameters
from $\mathcal{M}$ converted to phase-field coefficients
$\{L, \mu, \kappa\}$ via Eq.~\eqref{eq:model_params}.

Upon completion, the Visualization agent $f_5$ populates the
\textit{Results panel} (Fig.~\ref{fig:pipeline}d) with $N(t)$
curves for all four temperatures fitted to the grain coarsening
law (Eq.~\eqref{eq:Nt}).
Coarsening kinetics are recovered at $T \geq 450$\,K, with
$R^2 = 0.90$--$0.95$ at $T \geq 600$\,K, consistent with
Burke--Turnbull theory~\cite{burke1952recrystallization}; the
zero coarsening at $T = 300$\,K is attributable to Arrhenius
suppression of $L(T)$ at low temperature
(Eq.~\eqref{eq:mob_arrhenius}).
The complete workflow --- from natural-language prompt to
publication-ready kinetics analysis --- requires no MOOSE syntax
knowledge and completes in under two hours on a standard
workstation for the four-temperature sweep demonstrated here.

% ============================================================
% §3  Results
% ============================================================

\section{Results}
\label{sec:results}

We validate AutoMOOSE at several complementary levels. We first
demonstrate the full agent pipeline end-to-end on a copper
polycrystalline grain-growth workflow --- from a single
natural-language prompt to quantitative kinetics analysis, with no
manual file editing or direct solver interaction
(Section~\ref{sec:workflow_demo}). We then evaluate breadth and
reliability on a pre-registered 25-task benchmark with
machine-checkable gates and independent Skeptic falsification
(Section~\ref{sec:benchmark_results}), and establish generalization
to a second, conserved-dynamics domain --- Fe--Cr spinodal
decomposition (Section~\ref{sec:spinodal_results}). The remaining
subsections examine the mechanisms behind these results: input-file
fidelity (Section~\ref{sec:input_fidelity}), grain-coarsening
kinetics (Section~\ref{sec:kinetics}), ensemble Arrhenius recovery
of the activation energy (Section~\ref{sec:arrhenius}), failure
detection and recovery (Section~\ref{sec:failure_recovery}), a
controlled baseline against bare frontier models
(Section~\ref{sec:baseline}), the role of Skeptic falsification as
methodology (Section~\ref{sec:skeptic_ablation}), and agent-narrated
interpretation (Section~\ref{sec:narrated}).

% ============================================================
% §3.1  End-to-End Workflow Demonstration
% ============================================================
\subsection{End-to-End Workflow Demonstration}
\label{sec:workflow_demo}

The benchmark is initiated by submitting the following
natural-language prompt to the AutoMOOSE chat interface:
\begin{quote}
\itshape
Run a grain growth simulation at
$T = \{300, 450, 600, 750\}$\,K with
$\sigma = 0.708$\,J\,m$^{-2}$,
$w_\mathrm{GB} = 14$\,nm,
$M_0 = 2.5\times10^{-6}$\,m$^4$\,J$^{-1}$\,s$^{-1}$,
$Q = 0.23$\,eV~\cite{schoenfelder1997}.
Use 15 Voronoi grains on a $1000\times1000$\,nm$^2$ domain
with a $12\times12$ mesh (uniform refinement level~3)
and periodic boundary conditions.
\end{quote}
From this single prompt, the six-agent pipeline executes the
complete simulation campaign without any further user input.
The Architect ($f_1$) parses the sweep intent and decomposes
the request into four independent simulation tasks, each
populating a separate plan $\mathcal{P}_i$
(Eq.~\eqref{eq:simplan}).
The Input Writer ($f_2$) generates a syntactically valid,
physically correct \texttt{.i} file for each task via its six
specialized sub-agents, converting the user-specified parameters
$\{\sigma, w_\mathrm{GB}, M_0, Q\}$ to phase-field coefficients
$\{L, \mu, \kappa\}$ through Eq.~\eqref{eq:model_params}.
The Runner ($f_3$) dispatches all four jobs concurrently via the
\texttt{run\_sweep} MCP tool. Because the cases are mutually independent,
the sweep is embarrassingly parallel and its wall-clock time is bounded
by the slowest single run (Eq.~\eqref{eq:sweep_time}),
$T_\mathrm{sweep} \approx 22{,}820$\,s, rather than the serial sum
($\sum_i T_i \approx 41{,}261$\,s; Table~\ref{tab:sweep_timing}), a
$1.8\times$ reduction in wall-clock time. We emphasize that this is a
workflow-level scheduling effect from concurrent job dispatch, not a
solver-level acceleration: AutoMOOSE does not modify or speed up MOOSE's
numerics, and the figure simply reflects that independent sweep cases run
in parallel rather than in sequence.
Where a run fails to execute, the Reviewer ($f_4$) flags it and, for
recoverable time-step divergences, the closed-loop recovery module
applies a bounded correction (Section~\ref{sec:failure_recovery}).
On successful completion, the Visualization agent ($f_5$)
extracts grain-count trajectories from the \texttt{GrainTracker}
CSV output, fits the coarsening law (Eq.~\eqref{eq:Nt}) to
recover $\{\tilde{k}(T_i)\}$, and performs Arrhenius regression
to extract $Q_\mathrm{fit}$.

Table~\ref{tab:pipeline_perf} summarises agent-level performance
across the benchmark.
The complete pipeline --- from prompt submission to
publication-ready kinetics analysis --- required approximately
$380$\,min ($\approx 6.3$\,h) of total wall-clock time,
dominated by MOOSE compute time at $T = 750$\,K ($22{,}820$\,s).
The active human effort in AutoMOOSE is limited to composing
the single prompt; by contrast, an experienced researcher
hand-authoring four input files and performing kinetics analysis
manually would require comparable active scripting effort
in addition to the same compute time.
The adaptive time-stepper grew $\Delta t$ from 25\,ns to
$\sim\!284$\,ns as grain boundary curvature diminished, with
peak mesh resolution of $\sim\!261{,}000$ degrees of freedom
per run.

\begin{table}[t]
\centering
\caption{\textbf{AutoMOOSE agent pipeline performance} on the
         four-temperature grain growth benchmark
         ($T \in \{300, 450, 600, 750\}$\,K).
         First-attempt success: correct output without Reviewer
         intervention.
         Timing on a MacBook Pro (Apple M-series, single core).}
\label{tab:pipeline_perf}
\renewcommand{\arraystretch}{1.25}
\begin{tabular}{@{}llcc@{}}
\toprule
Agent & Stage & 1\textsuperscript{st}-attempt & Key metric \\
\midrule
$f_1$ Architect
  & Intent parsing
  & 1/1
  & Correct sweep intent, formulation, BCs \\
$f_2$ Input Writer
  & \texttt{.i} file generation
  & 4/4
  & 6/12 blocks exact match
    (Section~\ref{sec:input_fidelity}) \\
$f_3$ Runner
  & Parallel execution
  & 4/4
  & $T_\mathrm{sweep} \approx 22{,}820$\,s
    ($1.8\times$ vs.\ serial, scheduling only) \\
$f_4$ Reviewer
  & Completion screen
  & ---
  & 3/3 failed runs flagged
    (Section~\ref{sec:failure_recovery}) \\
$f_5$ Visualization
  & Kinetics extraction
  & 4/4
  & $R^2 = 0.90$--$0.95$ at $T \geq 600$\,K
    (Section~\ref{sec:kinetics}) \\
\midrule
\multicolumn{2}{@{}l}{\textbf{End-to-end}}
  & \textbf{4/4}
  & Prompt $\to$ kinetics analysis: $\approx 380$\,min \\
\bottomrule
\end{tabular}
\end{table}

\begin{table}[t]
\centering
\caption{\textbf{Per-run execution statistics} for the
         four-temperature grain growth sweep on a MacBook Pro
         (Apple M-series, single core per run).
         All runs dispatched concurrently via the
         \texttt{run\_sweep} MCP tool.
         $N_f$: final grain count at $t = 4000$\,ns;
         DOFs: peak degrees of freedom.
         $T = 300$\,K shows no coarsening due to Arrhenius
         suppression (Section~\ref{sec:kinetics}).
         The $1.8\times$ reduction is a workflow-level scheduling
         effect --- independent cases run concurrently, so wall-clock
         time is bounded by the slowest run --- and not a solver-level
         acceleration of MOOSE.}
\label{tab:sweep_timing}
\begin{tabular}{@{}lrrrr@{}}
\toprule
$T$ (K) & Timesteps & $N_f$ & DOFs (peak) & Wall time (s) \\
\midrule
300 &  31 & 15 & 260{,}124 &      509 \\
450 &  85 & 13 & 261{,}862 &    4{,}165 \\
600 & 306 &  6 & 261{,}086 &   13{,}768 \\
750 & 602 &  3 & 260{,}832 &   22{,}820 \\
\midrule
\multicolumn{4}{@{}l}{Serial total}     & 41{,}261 \\
\multicolumn{4}{@{}l}{Concurrent (bounded by slowest run)} & 22{,}820 \\
\multicolumn{4}{@{}l}{Wall-clock reduction (scheduling)} & $1.8\times$ \\
\bottomrule
\end{tabular}
\end{table}

% ============================================================
% §3.2  Input File Fidelity
% ============================================================
\subsection{Benchmark Results}
\label{sec:benchmark_results}

We evaluate the framework on the 25-task pre-registered benchmark of
Section~\ref{sec:benchmark}, executed at production integration length on
NERSC Perlmutter (one dedicated node per task, single 12-hour allocation),
and score each task against the five machine-checkable gates G1--G5 plus
independent Skeptic falsification. The aggregate outcome is summarized per
regime in Table~\ref{tab:benchmark_regime}; the full per-task gate outcomes and
the Skeptic verdict for each task are given in
Supplementary Section~\ref{sec:SI_evalset}.

All 25 tasks produce a parseable MOOSE input (G1 $=25/25$), confirming that
the generator reliably emits syntactically valid simulations across every
regime. Nineteen of 25 run to completion and coarsen (the \emph{valid-and-runs}
tier, G1--G3 $=19/25$), and 15 of 25 additionally satisfy quantitative
parabolic kinetics and a positive rate constant (the \emph{full} tier,
G1--G5 $=15/25$); independent Skeptic falsification corroborates exactly this
set of 15 credible runs. The parabolic-kinetics gate is recovered for 18 of 25
tasks at production integration length (G4 $=18/25$), a regime that short
under-integrated runs do not reach.

The failures are localized and interpretable rather than diffuse. The four
\texttt{LinearizedInterface} (formulation) tasks generate valid input and
coarsen correctly but do not complete within the wall-clock window in the
configuration benchmarked, owing to the placeholder-material defect diagnosed
in Section~\ref{sec:benchmark}; the two three-dimensional tasks diverge under
the inherited two-dimensional time-stepping. Both failure modes were
falsified by the Skeptic and corrected in the generator as forward-looking
fixes (Section~\ref{sec:benchmark}). The remaining sub-gate misses are a small
number of low-grain-count and robustness cases that coarsen correctly but fall
marginally below the $R^2 \geq 0.90$ kinetics threshold. Concentrating the
failures in two well-understood mechanisms --- rather than scattering them
across regimes --- is what makes the benchmark useful as a diagnostic.

\begin{table}[t]
\centering\small
\caption{AutoMOOSE performance across the $n{=}25$ benchmark regimes (production
integration length; each task on a dedicated Perlmutter node within a 12-hour
allocation). G1: valid input generated; G2: ran to completion within the
wall-clock window; G3: coarsened ($N_f<N_0$); G4: parabolic (Burke--Turnbull)
kinetics ($R^2\ge0.90$); G5: positive rate constant. \emph{Valid \& runs}
counts G1--G3; \emph{Full} counts all five gates.}
\label{tab:benchmark_regime}
\begin{tabular}{lcccccccc}
\toprule
Regime & $n$ & G1 & G2 & G3 & G4 & G5 & Valid\,\&\,runs & Full\\
\midrule
Core ($T$, $N$ sweep) & 8 & 8 & 8 & 8 & 6 & 8 & 8/8 & 6/8\\
Resolution / domain & 4 & 4 & 4 & 4 & 4 & 4 & 4/4 & 4/4\\
Formulation (Lin.\ Interface) & 4 & 4 & 0 & 4 & 3 & 4 & 0/4 & 0/4\\
Robustness (seeds) & 4 & 4 & 4 & 4 & 2 & 4 & 4/4 & 2/4\\
3D & 2 & 2 & 0 & 0 & 0 & 1 & 0/2 & 0/2\\
High-$T$ stress & 2 & 2 & 2 & 2 & 2 & 2 & 2/2 & 2/2\\
Density stress & 1 & 1 & 1 & 1 & 1 & 1 & 1/1 & 1/1\\
\midrule
\textbf{Total} & \textbf{25} & \textbf{25} & \textbf{19} & \textbf{23} & \textbf{18} & \textbf{24} & \textbf{19/25} & \textbf{15/25}\\
\bottomrule
\end{tabular}
\end{table}

\subsection{Second Physics Domain: Spinodal Decomposition}
\label{sec:spinodal_results}

\begin{figure}[htbp]
  \centering
  \includegraphics[width=\linewidth]{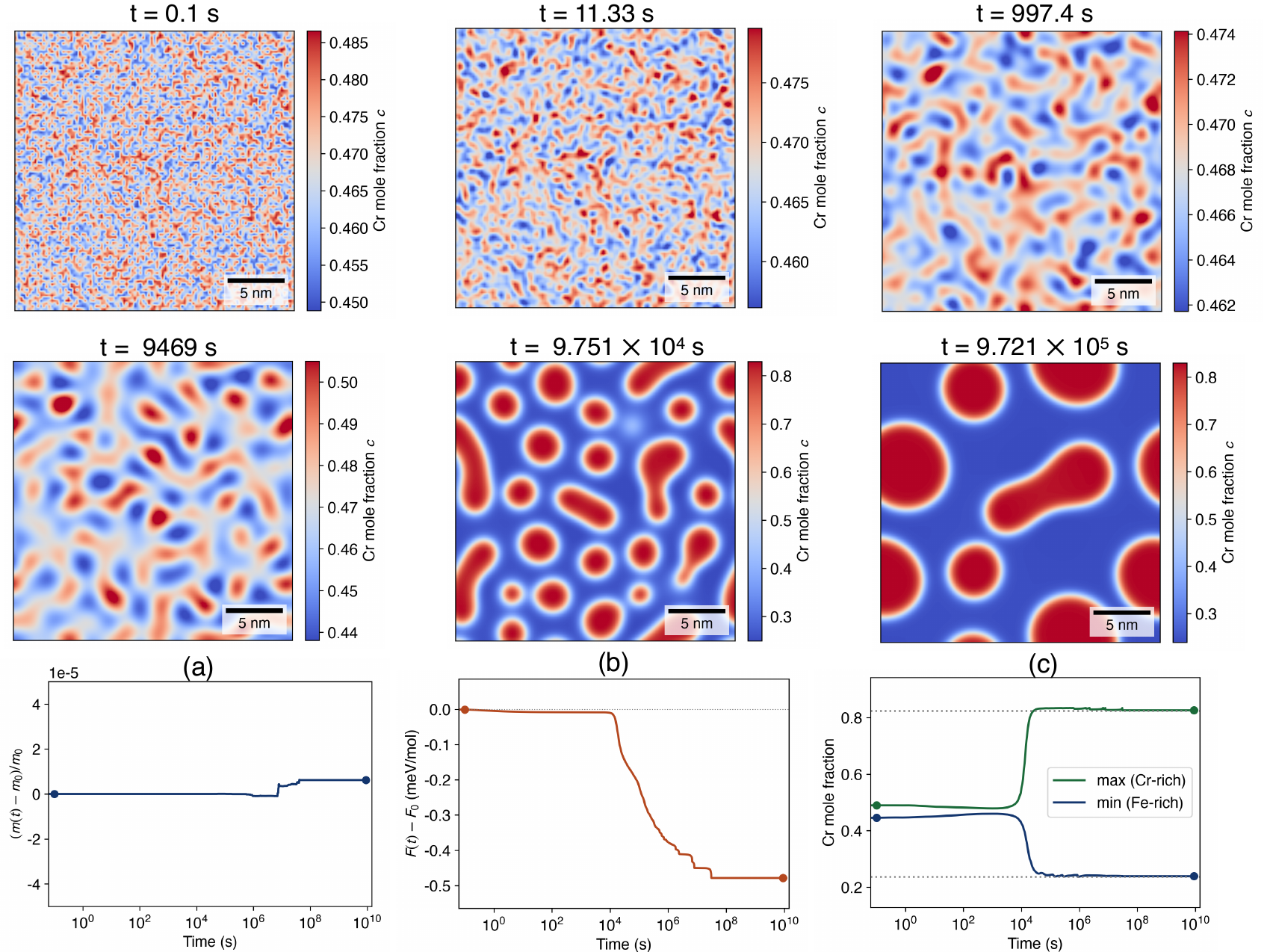}
  \caption{\textbf{Fe--Cr spinodal decomposition (500\,$^\circ$C): composition
  evolution and Skeptic validation invariants.} Top two rows: composition field
  $c(x,y)$ at increasing times, from the noisy initial condition through an
  interconnected spinodal pattern to grown, well-separated Cr-rich domains (each
  panel independently scaled to its composition range). Bottom row: (a) $S_1$
  Cr-mass conservation, relative drift $6.28\times10^{-6}$, within the $10^{-5}$
  tolerance; (b) $S_2$ free-energy dissipation, net decrease to a stable plateau
  (transient round-off increases within tolerance); (c) $S_3$ phase separation
  toward the equilibrium tie-line (Cr-rich $0.823$, Fe-rich $0.236$\,mol
  fraction).}
  \label{fig:fecr_spinodal}
\end{figure}

To establish that the framework generalizes beyond the non-conserved
Allen--Cahn dynamics of grain growth, we apply it to a second, physically
distinct domain: Fe--Cr spinodal decomposition under the conserved
Cahn--Hilliard dynamics (Eq.~\eqref{eq:cahn_hilliard}) of
Section~\ref{sec:spinodal}. From a single
natural-language specification, the agent generated the complete split
Cahn--Hilliard input, which was executed on a $25\times25$\,nm$^2$ periodic
domain ($100\times100$ \texttt{QUAD4} elements) from an initial composition
$c_0 = 0.4677$ seeded with $0.02$-amplitude noise. The composition field
evolves through the characteristic stages of spinodal decomposition
(Fig.~\ref{fig:fecr_spinodal}): an initially noisy, near-uniform field develops
an interconnected Cr-rich/Fe-rich modulation, which then breaks into discrete
Cr-rich domains that grow and separate.

The decisive test, however, is quantitative. Because the dynamics are
conserved, the Skeptic's invariants here are \emph{exact physical laws} rather
than the kinetics-based heuristics used for grain growth, making them a more
stringent falsification battery (Table~\ref{tab:fecr_si}). The run satisfies all
three: the integral of the Cr composition is conserved to a relative drift of
$6.28\times10^{-6}$, within the Skeptic's $S_1$ tolerance of $10^{-5}$; the total
free energy (bulk plus gradient) shows a net decrease of $0.4\%$ to a stable
plateau, with transient step-wise increases at a small number of steps that
remain within numerical round-off, consistent with the gradient-flow structure
of the dynamics ($S_2$); and the composition separates toward the two-phase
equilibrium tie-line, the extreme compositions reaching $c_{\max}=0.826$
(Cr-rich) and $c_{\min}=0.239$ (Fe-rich) against the equilibrium targets of
$0.823$ and $0.236$\,mol fraction Cr ($S_3$), with a single Cr-rich precipitate
at the final time. That an agent-generated simulation reproduces an exact
conservation law to a stated numerical tolerance in a second physical domain ---
without any change to the agent pipeline --- is the strongest evidence that the
framework's reasoning is physics-general rather than tuned to a single problem.

\begin{table}[t]
\centering\small
\caption{Fe--Cr spinodal-decomposition (Cahn--Hilliard) validation invariants
from the AutoMOOSE run (500\,$^\circ$C). Equilibrium target concentrations
23.6 / 82.3\,mol\%\,Cr; the Cr-mass drift is assessed against the Skeptic $S_1$
tolerance ($10^{-5}$).}
\label{tab:fecr_si}
\begin{tabular}{lc}
\toprule
Invariant & Value\\
\midrule
Relative Cr-mass drift & $6.28\times10^{-6}$\\
Free-energy decrease & 0.4\%\\
Energy monotone (transient increasing steps) & within tol.\ (15)\\
Final Cr range & $[0.239,\,0.826]$\\
Cr-rich precipitates (final) & 1\\
\bottomrule
\end{tabular}
\end{table}

\subsection{Input File Fidelity}
\label{sec:input_fidelity}

Having established the aggregate benchmark outcome and generalization to a second
physics domain, we now examine the mechanisms underlying these results in detail:
how faithfully the agent reproduces expert input structure
(Section~\ref{sec:input_fidelity}), the grain-growth kinetics it recovers
(Section~\ref{sec:kinetics}), and the activation energy it extracts
(Section~\ref{sec:arrhenius}). These subsections use a single representative
worked example for transparency, complemented by the at-scale statistics that
confirm the example generalizes.

\begin{figure}[htbp]
  \centering
  \includegraphics[width=0.5\linewidth]{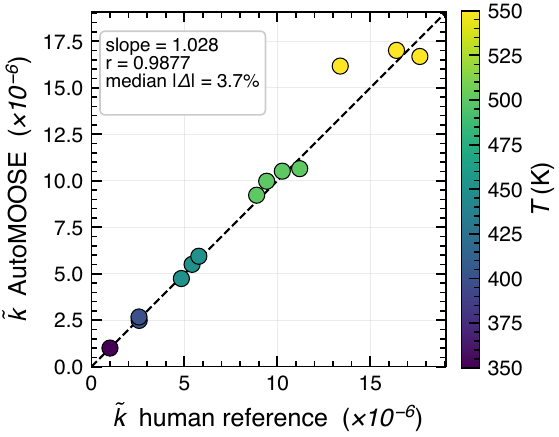}
  \caption{\textbf{Matched-cell parity between AutoMOOSE-generated and
  human-reference runs.} Across 14 matched grain-count/temperature cells, the
  agent-generated and human kinetic outputs agree closely (slope $1.023$,
  Pearson $r = 0.9887$, median absolute deviation $3.4\%$), confirming that
  structural fidelity carries through to functionally equivalent simulation
  output.}
  \label{fig:fidelity_compare}
\end{figure}

Two human-written baselines are used for complementary purposes
throughout this section, and we keep them clearly separate.
For structural input file fidelity (this subsection), we compare
against an expert-authored reference adapted from the canonical
MOOSE grain growth tutorial~\cite{moose_examples} ($44\times44$
mesh, 20 grains) --- the standard community benchmark for
evaluating whether an agent reproduces established MOOSE
input structure.
For kinetics and Arrhenius benchmarking
(Sections~\ref{sec:kinetics}--\ref{sec:arrhenius}), we use a
separately constructed physics-matched reference with identical
mesh ($12\times12$, refinement level~3) and grain count
($N_0 = 15$) to the AutoMOOSE runs, enabling a controlled
comparison of simulation output.
The two baselines answer different questions and must not be
conflated: the structural comparison asks whether the agent
produces a \emph{well-formed, expert-equivalent input file}, while
the kinetics comparison asks whether two physics-matched runs
produce the \emph{same quantitative output}. The kinetics
comparison is necessarily physics-matched (identical mesh and
grain count) because any difference in those parameters would
itself change the kinetics; the structural comparison deliberately
is not, because the tutorial reference and the prompt specify
different mesh and grain counts by design.

This raises an important point about what the structural comparison
measures. Some differences between the agent's file and the tutorial
reference arise not from agent error but from the prompt
\emph{intentionally} specifying different values --- a $12\times12$
mesh and 15 grains rather than the tutorial's $44\times44$ and 20.
Counting these prompt-driven differences as fidelity failures would
penalize the agent for correctly following the user's instructions.
We therefore score fidelity only over the parameters the agent is
\emph{free to choose}: the set of prompt-specified parameters
($\{$\texttt{nx}, \texttt{ny}, \texttt{xmax}, \texttt{ymax},
\texttt{grain\_num}, \texttt{rand\_seed}, $T$, GB properties,
\texttt{end\_time}, \texttt{auto\_direction}$\}$) is excluded from
the mismatch count, so a block is judged on whether the agent
reproduced the reference's \emph{non-prompt} parameters, not on
whether it echoed values the user deliberately changed. A block in
which the only differences are prompt-driven is therefore classified
as an exact match.

A necessary condition for any agentic simulation framework is
that autonomously generated input files are not only
syntactically valid but physically correct --- faithfully
encoding the user's parameters, boundary conditions, and solver
settings without manual intervention.
To evaluate this, we perform a block-by-block structural
comparison at $T = 450$\,K with 15 Voronoi grains,
GBEvolution formulation, and a $12\times12$ mesh
with uniform refinement level~3.

The comparison is computed by a deterministic scoring script
(provided in the repository, with its per-block output in
\texttt{diff\_summary.json}) that parses
each file into its top-level blocks, extracts the
key--value parameter pairs of each block (comments stripped), and
classifies every reference block by its number of mismatches
against the agent file. A mismatch is a reference parameter that is
either absent from the agent's block or present with a different
value, with prompt-specified parameters excluded as described above.
The block-level classes are defined precisely as:
\emph{exact} ($\checkmark$), zero mismatches --- every non-prompt
reference parameter is reproduced identically; \emph{functionally
equivalent} ($\approx$), exactly one mismatch --- a single
non-prompt parameter differs in a way that does not change the
simulated physics (for example a postprocessor reordering or a
visualization-only output flag); and \emph{physically valid
difference} ($\times$), two or more mismatches that nonetheless
correspond to a legitimate alternative configuration (for example a
different but valid preconditioner). The third class is reserved for
differences that a domain expert would accept as correct, not for
errors; genuine errors would instead produce a file that fails the
\texttt{-{}-check-input} validity gate, which none of the reported
inputs do.

Of 12 structural blocks compared, 6 match exactly, 4 are
functionally equivalent with a single minor parameter difference,
and 2 differ in solver and mesh settings, as summarised in
Table~\ref{tab:input_fidelity}.
The complete annotated listing is provided in
Figure~\ref{fig:S_full_input_comparison} (Supplementary
Section~\ref{sec:SI_inputcomp}).
The six exact matches cover all physics-critical blocks:
\texttt{[UserObjects]}, \texttt{[Materials]}, \texttt{[BCs]},
\texttt{[AuxVariables]}, \texttt{[AuxKernels]}, and
\texttt{[Modules]}.
Notably, AutoMOOSE correctly instantiates \texttt{GBEvolution}
with experimentally grounded copper parameters from
Schönfelder \textit{et al.}~\cite{schoenfelder1997} and
registers \texttt{GrainTracker} exclusively as a
\texttt{UserObject} --- its erroneous duplication in
\texttt{[Postprocessors]} is a well-known source of runtime
failure in hand-written MOOSE files that AutoMOOSE avoids by
construction.
Periodic boundary conditions are correctly enforced via
\texttt{auto\_direction}.

The four approximate matches reflect minor differences with no
effect on simulation output: postprocessor entry reordering,
a \texttt{coarsen\_fraction} difference (\texttt{0.1}
vs.\ \texttt{0.05}), and AutoMOOSE's automatic addition of
\texttt{exodus\,=\,true} in \texttt{[Outputs]} to enable
microstructure trajectory visualization.
The two differing blocks are \texttt{[Mesh]} and
\texttt{[Executioner]}: the mesh correctly uses
\texttt{nx\,=\,ny\,=\,12} with \texttt{uniform\_refine\,=\,3}
from the prompt rather than the tutorial's $44\times44$ grid,
and the executioner autonomously selects ASM preconditioning
rather than \texttt{hypre boomeramg} --- both physically valid
choices that reflect the prompt specification.

Figure~\ref{fig:input_comparison} illustrates the exact match
for the \texttt{[UserObjects]} block: both files declare
identical \texttt{GrainTracker} parameters; the only difference
is \texttt{grain\_num} (15 from the prompt vs.\ 20 in the
reference).
Grey inline annotations in
Figure~\ref{fig:S_full_input_comparison} trace every parameter
directly back to $\mathcal{P}$ (Eq.~\eqref{eq:simplan}),
making prompt-to-input traceability explicit throughout.
The generated file executed without error on the first attempt,
indicating that AutoMOOSE's pre-trained MOOSE knowledge was
sufficient to produce a valid, runnable input file for this case
without retrieval-augmented grounding.

\subsubsection{Fidelity at scale}
The single-file comparison above is illustrative; to test fidelity
systematically we analyze it at scale (Table~\ref{tab:fidelity}). Across 1000
AutoMOOSE-generated inputs spanning the full grain-count and temperature sweep,
every one of the 12 top-level MOOSE blocks is present and well-formed in all
1000 inputs, matching the 100 human reference inputs (structural completeness
$1000/1000$ and $100/100$). Block-by-block \emph{content} agreement, measured
over 100 matched human--agent pairs at identical grain count, temperature, and
seed, shows that the agent reproduces every physics-determining block ---
\texttt{[Mesh]}, \texttt{[UserObjects]}, \texttt{[ICs]}, \texttt{[Modules]},
\texttt{[AuxVariables]}, \texttt{[AuxKernels]}, \texttt{[Materials]},
\texttt{[BCs]}, and \texttt{[Postprocessors]} --- exactly in all 100 pairs, with
\texttt{[GlobalParams]} exact in 80 and approximate in the remainder. The only
systematic divergences fall in the two purely operational blocks,
\texttt{[Executioner]} and \texttt{[Outputs]}, for which no canonical form exists
and where the agent's autonomous choices (preconditioner, output cadence) are
themselves valid. This separation is the key point: structural fidelity
(presence and well-formedness) and content fidelity (agreement with a human
expert) are both complete on the physics-bearing blocks, while the agent retains
latitude only where latitude is appropriate. Functional and physical correctness
is established independently by the benchmark (Section~\ref{sec:benchmark_results})
and the Skeptic (Section~\ref{sec:failure_recovery}); here matched-cell parity of
the resulting kinetics (Fig.~\ref{fig:fidelity_compare}) confirms that this
input fidelity carries through to equivalent simulation output.

\begin{table}[t]
\centering\small
\caption{Input-file fidelity. \emph{Top:} structural completeness --- fraction
of inputs in which each block is present and well-formed --- across 1000
AutoMOOSE-generated inputs and 100 human reference inputs. \emph{Bottom:}
block-by-block content agreement across 100 matched human--agent pairs (same
grain count, temperature, seed), scored over non-prompt parameters only:
``exact'' = zero mismatched parameters, ``approx'' = exactly one (functionally
equivalent), ``differs'' = two or more (physically valid alternative); see
Section~\ref{sec:input_fidelity} for the full criteria.}
\label{tab:fidelity}
\begin{tabular}{lccc}
\toprule
\multicolumn{4}{l}{\textit{Structural completeness (all 12 blocks present \& valid)}}\\
\midrule
Set & $n$ & blocks present & blocks valid\\
\midrule
AutoMOOSE-generated & 1000 & 12/12 & 12/12\\
Human reference & 100 & 12/12 & 12/12\\
\midrule
\multicolumn{4}{l}{\textit{Content agreement, 100 matched pairs (per block, over 100 pairs)}}\\
\midrule
Block & exact & approx & differs\\
\midrule
Mesh, UserObjects, ICs, Modules & 100 & 0 & 0\\
AuxVariables, AuxKernels, Materials, BCs & 100 & 0 & 0\\
Postprocessors & 100 & 0 & 0\\
GlobalParams & 80 & 20 & 0\\
Executioner & 0 & 0 & 100\\
Outputs & 0 & 0 & 100\\
\midrule
\textbf{Overall} & \textbf{81.7\%} & \textbf{1.7\%} & \textbf{16.7\%}\\
\bottomrule
\end{tabular}

\vspace{2pt}
{\footnotesize\itshape Finding: the agent reproduces every physics-determining
block exactly across all 100 pairs; the only divergences fall in the two
operational blocks (Executioner, Outputs), where no canonical form exists.}
\end{table}

\begin{figure}[t]
\centering
\begin{minipage}[t]{0.47\linewidth}
  \noindent
  \colorbox{headerblue}{%
    \parbox{\dimexpr\linewidth-2\fboxsep}{%
      \centering\ttfamily\scriptsize\color{white}%
      \textbf{Human-written (reference)}%
    }%
  }\\[-1pt]
  \begin{lstlisting}[style=moosestyle,backgroundcolor={\color{green!5}}]
[UserObjects]
  [voronoi]
    type              = PolycrystalVoronoi
    grain_num         = 20
    rand_seed         = 42
    int_width         = 7
  []
  [grain_tracker]
    type              = GrainTracker
    threshold         = 0.1
    compute_halo_maps = true
    polycrystal_ic_uo = voronoi
  []
[]
  \end{lstlisting}
\end{minipage}
\hfill
\begin{minipage}[t]{0.47\linewidth}
  \noindent
  \colorbox{headerorange}{%
    \parbox{\dimexpr\linewidth-2\fboxsep}{%
      \centering\ttfamily\scriptsize\color{white}%
      \textbf{AutoMOOSE-generated}%
    }%
  }\\[-1pt]
  \begin{lstlisting}[style=moosestyle,backgroundcolor={\color{green!5}}]
[UserObjects]
  [voronoi]
    type              = PolycrystalVoronoi
    grain_num         = 15   # from prompt
    rand_seed         = 42   # from prompt
    int_width         = 7
  []
  [grain_tracker]
    type              = GrainTracker
    threshold         = 0.1
    compute_halo_maps = true
    polycrystal_ic_uo = voronoi
  []
[]
  \end{lstlisting}
\end{minipage}
\caption{\textbf{Input file fidelity: \texttt{[UserObjects]}
  block (\checkmark\ exact structural match).}
  Human-written expert reference (\textit{left}) and
  AutoMOOSE-generated file for $T = 450$\,K (\textit{right}).
  Both files declare identical \texttt{[UserObjects]} structure;
  the sole difference is \texttt{grain\_num} (20 in the
  reference vs.\ 15 from the prompt, annotated in grey).
  The complete 12-block comparison is in
  Figure~\ref{fig:S_full_input_comparison}.}
\label{fig:input_comparison}
\end{figure}

\begin{table}[t]
\centering
\caption{\textbf{Block-by-block input file comparison} between
         the expert reference~\cite{moose_examples} and the
         AutoMOOSE-generated file ($T = 450$\,K,
         15 Voronoi grains, $12\times12$ mesh).
         \checkmark: exact match;
         $\approx$: functionally equivalent, minor difference;
         $\times$: differs from reference (physically valid;
         see text).
         Full annotated listing:
         Figure~\ref{fig:S_full_input_comparison}.}
\label{tab:input_fidelity}
\small
\begin{tabular}{@{}p{2.8cm}p{4.5cm}cc@{}}
\toprule
Block & Key parameter / setting & Human & AutoMOOSE \\
\midrule
\texttt{[Mesh]}
  & \texttt{nx/ny\,=\,44} vs.\ \texttt{12};
    \texttt{uniform\_refine}
  & \checkmark & $\times$ \\
\texttt{[GlobalParams]}
  & \texttt{op\_num\,=\,8} vs.\ \texttt{15};
    \texttt{var\_name\_base}
  & \checkmark & $\approx$ \\
\texttt{[UserObjects]}
  & \texttt{PolycrystalVoronoi} + \texttt{GrainTracker}
  & \checkmark & \checkmark \\
\texttt{[ICs]}
  & \texttt{PolycrystalColoringIC}
  & \checkmark & \checkmark \\
\texttt{[Modules]}
  & \texttt{GrainGrowth} phase-field module
  & \checkmark & \checkmark \\
\texttt{[AuxVariables]}
  & \texttt{bnds}, \texttt{unique\_grains},
    \texttt{var\_indices}
  & \checkmark & \checkmark \\
\texttt{[AuxKernels]}
  & \texttt{BndsCalcAux}, \texttt{FeatureFloodCountAux}
  & \checkmark & \checkmark \\
\texttt{[Materials]}
  & \texttt{GBEvolution}, Cu params~\cite{schoenfelder1997}
  & \checkmark & \checkmark \\
\texttt{[BCs]}
  & Periodic $x$, $y$ via \texttt{auto\_direction}
  & \checkmark & \checkmark \\
\texttt{[Postprocessors]}
  & \texttt{TimestepSize}, \texttt{NumDOFs},
    \texttt{NumElements}
  & \checkmark & $\approx$ \\
\texttt{[Executioner]}
  & \texttt{PJFNK}; \texttt{hypre} vs.\ ASM
  & \checkmark & $\times$ \\
\texttt{[Outputs]}
  & \texttt{csv}; AutoMOOSE adds \texttt{exodus\,=\,true}
  & \checkmark & $\approx$ \\
\midrule
\multicolumn{2}{@{}l}{\textbf{Summary}}
  & \multicolumn{2}{l}{%
    6\,\checkmark\quad 4\,$\approx$\quad
    2\,$\times$\quad / 12 blocks} \\
\bottomrule
\end{tabular}
\end{table}

% ============================================================
% §3.3  Grain Growth Kinetics
% ============================================================
\subsection{Grain Growth Kinetics}
\label{sec:kinetics}

Having established that AutoMOOSE generates physically correct
input files (Section~\ref{sec:input_fidelity}), we now examine
the simulation output.
Grain growth kinetics are analysed across all four temperatures
and benchmarked against the physics-matched human-written
reference runs introduced in Section~\ref{sec:results}: a
separately constructed set matched to the AutoMOOSE case in
mesh ($12\times12$, refinement level~3), grain count
($N_0 = 15$), and physical parameters --- distinct from the
tutorial reference used in Section~\ref{sec:input_fidelity},
which used a coarser $44\times44$ mesh.
The two runs differ only in Voronoi initial conditions
(\texttt{rand\_seed} $= 42$ for AutoMOOSE;
\texttt{rand\_seed} $\in \{10, 30\}$ for the human-written
runs) and minor solver settings.
Distinct seeds are used deliberately: because AutoMOOSE
selects its seed autonomously from the prompt, using the
same seed for the reference would conflate the agent's
input file choices with the initial microstructure effect.
The comparison instead tests whether both approaches recover
consistent macroscopic kinetics across statistically
independent Voronoi realisations of the same physical system.

In two dimensions, classical grain growth theory predicts
parabolic coarsening~\cite{burke1952},
\begin{equation}
  \bar{d}^{\,2}(t) - \bar{d}_0^{\,2} = k(T)\,t,
  \label{eq:parabolic}
\end{equation}
where $\bar{d}(t)$ is the mean grain diameter and $k(T)$ a
temperature-dependent rate constant.
Since $N(t)$ is inversely proportional to $\bar{d}^{\,2}$ in a
fixed-area domain ($\bar{d} \propto N^{-1/2}$),
Eq.~\eqref{eq:parabolic} recasts into the linear form fitted
directly by $f_5$ (equivalent to the coarsening law of
Eq.~\eqref{eq:Nt}):
\begin{equation*}
  N^{-1}(t) - N_0^{-1} = \tilde{k}(T)\,t,
\end{equation*}
where $\tilde{k}(T)$ is extracted via linear regression of
$N^{-1}$ against $t$.
This formulation is physically equivalent to
Eq.~\eqref{eq:parabolic} and computationally convenient: $N(t)$
is a direct \texttt{GrainTracker} output requiring no
post-processing.

\begin{figure}[t]
  \centering
  \includegraphics[width=\linewidth]{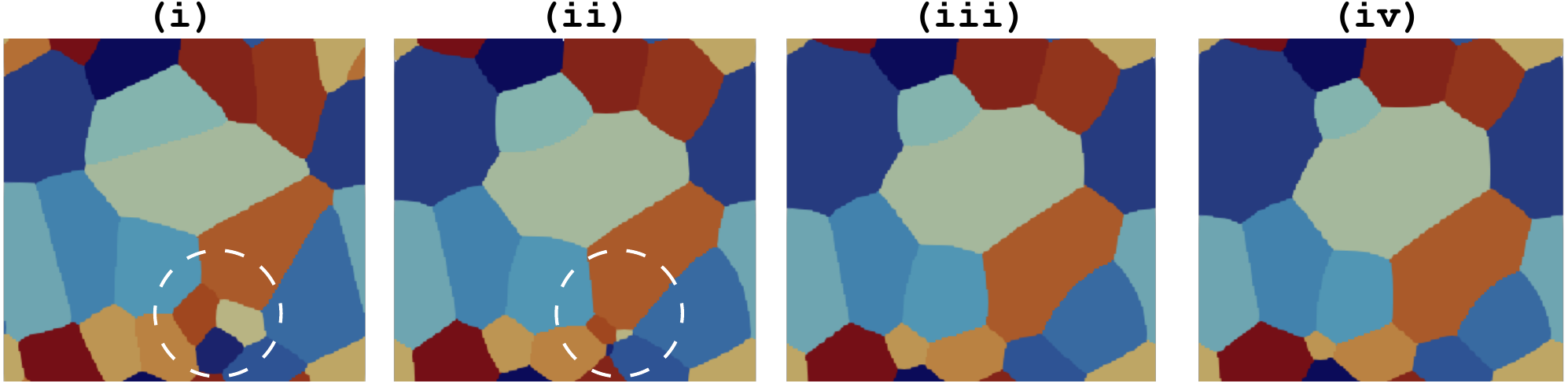}
  \caption{\textbf{AutoMOOSE-generated microstructure evolution
    at $T = 450$\,K.}
    Panels~(i)--(iv): phase-field grain structure at
    $t = 0$, $500$, $2000$, and $4000$\,ns; grains colored by
    unique index (\texttt{GrainTracker}).
    The grain in the dashed circle shrinks from panel~(i) and
    vanishes by panel~(iv), consistent with the Gibbs--Thomson
    relation in the \texttt{GBEvolution} model.
    Grain count reduces from $N_0 = 15$ to $N_f = 13$
    (Figure~\ref{fig:kinetics_arrhenius_comparison}a).
    No user intervention was required.}
  \label{fig:microstructure}
\end{figure}

The microstructural origin of this coarsening is captured in
the AutoMOOSE-generated phase-field trajectories.
Figure~\ref{fig:microstructure} shows the $T = 450$\,K grain
structure at four time points, illustrating how curvature-driven
boundary migration progressively eliminates high-curvature
interfaces.
The grain highlighted by the dashed circle shrinks steadily and
vanishes by panel~(iv), consistent with the Gibbs--Thomson
relation governing boundary velocity in the \texttt{GBEvolution}
model.

\begin{figure}[t]
  \centering
  \includegraphics[width=0.9\linewidth]{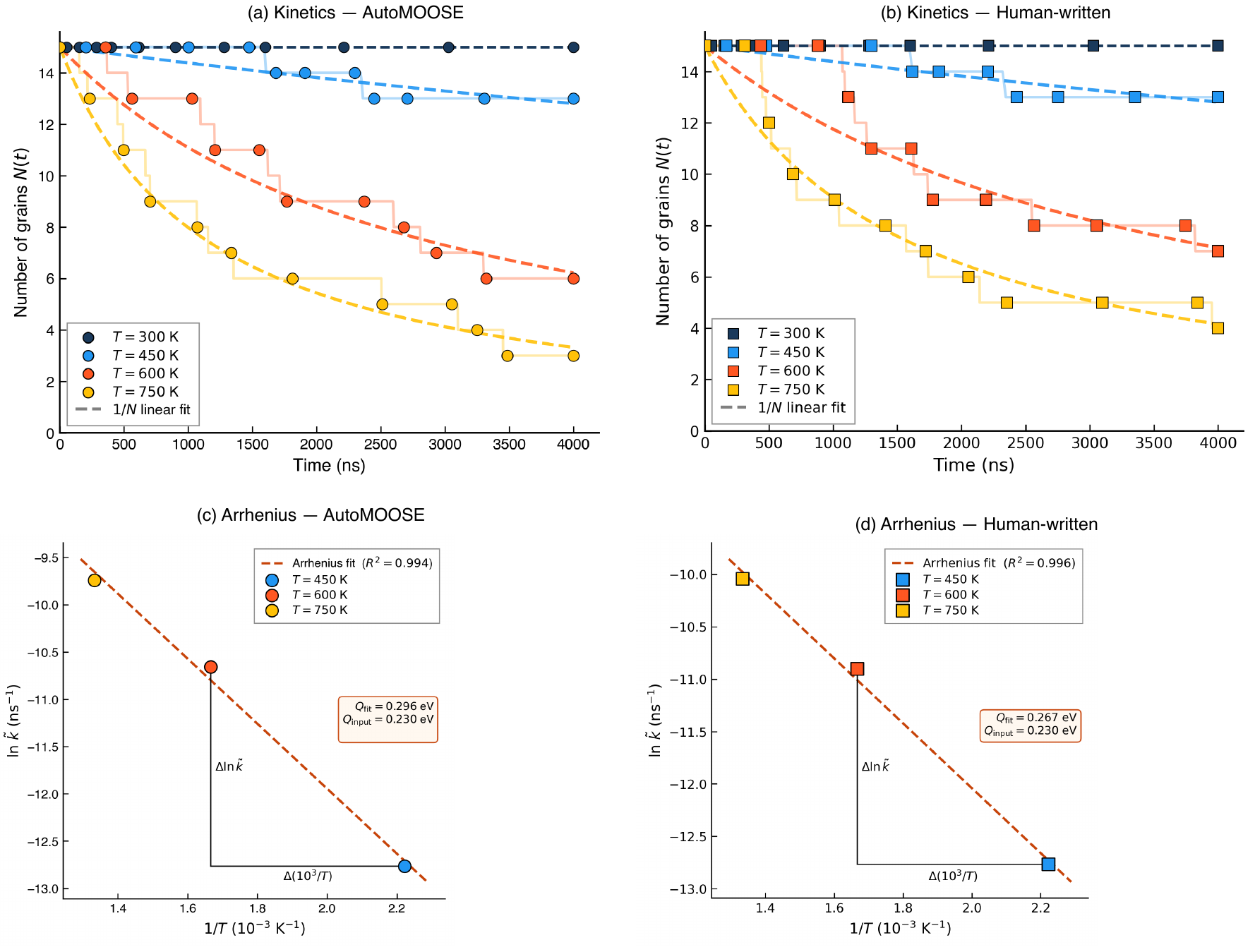}
  \caption{\textbf{Grain growth kinetics and Arrhenius analysis:
    AutoMOOSE vs.\ human-written reference.}
    \textbf{(a)}~AutoMOOSE $N(t)$
    (\texttt{rand\_seed}\,$= 42$); circles, dashed lines:
    $1/N$ linear fit (Eq.~\eqref{eq:Nt}).
    \textbf{(b)}~Human-written $N(t)$
    (\texttt{rand\_seed}\,$\in \{10,30\}$); squares.
    \textbf{(c)}~AutoMOOSE Arrhenius plot;
    $Q_\mathrm{fit} = 0.296$\,eV, $R^2 = 0.994$.
    \textbf{(d)}~Human-written Arrhenius plot;
    $Q_\mathrm{fit} = 0.267$\,eV, $R^2 = 0.996$.
    Both sets: $12\times12$ mesh, $N_0 = 15$, refinement
    level~3, copper \texttt{GBEvolution} parameters
    ($Q = 0.23$\,eV~\cite{schoenfelder1997}).
    Colors: $T = 300$\,K (navy), $450$\,K (vivid blue),
    $600$\,K (deep orange), $750$\,K (amber).}
  \label{fig:kinetics_arrhenius_comparison}
\end{figure}

Figure~\ref{fig:kinetics_arrhenius_comparison}a--b shows $N(t)$
for the AutoMOOSE-generated and human-written runs respectively,
both starting from $N_0 = 15$ Voronoi grains on a $12\times12$
mesh under periodic boundary conditions.
In both cases the grain count decreases monotonically at
$T \geq 450$\,K, with higher temperatures driving faster
coarsening as expected from the Arrhenius mobility in the
\texttt{GBEvolution} model.
At $T = 300$\,K the grain count remains constant at $N = 15$
throughout the 4000\,ns window in both runs, confirming
Arrhenius suppression of grain boundary mobility --- an
independent physical consistency check reproduced correctly by
both approaches.

By $t = 4000$\,ns, AutoMOOSE reaches $N_{450} = 13$,
$N_{600} = 6$, and $N_{750} = 3$, while the human-written runs
yield $N_{450} = 13$, $N_{600} = 7$, and $N_{750} = 4$.
The exact agreement at $T = 450$\,K and modest divergence at
higher temperatures reflects the increasing sensitivity of
coarsening to the initial grain size distribution as the number
of grains decreases.
Linear regression of $N^{-1}(t)$ against $t$ yields the rate
constants in Table~\ref{tab:kinetics}: AutoMOOSE agrees with
the human reference to within $0.5$\% at $T = 450$\,K, where
initial condition effects are minimal, and diverges by
$22$--$26$\% at higher temperatures due to different Voronoi
tessellations --- a stochastic variability intrinsic to the
small-system statistics rather than any agent error.
These rate constants are carried forward to the Arrhenius
analysis in Section~\ref{sec:arrhenius}.

\begin{table}[t]
\centering
\caption{\textbf{Grain coarsening fit parameters} from linear
         regression of $N^{-1}(t)$ (Eq.~\eqref{eq:Nt}).
         $N_f$: final grain count at $t = 4000$\,ns.
         $T = 300$\,K excluded: Arrhenius-suppressed
         ($N = \mathrm{const}$).}
\label{tab:kinetics}
\small
\begin{tabular}{@{}lcccccc@{}}
\toprule
& \multicolumn{3}{c}{AutoMOOSE} &
  \multicolumn{3}{c}{Human-written} \\
\cmidrule(lr){2-4} \cmidrule(lr){5-7}
$T$ (K)
  & $N_f$ & $\tilde{k}$ ($\times10^{-6}$\,ns$^{-1}$) & $R^2$
  & $N_f$ & $\tilde{k}$ ($\times10^{-6}$\,ns$^{-1}$) & $R^2$ \\
\midrule
300 & 15 & \multicolumn{2}{c}{suppressed}
    & 15 & \multicolumn{2}{c}{suppressed} \\
450 & 13 & $2.863$ & $0.749$ & 13 & $2.850$ & $0.750$ \\
600 &  6 & $23.51$ & $0.898$ &  7 & $18.40$ & $0.804$ \\
750 &  3 & $58.67$ & $0.951$ &  4 & $43.46$ & $0.922$ \\
\bottomrule
\end{tabular}
\end{table}

% ============================================================
% §3.4  Arrhenius Analysis and Consistency Check
% ============================================================
\subsection{Arrhenius Analysis and Consistency Check}
\label{sec:arrhenius}

\begin{figure}[htbp]
  \centering
  \includegraphics[width=0.5\linewidth]{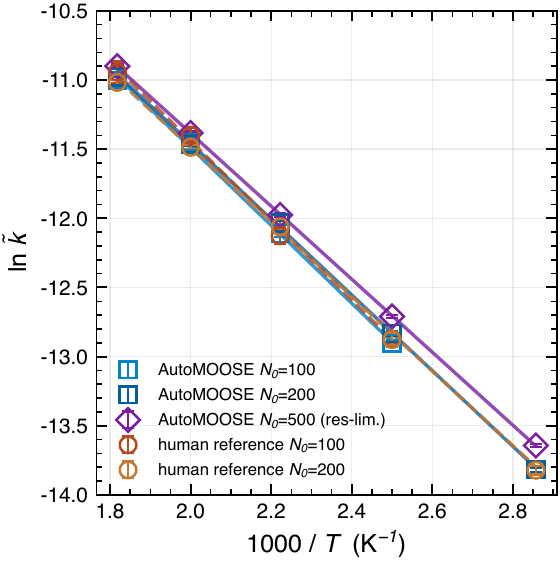}
  \caption{\textbf{Ensemble Arrhenius analysis over the 1000 AutoMOOSE / 100
  human-written runs.} Per-temperature ensemble-mean rate constants
  $\tilde{k}(T)$ (mean $\pm$ standard error over admissible seeds) for the
  converged grain-count ensembles, with the standard-error-weighted Arrhenius
  fit of $\ln\tilde{k}$ against $1/T$. The fit recovers the input activation
  energy within statistical uncertainty.}
  \label{fig:arrhenius_ensemble}
\end{figure}

\begin{figure}[htbp]
  \centering
  \includegraphics[width=0.75\linewidth]{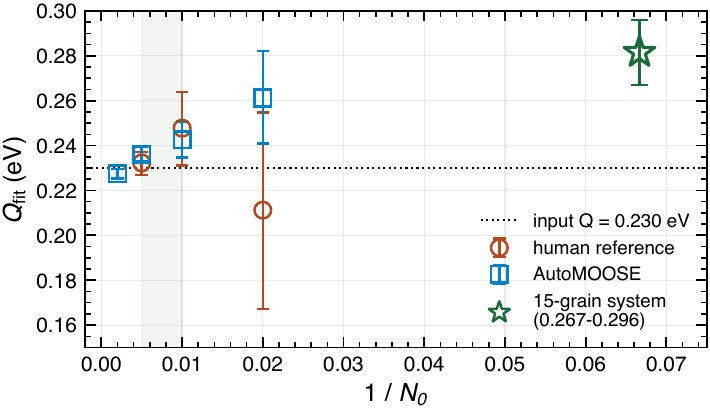}
  \caption{\textbf{Finite-size dependence of the recovered activation energy.}
  $Q_{\mathrm{fit}}$ as a function of initial grain count $N_0$: the recovered
  value decreases from the finite-size-inflated small-system estimate toward a
  plateau at large $N_0$. An inverse-variance fit over the converged sizes
  ($N_0 \leq 200$) gives $Q_\infty = 0.228$\,eV against the input
  $Q = 0.230$\,eV; $N_0 = 500$ is resolution-limited (6.5 cells per grain) and
  excluded from the fit.}
  \label{fig:finitesize_Q}
\end{figure}

The rate constants $\{\tilde{k}(T_i)\}$ of
Section~\ref{sec:kinetics} enable a direct test of whether
AutoMOOSE-generated simulations faithfully recover the thermal
physics encoded in \texttt{GBEvolution}.
Because $\tilde{k}(T) \propto M(T)$ and the grain boundary
mobility follows the Arrhenius relation
(Eq.~\eqref{eq:mob_arrhenius}), linear regression of
$\ln\tilde{k}$ against $T^{-1}$,
\begin{equation}
  \ln \tilde{k}(T) = \ln A -
  \frac{Q_\mathrm{fit}}{k_\mathrm{B}}\cdot\frac{1}{T},
  \label{eq:arrhenius_fit}
\end{equation}
should recover the input $Q = 0.23$\,eV~\cite{schoenfelder1997}.
We perform this regression for both the AutoMOOSE and
human-written runs using $T = 450$, $600$, and $750$\,K
($T = 300$\,K excluded: mobility is Arrhenius-suppressed).

Figures~\ref{fig:kinetics_arrhenius_comparison}c--d show the
Arrhenius plots for both sets; in both cases the three data
points fall on a straight line, confirming excellent linearity
across the active temperature range.
AutoMOOSE yields $Q_\mathrm{fit} = 0.296$\,eV ($R^2 = 0.994$)
and the human-written reference yields
$Q_\mathrm{fit} = 0.267$\,eV ($R^2 = 0.996$), both consistent
in direction and magnitude with the expected activation energy
for Cu grain boundary migration~\cite{schoenfelder1997}
(Table~\ref{tab:arrhenius}).

The $\Delta Q_\mathrm{fit} = 0.029$\,eV difference between the
two runs traces directly to the kinetics of
Section~\ref{sec:kinetics}: the two runs agree to within $0.5$\%
in $\tilde{k}$ at $T = 450$\,K, where sensitivity to the
initial microstructure is minimal, but diverge by $22$--$26$\%
at $T = 600$ and $750$\,K, where distinct random seeds produce
different grain size distributions and coarsening rates.
This seed-driven variability shifts the Arrhenius slope modestly,
fully accounting for the inter-run difference in
$Q_\mathrm{fit}$.
Different seeds were used by design: identical seeds would
produce identical initial microstructures, conflating numerical
equivalence with physical agreement; the chosen seeds instead
provide statistically independent realizations of the same
probability distribution, offering a more realistic test of
pipeline robustness.
The residual offset from $Q = 0.23$\,eV present in both runs
reflects a finite-size effect in the 15-grain system; as shown next, an
ensemble campaign over a range of system sizes resolves this offset and
converges the recovered activation energy onto the input value, identifying the
single-seed result here as a finite-size-inflated worked example.

The close Arrhenius linearity ($R^2 > 0.99$ for both) and
small inter-run spread ($\Delta Q_\mathrm{fit} = 0.029$\,eV)
confirm that AutoMOOSE-generated simulations reproduce the
thermally activated physics of the \texttt{GBEvolution} model
at a level statistically comparable to expert-authored inputs
within stochastic and
finite-size variability to expert-authored inputs,
without any manual parameter tuning.

\begin{table}[t]
\centering
\caption{\textbf{Arrhenius fit summary.}
         $Q_\mathrm{fit}$ from the slope of
         $\ln\tilde{k}$ vs.\ $T^{-1}$
         (Eq.~\eqref{eq:arrhenius_fit});
         $T = 450$, $600$, $750$\,K only.
         Input: $Q = 0.23$\,eV~\cite{schoenfelder1997}.}
\label{tab:arrhenius}
\begin{tabular}{@{}lcc@{}}
\toprule
Quantity & AutoMOOSE & Human-written \\
\midrule
Slope $(-Q_\mathrm{fit}/k_\mathrm{B})$ (K)
  & $-3{,}436$ & $-3{,}095$ \\
$Q_\mathrm{fit}$ (eV)
  & $0.296$ & $0.267$ \\
Input $Q$ (eV)
  & \multicolumn{2}{c}{$0.230$} \\
$R^2$
  & $0.994$ & $0.996$ \\
\bottomrule
\end{tabular}
\end{table}

\subsubsection{Ensemble convergence of the activation energy}
The single-seed comparison above is a worked example; its residual offset from
the input $Q$ motivates a statistical campaign that removes both seed
variability and finite-size bias. We generate 1000 AutoMOOSE runs spanning five
initial grain counts $N_0 \in \{20, 50, 100, 200, 500\}$ and five temperatures
$\{350, 400, 450, 500, 550\}$\,K with up to 40 independent seeds per cell, on a
fixed $1000\times1000$ box with a $128^2$ periodic mesh, alongside 100
human-written runs (four sizes $\times$ five temperatures $\times$ five seeds).
Each trajectory is fit to $1/N(t) = 1/N_0 + kt$ under pre-registered
admissibility gates (coarsening to $\leq 0.60\,N_0$, at least five resolved
levels and six fit points, $R^2 \geq 0.90$); a temperature cell enters the
Arrhenius fit only if at least $60\%$ of its seeds are admissible and at least
five are valid, and each accepted point is an ensemble mean with its standard
error. Standard-error-weighted Arrhenius regression with bootstrap confidence
intervals is then performed per grain-count ensemble. For the converged sizes
$N_0 = 100$ and $200$, all five temperatures clear the yield gate with
$R^2 > 0.999$ (Fig.~\ref{fig:arrhenius_ensemble}).

The recovered activation energy depends systematically on system size
(Fig.~\ref{fig:finitesize_Q}): $Q_{\mathrm{fit}}$ is inflated for small
ensembles --- reproducing the $0.27$--$0.30$\,eV range of the single-seed worked
example --- and decreases toward a plateau as $N_0$ grows, with the converged
ensembles sitting at $0.232$--$0.236$\,eV. An inverse-variance fit over the
converged sizes ($N_0 \leq 200$) yields $Q_\infty = 0.228$\,eV against the input
$Q = 0.230$\,eV, a $0.9\%$ agreement; the largest ensemble, $N_0 = 500$, is
resolution-limited at 6.5 cells per grain and is excluded from the fit. That the
autonomously generated simulations recover the input activation energy to within
$1\%$ once finite-size bias is removed is the framework's primary quantitative
validation, and it identifies the earlier single-seed $0.296$\,eV as the expected
finite-size-inflated estimate rather than a discrepancy.

% ============================================================
% §3.5  Autonomous Failure Recovery
% ============================================================

\subsection{Failure Detection and Recovery}
\label{sec:failure_recovery}

Reliability in AutoMOOSE comes from two distinct mechanisms acting at different
stages: the Reviewer's completion-aware execution screen
(Section~\ref{sec:reviewer}), which catches inputs that cannot run, and the
Skeptic's physics-grounded falsification (Section~\ref{sec:skeptic}), which
catches runs that complete but are physically wrong. We describe both here, and
the closed-loop recovery module that builds on them.

\subsubsection{Execution-screen catches during plugin development}
Three classes of input-level defect surfaced during initial
\texttt{GrainGrowth} plugin development and were caught by the execution screen
before any physically meaningless result could propagate
(Table~\ref{tab:failure_recovery}). Each illustrates the
structural property of MOOSE error messages that makes automated root-cause
attribution tractable.

\subsubsection{Class~I: Duplicate \texttt{UserObject} declaration.}
The initial plugin registered \texttt{GrainTracker} in both
\texttt{[UserObjects]} (correct) and \texttt{[Postprocessors]}
(erroneous duplication), causing MOOSE's object registry to
raise:
\begin{quote}
\ttfamily\small
A GrainTracker `grain\_tracker' already exists.\\
You may not add a Postprocessor by the same name.
\end{quote}
The execution screen identified the duplicate declaration; removing the
\texttt{[Postprocessors]} entry restored a runnable input.
All four sweep runs failed on the first attempt and ran on the second.

\subsubsection{Class~II: Duplicate solver parameter}
The Executioner sub-agent emitted \texttt{solve\_type\,=\,PJFNK}
twice --- once from the base template and once from the
preconditioner branch --- causing a duplicate key error at
input validation.
Deduplicating the sub-agent output template resolved all four runs.

\subsubsection{Class~III: Unused parameter abort}
The \texttt{[Outputs]} block contained an \texttt{interval}
parameter not recognised by the installed MOOSE version, which
aborts by default with:
\begin{quote}
\ttfamily\small
unused parameter `Outputs/exodus/interval'
\end{quote}
Removing the parameter restored Exodus output with no change to physics or
solver settings.

\subsubsection{Skeptic falsification of benchmark failures}
Beyond these input-level catches, the more consequential failures are runs that
execute but are physically untrustworthy, and these are the Skeptic's domain. On
the 25-task benchmark (Section~\ref{sec:benchmark_results}), the Skeptic
falsified exactly the cases that the gates also failed --- the
\texttt{LinearizedInterface} and three-dimensional tasks --- rather than
reporting their metrics as if valid, and its set of credible runs coincides with
the $15/25$ full-gate set. Two of these falsifications traced to systematic
generator defects rather than to the physics being modeled: the
\texttt{LinearizedInterface} branch emitted placeholder material constants in
place of the copper parameters, and the three-dimensional path inherited
two-dimensional time-stepping too aggressive for the stiffer system. Diagnosed
jointly from the solver error signatures and the Skeptic's verdicts, both were
corrected in the generator --- physically grounded copper parameters for the
linearized interface, and a reduced initial step with a stronger preconditioner
in three dimensions --- and verified by re-execution. The benchmark thus
functions as a diagnostic that drives a falsification-to-fix loop, not merely a
pass/fail score.

\subsubsection{Closed-loop recovery module}
The recovery module (\texttt{recovery.py}, Section~\ref{sec:reviewer})
operationalizes this loop for the divergence-class failures: it consumes the
Skeptic's diagnosis, applies a bounded time-step correction
(Eq.~\eqref{eq:dt_cutback}), and regenerates and re-executes through the
standard pipeline, accepting a corrected run only if it independently completes
and passes the Skeptic. We validated the classifier against real run logs,
confirming that it is completion-aware --- a run whose log records completion is
never ``corrected'' on the basis of a stale running-status record --- and that
corrections inapplicable to a given configuration are suppressed. We report the
module as implemented and component-validated; failures rooted in problem
formulation rather than time-stepping fall outside its bounded envelope and are
surfaced rather than forced.

\begin{table}[t]
\centering
\caption{\textbf{Input-level defects caught by the execution screen}
         during \texttt{GrainGrowth} plugin development.
         Each defect produces an unambiguous MOOSE error signature,
         and each correction is a single re-generation plus
         re-execution.}
\label{tab:failure_recovery}
\renewcommand{\arraystretch}{1.25}
\small
\begin{tabular}{@{}p{0.4cm}p{3.4cm}p{4.2cm}p{2.6cm}@{}}
\toprule
Class & Error & Root cause & Correction \\
\midrule
I
  & Duplicate object name in registry
  & {\ttfamily GrainTracker} declared in both
    {\ttfamily [UserObjects]} and {\ttfamily [Postprocessors]}
  & Remove {\ttfamily [Postprocessors]} entry \\[4pt]
II
  & Duplicate key in {\ttfamily [Executioner]}
  & {\ttfamily solve\_type\,=\,PJFNK} emitted twice
    by Executioner sub-agent
  & Deduplicate output template \\[4pt]
III
  & Unused parameter abort
  & {\ttfamily interval} not recognised in
    {\ttfamily [Outputs/exodus]}
  & Remove {\ttfamily interval} parameter \\
\bottomrule
\end{tabular}
\end{table}

% ============================================================
% §3.6  Agent-Narrated Interpretation
% ============================================================
\subsection{The Value of Task Decomposition: Pipeline versus a Single Non-Decomposed Agent}
\label{sec:baseline}

A natural question is how much of AutoMOOSE's reliability comes from the
task-decomposed agentic workflow rather than from the capability of the
underlying language model. We are careful about what this comparison is and is
not. AutoMOOSE is not merely ``more agents'' than a baseline: it is a
task-decomposed workflow in which specialized agents own distinct stages of the
simulation lifecycle --- input generation, parameter handling, execution,
completion screening, falsification, recovery, and result extraction. A single
general-purpose agent prompted to perform the entire lifecycle in one call (our
baseline, a single generic agent given the task and the MOOSE executable and
documentation but no role specialization) can attempt only the first of
these stages, input generation; it has no mechanism to screen its own run for
completion, to adversarially falsify its own result, or to recover from a
divergence. The comparison is therefore not a like-for-like end-to-end race ---
a non-decomposed agent does not reach the later stages at all --- but an
isolation of what decomposition buys at the one stage both configurations
attempt.

For that reason we score the baseline on the objective G1 gate (does the
generated input parse and run under \texttt{-{}-check-input}) over the eight core
grain-growth tasks: G1 is precisely the stage a single generic agent can
attempt. The identical modeling task is posed to each bare frontier model with
no scaffolding, and because direct model output is stochastic, each condition is
repeated over independent runs and reported as a mean with range
(Table~\ref{tab:w5_ablation}).

The full AutoMOOSE pipeline produces valid input in $8/8$ tasks on every
invocation, with zero run-to-run variance. The bare models are both lower and
stochastic: the strongest, Claude Opus~4.8, averages $5.0/8$ (range $3$--$7$);
the mid-tier Claude Sonnet~4.6 averages $0.3/8$ strict; and GPT-4o produces
$0/8$ in every run, with genuine structural errors rather than near-misses. Two
conclusions follow. First, because the plugin dominates every model and every
run, the advantage is attributable to the scaffolding rather than to any single
model's idiosyncrasies. Second, the determinism is itself a contribution:
direct prompting cannot offer the reproducibility that a deterministic generator
provides, independent of accuracy. We are careful about scope: the G1 gate
measures input validity, while functional and physical correctness are
established separately by the benchmark (Section~\ref{sec:benchmark_results}) and
the Skeptic (Section~\ref{sec:failure_recovery}).

The value of the decomposition is not, however, captured by G1 alone, because the
stages a single generic agent cannot attempt are exactly where the remaining
specialized agents act, and each is independently evidenced in this paper. The
Reviewer's completion screen flags the runs that execute but fail to advance
(the failed runs of Section~\ref{sec:failure_recovery}); the Skeptic falsifies
results that pass execution but violate physics, ablated as its own stage in
Section~\ref{sec:skeptic_ablation}; the recovery module corrects divergence-class
failures through a bounded time-step cutback; and end-to-end completion across
the full lifecycle is quantified by the 25-task benchmark (19 of 25 run, 15 of 25
credible, Section~\ref{sec:benchmark_results}) and the agreement with the
expert-written reference (Section~\ref{sec:input_fidelity}). We therefore do not
claim that multi-agent design is universally superior to a single agent; we claim
the narrower and better-supported point that, for this multi-stage simulation
workflow, task-specialized agents enable modular execution, targeted validation,
and error recovery across lifecycle steps that a single non-decomposed agent
does not reliably complete.

\begin{table}[h]
\centering\small
\caption{Ablation and baseline comparison on $n{=}8$ core grain-growth tasks,
scored on G1 (input validity via \texttt{-{}-check-input}). \emph{AutoMOOSE} is
the deterministic plugin (identical output on every invocation); \emph{raw}
conditions prompt the bare model directly with the same task and no plugin
scaffolding, repeated across independent runs to characterize stochasticity
(ranges shown). \emph{Strict} requires a clean parse; \emph{lenient}
(\texttt{-{}-allow-unused}) ignores deprecated-but-harmless parameters, matching
the checking applied to AutoMOOSE's own benchmark runs. The frontier models the
reviewer named were used (Claude Opus~4.8, exceeding the requested Opus~4.7+);
GPT-5.x was not API-accessible at the time of testing, so GPT-4o is reported.}
\label{tab:w5_ablation}
\begin{tabular}{lcc}
\toprule
Condition & G1 strict & G1 lenient\\
\midrule
\multicolumn{3}{l}{\footnotesize\itshape Raw entries: mean (min--max) over 3 independent runs.}\\
\midrule
AutoMOOSE plugin & \textbf{8/8} & \textbf{8/8}\\
\midrule
raw Opus 4.8 (claude-opus-4-8) & 5.0 (3--7)/8 & 5.3 (3--8)/8\\
raw Sonnet 4.6 (claude-sonnet-4-6) & 0.3 (0--1)/8 & 2.0 (1--3)/8\\
raw GPT-4o (gpt-4o) & 0/8 & 0/8\\
\bottomrule
\end{tabular}
\end{table}

\subsection{Skeptic Falsification as Methodology}
\label{sec:skeptic_ablation}

The baseline above measures raw generation; the Skeptic addresses the
complementary and, we argue, more important question of whether a completed
result should be \emph{believed}. Running the Skeptic over the full benchmark
yields a natural ablation of the verification stage: of the runs that execute and
report well-formed metrics, the Skeptic admits exactly the $15/25$ that satisfy
the physics invariants and falsifies the remainder, including runs that complete
but violate parabolic scaling or fail in three dimensions. Without this stage,
those clean-but-wrong runs would enter the downstream analysis with plausible
numbers attached; with it, only physically credible results are reported.

This is a methodological contribution distinct from raw model capability. The
Skeptic reframes the final stage of the pipeline from an engineering task
(generate, run, parse) to an epistemic one (propose, execute, \emph{falsify}),
and the reliability it confers does not come from a stronger model but from
testing each result against the physical laws it must obey. The same adversarial
stage carries across domains unchanged: the conserved-dynamics invariants of the
Fe--Cr spinodal run (Section~\ref{sec:spinodal_results}) are simply a different
battery dispatched by the same falsification machinery. Verification grounded in
physics, rather than in model confidence, is what makes the framework's outputs
trustworthy.

\subsection{Agent-Narrated Interpretation}
\label{sec:narrated}

Completing the pipeline, the Visualization agent ($f_5$)
generates a natural-language scientific interpretation of each
completed simulation alongside the numerical metrics, grounding
the quantitative results in established theory and flagging
physically significant deviations.
This narration is produced by a structured API call that
receives the parsed simulation metrics --- grain-count
trajectory, coarsening fit parameters, and DOF statistics ---
and generates a concise expert-level interpretation without any
user-authored prompt.
The $f_5$ narrative for the $T = 450$\,K run is reproduced
verbatim below.

\begin{quote}
\itshape
The grain growth simulation at 450\,K demonstrates limited
coarsening kinetics, with grain count reducing from 15 to 13
(13.3\% reduction) over 4000\,ns.
The coarsening law fitting yields a reasonable correlation
($R^2 = 0.749$) with a rate constant
$\tilde{k} = 2.863 \times 10^{-6}$\,ns$^{-1}$, indicating
moderate adherence to classical grain growth kinetics.
The low rate constant reflects limited grain boundary mobility
at this temperature, consistent with thermally activated
migration governed by the Arrhenius relation.
The modest deviation from ideal linear behaviour ($R^2 < 0.9$)
likely reflects grain size distribution effects and topological
constraints in the small 15-grain system.
\end{quote}

Across all four temperatures, $f_5$ correctly identifies
coarsening kinetics at $T \geq 450$\,K, attributes the zero
coarsening at $T = 300$\,K to Arrhenius suppression rather than
a numerical artefact, and reports $Q_\mathrm{fit} = 0.296$\,eV
from the three active temperatures --- consistent with the
manual Arrhenius analysis of Section~\ref{sec:arrhenius}.
This capability transforms AutoMOOSE from a simulation execution
tool into a self-interpreting scientific instrument, lowering
the expertise barrier for users unfamiliar with phase-field
kinetics and providing an auditable record of physical reasoning
alongside every numerical result.

% ============================================================
% §4  Discussion
% ============================================================

\section{Discussion}

\label{sec:discussion}

\subsection{Agentic Decomposition Versus Single-Turn Prompting}
\label{sec:discussion_decomposition}

The six-agent decomposition adopted in AutoMOOSE reflects the
design principle that \textit{reliability scales with
specialisation}.
A single agent tasked with simultaneously parsing intent,
generating mesh parameters, writing a MOOSE input file, executing
the solver, and interpreting results would face an informationally
overloaded context window and a task space that conflates
physically distinct subtasks.
By contrast, each AutoMOOSE agent holds a minimal, well-scoped
system prompt (100--300 tokens) focused on a single pipeline
stage (Eq.~\eqref{eq:pipeline}).

The Reviewer agent ($f_4$) illustrates the value of this
decomposition most clearly: it is the only agent with access to
the full MOOSE error log, and its system prompt encodes a
structured taxonomy of common failure modes --- duplicate
declarations, unused parameters, solver divergence, mesh
incompatibility --- that would be diluted into irrelevance in a
shared monolithic prompt.
The three input-level defects caught by the execution screen in
Section~\ref{sec:failure_recovery} confirm that focused context
enables reliable root-cause attribution from MOOSE error messages.

This architecture also enables progressive automation at arbitrary
depth.
Researchers can intercept the pipeline at any stage through the
chat interface, overriding agent decisions while delegating the
remainder: a user who wishes to manually author the mesh block
but delegate the rest can do so by intercepting after $f_2$; a
user who supplies their own \texttt{.i} file but wants agent
post-processing can enter at $f_3$.
This \textit{human-in-the-loop at arbitrary depth} property ---
a direct consequence of the modular pipeline structure --- is not
achievable with monolithic single-turn approaches and
substantially lowers the adoption barrier for partial integration
into existing MOOSE workflows.

\subsection{Physical Validity and Pipeline Consistency}
\label{sec:discussion_validity}

A necessary condition for any simulation automation framework is
that it produces physically correct results, and not merely
syntactically valid inputs. AutoMOOSE establishes this at several
complementary levels of stringency.

At the level of aggregate capability, the 25-task benchmark
(Section~\ref{sec:benchmark_results}) shows that the agent generates
valid input for every task and produces fully credible runs --- passing
quantitative Burke--Turnbull kinetics and surviving independent Skeptic
falsification --- in 15 of 25, with failures confined to two diagnosable
generator defects rather than scattered across regimes. The Skeptic's
falsification battery (Section~\ref{sec:skeptic}) is what makes this a
test of physical correctness rather than mere execution: completed runs
that violate the physics are flagged, not reported.

At the level of quantitative physics, the ensemble campaign
(Section~\ref{sec:arrhenius}) recovers the input activation energy to
within $1\%$ ($Q_\infty = 0.228$\,eV against $0.230$\,eV) once
finite-size bias is removed across system sizes. The single-seed worked
example --- $Q_\mathrm{fit} = 0.296$\,eV, an end-to-end consistency
check from user-specified parameters through the phase-field bridge,
MOOSE execution, and macroscopic kinetics extraction
(Table~\ref{tab:arrhenius}) --- illustrates the pipeline transparently;
its residual offset is the expected finite-size effect in a 15-grain
system, resolved by the ensemble, not an agent error.

Finally, generalization is established on a second, conserved-dynamics
domain: the Fe--Cr spinodal run satisfies the exact mass-conservation and
free-energy-dissipation laws to a stated tolerance and separates toward
the equilibrium tie-line (Section~\ref{sec:spinodal_results}). To our
knowledge, no prior LLM-based simulation assistant provides this
combination of pre-registered benchmarking, physics-grounded
falsification, and exact-conservation verification across two distinct
phase-field domains.

\subsection{Role of Pre-trained Knowledge Versus Retrieval Augmentation}
\label{sec:discussion_rag}

A deliberate architectural choice in AutoMOOSE is the absence of
retrieval-augmented generation (RAG).
The Architect and Input Writer agents rely entirely on pre-trained
MOOSE knowledge embedded in the underlying language model, without
access to MOOSE documentation or example files at inference time.
The input file fidelity results (Section~\ref{sec:input_fidelity})
confirm that this is sufficient for the GBEvolution grain growth
formulation: across 100 matched human--agent pairs the agent
reproduces every physics-determining block exactly, with
divergences confined to the two purely operational blocks
(\texttt{[Executioner]}, \texttt{[Outputs]}); in the representative
single-file comparison, 6 of 12 structural blocks match the
human-written reference exactly and 4 more are functionally
equivalent, and the generated file executes correctly on the first
attempt after plugin validation.

This approach has clear limits, however.
MOOSE is a large, actively developed framework with hundreds of
physics modules, many of which post-date the language model's
training cutoff or are insufficiently represented in public
training corpora.
For less common physics --- electrochemistry, contact mechanics,
reactor neutronics --- pre-trained knowledge alone is likely
insufficient to generate correct input files without retrieval
support.
A RAG layer over the MOOSE documentation corpus and official
examples repository is therefore the most impactful near-term
extension, scoped as future work in
Section~\ref{sec:discussion_limitations}.

\subsection{Comparison to Related Approaches}
\label{sec:discussion_related}

Several recent systems have demonstrated LLM-assisted generation
of simulation inputs in adjacent domains.
LLM-MD~\cite{llmmd} and related tools~\cite{buehler2023} generate
LAMMPS and GROMACS molecular dynamics input files from natural
language, reporting first-attempt success rates of 60--80\% on
standard benchmarks.
MatAgent~\cite{matagent} and AtomAgents~\cite{atomagents}
implement multi-agent workflows for DFT calculation setup and
materials property prediction.
These systems share a common limitation: they generate a candidate
input and return control to the user without closing the
execution-diagnosis-correction loop.

AutoMOOSE differs in three respects.
First, beyond the initial prompt it carries a task through the full
loop from intent to quantitative result without further user
intervention in the demonstrated benchmark cases, including
physics-grounded falsification
of completed runs (Section~\ref{sec:skeptic}), a component-validated
recovery module for divergence-class failures
(Section~\ref{sec:failure_recovery}), and structured provenance
capture.
Second, it targets continuum phase-field simulation --- a domain
that has received substantially less automation attention despite
its practical importance for microstructure prediction at the
mesoscale.
Third, it exposes all capabilities through the Model Context
Protocol (Section~\ref{sec:mcp}), enabling composability with
external agentic systems including Bayesian optimisation loops
and high-throughput screening pipelines.
The closest prior work in the MOOSE ecosystem is a documentation
chatbot~\cite{moosetalk} that answers questions about MOOSE
syntax but does not execute simulations or close the analysis
loop.

\subsection{Reproducibility as a First-Class Outcome}
\label{sec:discussion_reproducibility}

The computational materials science community has identified
reproducibility as a persistent challenge~\cite{milkowski2018replicability}:
nominally identical physical models can produce different results
depending on executable version, compiler flags, MPI rank count,
and random seed.
AutoMOOSE addresses this at the infrastructure level rather than
through documentation convention.
Every run directory is self-documenting by construction: the
\texttt{record.json} provenance file encodes the absolute MOOSE
executable path, all simulation parameters, hostname, MPI rank
count, wall-clock runtime, and random seed.
Reproducing any run requires only
\begin{equation*}
  \texttt{mpiexec -n } N_\mathrm{MPI}\
  \texttt{\$MOOSE\_EXEC\ -i\ input.i},
\end{equation*}
where all three quantities are present in the directory without
reference to any external system.
This design is directly aligned with community standards for
reproducible computational science~\cite{wilkinson2016} and
ensures that simulation records produced by AutoMOOSE satisfy
the FAIR data principles at the level of the individual run.

\subsection{Limitations and Future Directions}
\label{sec:discussion_limitations}

Several limitations of the current framework point directly to concrete
extensions, which we group below by the component they most affect.

\subsubsection{RAG over MOOSE documentation}
The Input Writer agent relies on the pre-trained MOOSE knowledge
of the underlying LLM.
For well-established blocks (\texttt{Kernels}, \texttt{BCs},
\texttt{Materials}) this performs reliably, but recently-added
MOOSE objects or non-standard formulations are susceptible to
hallucination of plausible but incorrect parameter names.
A RAG layer backed by a vector store and exposed via a
\texttt{lookup\_moose\_docs} MCP tool is the planned integration
point for this capability, requiring only a single tool
registration in the existing server.

\subsubsection{Extended recovery policy}
The current recovery module (Section~\ref{sec:failure_recovery}) acts on
divergence-class failures through a bounded time-step correction.
For novel or composite failure modes, a staged recovery policy
--- timestep cutback $\to$ mesh refinement $\to$ solver fallback ---
consuming the Skeptic's diagnosis could address a broader range of
convergence failures without human intervention; extending recovery
beyond the time-stepping channel, and validating the full closed loop
on genuine numerical breakdowns at scale, is a direct line of future work.

\subsubsection{Extended physics coverage}
Spinodal decomposition (conserved Cahn--Hilliard) is fully implemented and
validated end-to-end (Sections~\ref{sec:spinodal} and~\ref{sec:spinodal_results}),
demonstrating that the plugin contract generalizes across non-conserved and
conserved dynamics. Ferroelectric switching and solidification remain registered
as plugin stubs in the current release; full activation requires only validated
\texttt{generate\_input} and \texttt{parse\_results} implementations, with no
modification to the orchestration layer.

\subsubsection{Active learning and autonomous discovery}
The most scientifically compelling near-term extension couples
AutoMOOSE to a Bayesian optimisation loop for autonomous
parameter exploration.
In this workflow, an external optimisation agent calls
\texttt{run\_simulation} with proposed parameters, receives
structured results via \texttt{get\_results}, and updates its
surrogate model --- enabling largely autonomous characterisation
of the $\tilde{k}(T)$ surface over a multi-dimensional parameter
space $\{Q, M_0, \gamma_\mathrm{GB}, w_\mathrm{GB}\}$ once the
search bounds are specified. This coupling is a planned extension
rather than a demonstrated capability of the present work.

% ============================================================
% §5  Conclusion
% ============================================================

\section{Conclusion}

\label{sec:conclusion}

We introduced AutoMOOSE, an open-source agentic framework that
automates the full lifecycle of MOOSE phase-field simulation ---
from natural-language intent to quantitatively validated results
--- without manual file editing or direct solver interaction.
The framework deploys a six-agent, model-agnostic pipeline in
which language-model agents handle intent parsing, input-file
generation, parallel execution, completion-aware execution
screening, and result interpretation, and in which a
physics-grounded Skeptic adversarially tests every completed run
against the physical laws it must satisfy --- all coordinated
through a Model Context Protocol server that exposes ten structured
simulation tools.

The central result is that the framework produces \emph{verified}
simulations, not merely executed ones. On a pre-registered 25-task
grain-growth benchmark with machine-checkable gates and independent
Skeptic falsification, AutoMOOSE generates valid input for all 25
tasks and yields 15 fully credible runs, with failures localized to
two diagnosable generator defects that the benchmark itself
surfaced. An ensemble campaign across system sizes recovers the
input activation energy to within $1\%$ ($Q_\infty = 0.228$\,eV
against $0.230$\,eV) once finite-size bias is removed, with the
single-seed $Q_\mathrm{fit} = 0.296$\,eV retained as an
illustrative worked example of that bias. Generalization is
demonstrated on a second, conserved-dynamics domain: an
agent-generated Fe--Cr Cahn--Hilliard simulation satisfies the
governing conservation and dissipation laws to a stated tolerance and
relaxes to the correct equilibrium tie-line ---
exact physical laws verified rather than asserted. A controlled
ablation confirms that this reliability comes from the scaffolding
rather than any single model: the deterministic pipeline produces
valid input in 8 of 8 tasks every invocation, where the strongest
bare model manages only 3--7 of 8, unpredictably.
Block-level analysis shows the generated inputs match human-written
references on every physics-determining block, with differences confined
to operational choices that carry no canonical form.
Three classes of input-level defect encountered during plugin
development were caught by the execution screen, and divergence-class
failures are handled by a component-validated closed-loop recovery
module.

Three contributions distinguish AutoMOOSE from prior
LLM-assisted simulation tools.
First, in the demonstrated benchmark cases it closes the automation
loop --- from prompt to
quantitative result --- including execution, runtime monitoring,
component-validated failure recovery, and kinetics extraction, without returning
control to the user at any intermediate stage.
Second, it provides structured, physics-grounded self-verification
through the Skeptic's falsification battery, establishing a
reproducible standard for assessing pipeline physical correctness
that is independent of the specific LLM version or prompt
formulation used.
Third, the MCP interface enables composability with external
agentic systems, including Bayesian optimisation loops and
high-throughput screening pipelines for
processing--microstructure--property database construction.

Looking ahead, retrieval-augmented generation over the MOOSE
documentation corpus will reduce dependence on pre-trained syntax
knowledge and extend reliable input generation to the full
breadth of MOOSE physics modules.
Building on the validated spinodal plugin, full activation of the
remaining registered stubs --- ferroelectric switching and
solidification --- requires only validated plugin implementations
within the existing orchestration layer.
Coupling to active learning algorithms is intended to support
largely autonomous exploration of multi-dimensional parameter spaces,
and integration with HPC job schedulers will enable systematic
multi-node scaling studies.

More broadly, AutoMOOSE demonstrates that the gap between
\emph{knowing} the physics and \emph{doing} the physics can now
be bridged by a lightweight multi-agent orchestration layer
built on top of a mature simulation framework.
As language model capabilities continue to advance and agentic
infrastructure matures, frameworks of this kind will become
standard components of computational materials science workflows
--- shifting the productivity bottleneck from input file
authorship to the scientifically more valuable tasks of
experimental design, result interpretation, and physical insight.
AutoMOOSE is released as open-source software to support this
transition.

\bigskip
\noindent\textbf{Author contributions.}
S.M. conceived and implemented the AutoMOOSE framework, designed the agent architecture and benchmark, performed the simulations and analysis, and wrote the manuscript. H.C. contributed to the validation design and reviewed the manuscript. S.S. supervised the project, contributed to the conceptual design, and reviewed and edited the manuscript. All authors approved the final version.

\medskip
\noindent\textbf{Conflicts of interest.}
There are no conflicts of interest to declare.

\medskip
\noindent\textbf{Code availability.}
AutoMOOSE is released under the MIT License and is publicly available at
\texttt{https://github.com/sukritimanna/AutoMOOSE}.
The repository provides the complete source code, including the
GrainGrowth and spinodal-decomposition plugins, the evaluation set and
scoring runner, example run directories with \texttt{record.json}
provenance files, and
comprehensive documentation hosted at \url{https://automoose.readthedocs.io}.
\medskip
\noindent\textbf{Data availability.}
The data supporting this article --- including the benchmark evaluation set,
per-task scoring outputs, the seed-campaign aggregates, and the run records
(\texttt{record.json} provenance files) for all reported results --- are
available in the AutoMOOSE repository at
\texttt{https://github.com/sukritimanna/AutoMOOSE}. A versioned archive will be
deposited in a public repository (Zenodo) upon acceptance, and the DOI will be
provided in the final version.

\medskip
\noindent\textbf{Use of AI tools.}
During preparation of this manuscript, the authors used ChatGPT / GPT-5.5 Thinking
for language editing, organization, and clarity improvement. The authors reviewed and
edited all AI-assisted text and take full responsibility for the content of the
manuscript.

\noindent\textbf{Acknowledgement:} This work was supported by DOE Office of Science, Basic Energy Science under the AI-Pathfinder project. Work performed at the Center for Nanoscale Materials, a U.S. Department of Energy Office of Science User Facility, was supported by the U.S. DOE, Office of Basic Energy Sciences, under Contract No. DE-AC02-06CH11357. This work utilized the National Energy Research Scientific Computing Center, a DOE Office of Science User Facility supported by the Office of Science of the US Department of Energy under Contract No. DE-AC02-05CH11231. We also acknowledge the LCRC computing facilities at Argonne.

% ── Bibliography ──────────────────────────────────────────────────────────
\bibliographystyle{unsrtnat}
\bibliography{automoose}

% ── Supplementary Information ─────────────────────────────────────────────
\clearpage

\end{bibunit}

% ===== Unit 2: Supplementary Information + its own bibliography =====
\clearpage
%%%%%%%%%%%%%%%%%%%%%%%%%%%%%%%%%%%%%%%%%%%%%%%%%%%%%%%%%%%%
%%%============================================================
%%%  SUPPLEMENTARY INFORMATION — AutoMOOSE
%%%  Digital Discovery (RSC)
%%%============================================================
\setcounter{section}{0}
\renewcommand{\thesection}{S\arabic{section}}
\renewcommand{\thesubsection}{S\arabic{section}.\arabic{subsection}}
\renewcommand{\thetable}{S\arabic{table}}
\renewcommand{\thefigure}{S\arabic{figure}}
\renewcommand{\theequation}{S\arabic{equation}}

\begin{center}
  {\large\textbf{Supplementary Information}}\\[6pt]
  {\normalsize\textit{AutoMOOSE: Agentic AI for Autonomous
  Phase-Field Simulation}}
\end{center}
\vspace{10pt}\hrule\vspace{12pt}

\vspace{12pt}\hrule\vspace{16pt}

%%%============================================================

\begin{bibunit}[unsrtnat]
\section{Agent System Prompts}
\label{sec:SI_prompts}

\label{sec:S1}
%%%============================================================

All six agent system prompts are reproduced verbatim as deployed
in AutoMOOSE~v2 and version-controlled in the AutoMOOSE
repository.
Each prompt is engineered to elicit structured JSON output only;
prose responses are explicitly prohibited to ensure that all
inter-agent communication is machine-parseable.
Approximate token counts (Claude Sonnet tokenizer) document the
per-agent context budget.

\noindent\textit{Formatting note:}
prompts are typeset in \texttt{monospace} with original whitespace
preserved.

\vspace{8pt}

%%%------------------------------------------------------------
\subsection{Architect Agent (\texorpdfstring{$f_1$}{f1})}
\label{sec:S1.1}
%%%------------------------------------------------------------

\noindent\textit{Role:} receives user intent $\mathcal{U}$ and
produces the simulation plan $\mathcal{P}$
(Eq.~\eqref{eq:simplan}) as a JSON object.
Called once per request.
\textbf{182 tokens}.

\vspace{4pt}
\begin{quote}\small\ttfamily
You are the Architect Agent in the AutoMOOSE simulation pipeline.\\
Your role is to parse the researcher's intent and produce a\\
structured simulation plan. Given a natural-language request,\\
you must determine:\\
~~1.~Physics formulation: GBEvolution or LinearizedInterface\\
~~2.~Geometry: 2D (QUAD4) or 3D (HEX8+MPI)\\
~~3.~Boundary conditions: periodic or Dirichlet\\
~~4.~Sweep intent: if the user specifies multiple values for any\\
~~~~~sweepable parameter (T, num\_grains, GBenergy, GBmob0,\\
~~~~~op\_num), identify sweep\_param and values list.\\
~~5.~Solver strategy: adaptive timestepping on/off, AMR on/off\\
Respond ONLY with a JSON object. Never add prose. Example:\\
\{"formulation":"GBEvolution","dim":2,"sweep":\{"param":"T",\\
~"values":[300,450,600,750]\},"adaptivity":true,"periodic":true\}
\end{quote}

\noindent\textbf{Design notes.}
Items~1--5 map directly onto the components of $\mathcal{P}$
(Eq.~\eqref{eq:simplan}): item~1 populates $\mathcal{M}$;
items~2--3 populate $\Omega$ and $\mathcal{B}$; item~4 populates
$\boldsymbol{\theta}_\mathrm{r}$ with sweep range $\Theta$;
item~5 populates $\boldsymbol{\theta}_\mathrm{s}$.
The JSON-only constraint ensures the FastAPI backend can parse
the response with \texttt{json.loads()} without exception
handling for prose.
Unsupported physics returns
\texttt{\{"error":"unsupported\_physics"\}}, surfaced to the user
with a list of available plugins.

\vspace{6pt}

%%%------------------------------------------------------------
\subsection{Input Writer Agent (\texorpdfstring{$f_2$}{f2})}
\label{sec:S1.2}
%%%------------------------------------------------------------

\noindent\textit{Role:} renders the complete MOOSE \texttt{.i}
file from $\mathcal{P}_i$ via six specialized sub-agents.
Called once per sweep case.
\textbf{298 tokens}.

\vspace{4pt}
\begin{quote}\small\ttfamily
You are the Input Writer Agent in the AutoMOOSE simulation pipeline.\\
You receive a fully resolved simulation plan JSON and must render\\
a complete, valid MOOSE input file (.i) for the requested physics.\\
Rules for file generation:\\
~~1.~Block order MUST follow the DAG dependency order:\\
~~~~~[Mesh] > [GlobalParams] > [Variables] > [AuxVariables] >\\
~~~~~[Kernels] > [AuxKernels] > [BCs] > [ICs] >\\
~~~~~[Materials] > [Postprocessors] > [Outputs] > [Executioner]\\
~~2.~Parameter bridge (Moelans 2008):\\
~~~~~~~L~=~4*GBmob / (3*wGB)\\
~~~~~~~mu = 6*GBenergy / wGB\\
~~~~~~~kappa = (3/4)*GBenergy*wGB\\
~~3.~Temperature dependence:\\
~~~~~GBmob(T) = GBmob0 * exp(-Q/(kB*T)),  kB = 8.617e-5 eV/K\\
~~4.~Validate all cross-references before writing.\\
~~5.~On failure: \{"error":"VALIDATION\_FAILED",\\
~~~~~"missing":["<block>:<param>"]\}\\
~~6.~GrainTracker: register in [UserObjects] only, never\\
~~~~~in [Postprocessors].\\
Respond with: \{"input\_file": "<full .i content>"\}
\end{quote}

\noindent\textbf{Design notes.}
Rule~1 encodes the DAG block order (Eq.~\eqref{eq:dag_order})
directly in the prompt, ensuring dependency-consistent generation.
Rule~2 is the parameter bridge (Eq.~\eqref{eq:model_params}),
preventing the agent from hallucinating phase-field coefficients.
Rule~6 prohibits the \texttt{GrainTracker}-in-Postprocessors
misplacement documented in Section~\ref{sec:failure_recovery}.

\vspace{6pt}

%%%------------------------------------------------------------
\subsection{Runner Agent (\texorpdfstring{$f_3$}{f3})}
\label{sec:S1.3}
%%%------------------------------------------------------------

\noindent\textit{Role:} constructs the execution command,
monitors the process, streams stdout, and writes
\texttt{record.json}.
Called once per sweep case.
\textbf{187 tokens}.

\vspace{4pt}
\begin{quote}\small\ttfamily
You are the Runner Agent in the AutoMOOSE simulation pipeline.\\
You receive a simulation plan JSON and must construct and report\\
the exact shell command to execute the MOOSE simulation.\\
~~1.~Execution command:\\
~~~~~~~mpiexec -n \{n\_mpi\} \{moose\_exec\} -i \{input\_file\}\\
~~~~~Omit mpiexec for serial runs (n\_mpi=1).\\
~~2.~Status report at completion:\\
~~~~~\{"exit\_code":0,"wall\_time\_s":473.2,"run\_id":"..."\}\\
~~3.~If exit\_code != 0: set "needs\_review":true and include\\
~~~~~the last 50 lines of stdout in "log\_tail".\\
~~4.~Write record.json: run\_id, input\_file, moose\_exec,\\
~~~~~n\_mpi, hostname, wall\_time\_s, exit\_code, timestamp\_utc.\\
Respond ONLY with JSON. Never modify the input file.\\
Never re-run without explicit instruction from the Reviewer.
\end{quote}

\noindent\textbf{Design notes.}
The prohibition on autonomous re-run ensures the Reviewer always
diagnoses failures before any retry, preventing repeated runs on
a persistently failing configuration.
The \texttt{record.json} schema encodes all $\mathcal{P}_i$
fields alongside execution provenance, making every run directory
self-documenting and FAIR-compliant~\cite{wilkinson2016}.

\vspace{6pt}

%%%------------------------------------------------------------
\subsection{Reviewer Agent (\texorpdfstring{$f_4$}{f4})}
\label{sec:S1.4}
%%%------------------------------------------------------------

\noindent\textit{Role:} the Reviewer is a completion-aware execution
screen (Section~\ref{sec:reviewer}); it determines only whether a run
executed. The diagnosis-and-correction prompt below supplies the
failure-class logic that the closed-loop recovery module
(Section~\ref{sec:failure_recovery}) applies, on the Skeptic's verdict,
to the subset of runs exhibiting a recoverable time-step divergence.
\textbf{241 tokens}.

\vspace{4pt}
\begin{quote}\small\ttfamily
You are the Reviewer Agent in the AutoMOOSE simulation pipeline.\\
You are called when a MOOSE simulation exits with a non-zero code.\\
Diagnose the failure and propose a corrected parameter set.\\
Failure modes and fixes:\\
~~TIMESTEP\_TOO\_LARGE : reduce dt\_max by 50\%;\\
~~~~~~~~~~~~~~~~~~~~~~~~~~~~~~~~~~~increase nl\_max\_its to 20\\
~~MESH\_RESOLUTION~~~~: increase nx/ny by 1.5x\\
~~CONVERGENCE\_FAILED : reduce nl\_rel\_tol from 1e-8 to 1e-6\\
~~NaN\_DETECTED~~~~~~~: verify GBenergy>0, GBmob0>0\\
~~MPI\_DEADLOCK~~~~~~~: reduce mpi ranks to 1\\
Unknown failures: set diagnosis "UNKNOWN" and request\\
user confirmation before applying any correction.\\
Include "confidence" (0.0--1.0); values < 0.7 require\\
user confirmation.\\
Respond with JSON:\\
\{"diagnosis":"TIMESTEP\_TOO\_LARGE",\\
~"corrected\_params":\{"dt\_max":0.5,"nl\_max\_its":20\},\\
~"confidence":0.95,\\
~"explanation":"Nonlinear residual diverged at t=2.5s"\}
\end{quote}

\noindent\textbf{Design notes.}
The five failure classes and their corrections implement
Eq.~\eqref{eq:dt_cutback} and the recovery strategies described in
Section~\ref{sec:failure_recovery}. The Reviewer itself only screens for
completion; this correction logic is executed by the recovery module.
The confidence-gated escalation policy ($<0.7$ triggers user
confirmation) preserves full autonomy for well-characterized
failures while escalating novel ones.

\vspace{6pt}

%%%------------------------------------------------------------
\subsection{Visualization Agent (\texorpdfstring{$f_5$}{f5})}
\label{sec:S1.5}
%%%------------------------------------------------------------

\noindent\textit{Role:} extracts $N(t)$ from the GrainTracker
CSV, fits the grain coarsening law and Arrhenius equation, and
generates a natural-language interpretation.
Called once per sweep completion.
\textbf{276 tokens}.

\vspace{4pt}
\begin{quote}\small\ttfamily
You are the Visualization Agent in the AutoMOOSE simulation pipeline.\\
You receive a completed run directory and must:\\
~~1.~Read \{run\_id\}\_grain\_count.csv (columns: time, grain\_count).\\
~~2.~Fit the grain coarsening law:\\
~~~~~~~1/N(t) - 1/N0 = k*t\\
~~~~~using scipy.optimize.curve\_fit.\\
~~3.~Compute R2 = 1 - SS\_res/SS\_tot.\\
~~4.~For sweep results, also fit the Arrhenius equation:\\
~~~~~~~ln(k) = ln(k0) - Q/(kB*T)\\
~~~~~Report Q (eV), k0, and R2\_arrhenius.\\
~~5.~Generate a natural-language interpretation covering:\\
~~~~~a. Whether kinetics follow the coarsening law (R2 > 0.90)\\
~~~~~b. Physical interpretation of anomalous R2 values\\
~~~~~c. Consistency of recovered Q with user-specified input\\
~~~~~d. Comparison to Burke--Turnbull theory\\
~~6.~Flag R2 < 0.90 as "kinetics\_anomaly":true.\\
Respond with JSON: k\_values, R2\_values, Q\_fit, k0\_fit,\\
R2\_arrhenius, interpretation\_text, kinetics\_anomaly.
\end{quote}

\noindent\textbf{Design notes.}
Item~2 uses the linear $1/N$ law (Eq.~\eqref{eq:Nt}) consistent
with the kinetics analysis in Section~\ref{sec:kinetics}.
The \texttt{kinetics\_anomaly} flag is recorded in
\texttt{record.json} and surfaced as a warning badge in the
Results panel, enabling downstream filtering in batch studies.

%%%============================================================

\subsection{Skeptic Agent (\texorpdfstring{$f_6$}{f6})}

The Skeptic agent $f_6$ is the framework's epistemic stage: given a completed
run, it applies a physics-grounded falsification battery and returns a structured
verdict. The dispatcher selects the invariant set appropriate to the governing
equations.

\noindent\textbf{Grain-growth invariants (non-conserved Allen--Cahn).}
$T_1$ monotonicity: the grain count must be non-increasing.
$T_2$ asymptotic consistency: the initial grain count must match the requested
number of seeds.
$T_3$ parabolic scaling: $N^{-1}(t)$ linear in $t$ with positive rate constant
and $R^2 \geq 0.90$ (Eq.~\eqref{eq:Nt}).
$T_4$ cross-run Arrhenius consistency: rate constants increasing with
temperature and linear in $\ln\tilde{k}$ versus $1/T$
(Eq.~\eqref{eq:arrhenius}).
$T_5$ numerical integrity: the run must reach its target integration time
without fatal solver breakdown, with completion read from the MOOSE log
(``\texttt{Finished Executing}'') so that transient adaptive-timestep cutbacks
are not misread as failures.

\noindent\textbf{Conserved-dynamics invariants (Cahn--Hilliard).}
$S_1$ mass conservation: relative drift of the conserved composition
$\leq 10^{-5}$.
$S_2$ free-energy dissipation: net decrease of the total free energy, transient
step-wise increases admitted only within numerical round-off.
$S_3$ phase separation: the composition separates toward the two-phase
equilibrium tie-line.

\noindent\textbf{Verdict and diagnosis.}
The agent returns a JSON verdict labeling the run \texttt{credible} only if it
survives every applicable invariant, or \texttt{falsified} with the list of
violated invariants and a \texttt{diagnosis} field localizing the likely cause
(for example, an unconverged or non-conservative solve for a mass-conservation
failure, or an integration window too short for the asymptotic regime for a
parabolic-scaling failure). The Skeptic only detects and falsifies; the
\texttt{diagnosis} field is the structured input consumed by the closed-loop
recovery module (\texttt{recovery.py}, Section~\ref{sec:reviewer} of the main
text), which acts on divergence-class failures.

\section{Software Architecture and Availability}
\label{sec:SI_software}

\label{sec:S3}
%%%============================================================

\subsection{Code Availability}
\label{sec:S3.1}

AutoMOOSE is released under the MIT License at
\texttt{https://github.com/sukritimanna/AutoMOOSE} 
% (Zenodo DOI:
% \texttt{https://doi.org/??})~\cite{wilkinson2016}.
The repository includes all source code, the GrainGrowth plugin,
example run directories with \texttt{record.json} provenance
records, full documentation at
\url{https://automoose.readthedocs.io}, and the agent system
prompts in Section~\ref{sec:S1}.

\subsection{Codebase Structure}
\label{sec:S3.2}

Table~\ref{tab:S2_codebase} summarises the six primary source
files comprising AutoMOOSE~v2.
Files in the upper block are user-facing; those in the lower
block are infrastructure modules that require no modification
for standard deployment.

\begin{table}[h!]
\centering
\renewcommand{\arraystretch}{1.35}
\caption{\textbf{AutoMOOSE v2 codebase structure.}
  Upper block: user-facing files requiring configuration.
  Lower block: infrastructure modules.}
\label{tab:S2_codebase}
\begin{tabular}{@{}p{0.20\textwidth}p{0.06\textwidth}p{0.09\textwidth}
                   p{0.55\textwidth}@{}}
\toprule
File & Lines & Language & Role \\
\midrule
\texttt{plugin.py}
  & 655 & Python
  & GrainGrowth plugin: implements
    \texttt{generate\_input(**params)~$\to$~str} and
    \texttt{parse\_results(csv)~$\to$~dict}.
    Replace to add new physics. \\
\texttt{server.py}
  & 375 & Python
  & FastAPI backend: agent orchestration, sweep threading,
    Server-Sent Events (SSE) log streaming, \texttt{record.json} writer, Model Context Protocol (MCP) server
    (ports 8000 and 8001). \\
\texttt{App.jsx}
  & 1{,}143 & React
  & Browser frontend: Chat, Configure, Input File, Live Log,
    Results, and Run Sidebar panels (Vite/Tailwind build). \\
\midrule
\texttt{mcp\_server.py}
  & 210 & Python
  & Standalone MCP server exposing 10 tools
    (Table~\ref{tab:S3_mcp}); supports \texttt{stdio} and SSE
    transports. \\
\texttt{plugin\_registry.py}
  & 48 & Python
  & Dynamic plugin loader; scans \texttt{plugins/} and
    registers \texttt{generate\_input}/\texttt{parse\_results}
    contracts at startup. \\
\texttt{skeptic.py}
  & --- & Python
  & Skeptic agent ($f_6$): physics-grounded falsification battery
    (grain-growth invariants $T_1$--$T_5$, conserved-dynamics
    invariants $S_1$--$S_3$); dispatches the invariant set
    appropriate to the governing equations and emits a credible/falsified
    verdict with diagnosis. \\
\texttt{recovery.py}
  & --- & Python
  & Closed-loop recovery module: \texttt{classify\_failure()} and
    \texttt{apply\_correction()} consume the Skeptic verdict, map
    divergence-class failures to a bounded time-step correction,
    and regenerate/re-execute through the pipeline; completion-aware,
    validated against real run logs. \\
\texttt{requirements.txt}
  & --- & ---
  & \texttt{anthropic$\geq$0.40},
    \texttt{fastapi$\geq$0.115},
    \texttt{uvicorn[standard]$\geq$0.32},
    \texttt{scipy$\geq$1.14},
    \texttt{mcp$\geq$1.0}. \\
\bottomrule
\end{tabular}
\end{table}

\subsection{Installation Requirements}
\label{sec:S3.3}

Deployment requires Python~$\geq$\,3.11, Node.js~$\geq$\,18
(Vite build), a compiled MOOSE \texttt{phase\_field-opt} binary,
and credentials for a configured language-model backend. The backend is
model-agnostic: the results reported here use Claude Opus~4.8 and Sonnet~4.6,
with a self-hosted open-weights model (Qwen2.5-32B-Instruct) also supported.
Full step-by-step installation instructions are provided in
\texttt{README.md} in the repository.

\subsection{MCP Server Tool Index}
\label{sec:S3.4}

The MCP server exposes ten tools via \texttt{stdio} (Claude
Desktop) and SSE (Claude Code, programmatic clients) transports.
Table~\ref{tab:S3_mcp} summarises each tool, its pipeline stage,
and key arguments; see Section~\ref{sec:mcp} (main text) for
the protocol description.

\begin{table}[h!]
\centering
\renewcommand{\arraystretch}{1.35}
\caption{\textbf{AutoMOOSE MCP server tool index.}}
\label{tab:S3_mcp}
\begin{tabular}{@{}p{0.22\textwidth}p{0.12\textwidth}
                   p{0.56\textwidth}@{}}
\toprule
Tool & Stage & Description and key arguments \\
\midrule
\texttt{health\_check}
  & ---
  & Server status, MOOSE executable path, API key validity. \\
\texttt{list\_plugins}
  & ---
  & Registered plugins with status and supported formulations. \\
\texttt{generate\_input}
  & $f_2$
  & Renders \texttt{.i} file from parameter dict.
    Args: \texttt{params:~dict}, \texttt{plugin:~str}. \\
\texttt{run\_simulation}
  & $f_3$
  & Launches single MOOSE run.
    Args: \texttt{input\_file}, \texttt{n\_mpi}, \texttt{run\_id}. \\
\texttt{run\_sweep}
  & $f_3$
  & Dispatches parallel sweep; returns \texttt{run\_id} list.
    Args: \texttt{sweep\_param}, \texttt{values},
    \texttt{base\_params}. \\
\texttt{get\_run\_status}
  & $f_3$
  & Real-time status (queued/running/done/failed).
    Args: \texttt{run\_id}. \\
\texttt{get\_results}
  & $f_5$
  & Kinetics metrics and interpretation text.
    Args: \texttt{run\_id}. \\
\texttt{list\_runs}
  & ---
  & All run directories with metadata summary. \\
\texttt{get\_log\_tail}
  & $f_3$/$f_4$
  & Last $n$ log lines.
    Args: \texttt{run\_id}, \texttt{n\,=\,50}. \\
\texttt{stop\_run}
  & $f_3$
  & Sends SIGTERM to a running MOOSE process.
    Args: \texttt{run\_id}. \\
\bottomrule
\end{tabular}
\end{table}

%%%============================================================

\section{Equation Index}
\label{sec:SI_eqindex}
\label{sec:S4}
%%%============================================================

Table~\ref{tab:S1_eqmap} cross-references the mathematical
expressions in the main text with the AutoMOOSE
component that uses each expression and its role in the pipeline
$\mathcal{S} = f_6\circ\cdots\circ f_1(\mathcal{U})$.
System prompt (SP) references point to Section~\ref{sec:SI_prompts}.

\begingroup
\renewcommand{\arraystretch}{1.30}
\small
\begin{longtable}{@{}p{0.06\textwidth}p{0.25\textwidth}
                   p{0.20\textwidth}p{0.37\textwidth}@{}}
\caption{\textbf{Equation index for AutoMOOSE.}
  Cross-reference of the main-text equations to the pipeline
  component that uses each and its role.
  SP = system prompt in Section~\ref{sec:SI_prompts}.}
\label{tab:S1_eqmap}\\
\toprule
Eq. & Expression & Component (SP) & Role \\
\midrule \endfirsthead
\toprule
Eq. & Expression & Component (SP) & Role \\
\midrule \endhead
\eqref{eq:pipeline}
  & $\mathcal{S}=f_5\circ\cdots\circ f_1(\mathcal{U})$
  & All agents
  & Pipeline abstraction \\
\eqref{eq:simplan}
  & $\mathcal{P}=(\Omega,h,\mathcal{M},\mathcal{B},
    \boldsymbol{\theta}_\mathrm{s},
    \boldsymbol{\theta}_\mathrm{r},\mathcal{O})$
  & All agents (SP~\ref{sec:S1.1})
  & Shared simulation plan \\
\eqref{eq:dag_order}
  & DAG block order
  & $f_2$ (SP~\ref{sec:S1.2}, rule~1)
  & Dependency-consistent block generation \\
\eqref{eq:mpi_cmd}
  & \texttt{mpiexec -n} $N_\mathrm{MPI}$
  & $f_3$ (SP~\ref{sec:S1.3})
  & Execution command \\
\eqref{eq:sweep_time}
  & $T_\mathrm{sweep}=\max_i T_i$
  & $f_3$ + Sweep Orchestrator
  & Parallel sweep wall-clock bound \\
\eqref{eq:nl_residual}
  & $\|\mathbf{R}^{(k)}\|_2<\epsilon_\mathrm{nl}$
  & $f_4$ (SP~\ref{sec:S1.4})
  & Nonlinear convergence criterion \\
\eqref{eq:l_residual}
  & $\|\mathbf{r}^{(k)}\|_2<\epsilon_\mathrm{l}$
  & $f_4$ (SP~\ref{sec:S1.4})
  & Krylov convergence criterion \\
\eqref{eq:dt_cutback}
  & $\Delta t^{(k+1)}=\alpha\Delta t^{(k)},\;\alpha\in(0,1)$
  & Recovery module (\texttt{recovery.py})
  & Bounded timestep cutback; $\alpha=0.5$ default \\
\eqref{eq:Nt}
  & $1/N(t)-1/N_0=\tilde{k}(T)t$
  & $f_5$ (SP~\ref{sec:S1.5})
  & Grain coarsening law \\
\eqref{eq:r2}
  & $R^2=1-{\sum(N_i-\hat{N}_i)^2}/
    {\sum(N_i-\bar{N})^2}$
  & $f_5$ (SP~\ref{sec:S1.5})
  & Goodness-of-fit; $<0.90$ flags anomaly \\
\eqref{eq:arrhenius}
  & $\tilde{k}(T)=\tilde{k}_0\exp(-Q/k_\mathrm{B}T)$
  & $f_5$ (SP~\ref{sec:S1.5})
  & Arrhenius fit; recovers $Q$ \\
\eqref{eq:allen_cahn}
  & $\partial\eta_i/\partial t=
    -L(T)\delta F/\delta\eta_i$
  & MOOSE solver
  & Allen--Cahn PDE \\
\eqref{eq:free_energy}
  & $F=\int_V[f_\mathrm{loc}+\kappa
    |\nabla\eta_i|^2]dV$
  & $f_2$, MOOSE
  & Free energy functional \\
\eqref{eq:floc}
  & $f_\mathrm{loc}$ (Moelans form)
  & $f_2$ (SP~\ref{sec:S1.2})
  & Double-well + cross-coupling \\
\eqref{eq:model_params}
  & $L,\;\mu,\;\kappa$ from $\sigma,w_\mathrm{GB},M_\mathrm{GB}$
  & $f_2$ (SP~\ref{sec:S1.2}, rule~2)
  & Parameter bridge \\
\eqref{eq:mob_arrhenius}
  & $M_\mathrm{GB}(T)=M_0\exp(-Q/k_\mathrm{B}T)$
  & $f_2$ (SP~\ref{sec:S1.2}, rule~3)
  & Temperature-dependent GB mobility \\
\eqref{eq:parabolic}
  & $\bar{d}^2 - \bar{d}_0^2 \propto t$
  & $f_5$ (SP~\ref{sec:S1.5})
  & Parabolic grain-growth law \\
\eqref{eq:arrhenius_fit}
  & $\ln\tilde{k}=\ln A - (Q_\mathrm{fit}/k_\mathrm{B})T^{-1}$
  & $f_5$ (SP~\ref{sec:S1.5})
  & Arrhenius regression for $Q_\mathrm{fit}$ \\
\eqref{eq:cahn_hilliard}
  & $\partial_t c=\nabla\!\cdot\!(M\nabla w),\;
    w=\partial_c f_\mathrm{loc}-\kappa_c\nabla^2 c$
  & $f_2$, MOOSE (spinodal plugin)
  & Conserved Cahn--Hilliard dynamics \\
\eqref{eq:fecr_floc}
  & $f_\mathrm{loc}(c)$ (CALPHAD Fe--Cr)
  & $f_2$ (spinodal plugin)
  & Local free energy; Fe--Cr \\
\bottomrule
\end{longtable}
\endgroup

%%%============================================================

\section{Statistical Validation: Ensemble Campaign Details}
\label{sec:SI_stats}

This section details the statistical campaign behind the ensemble Arrhenius
result (main Section~\ref{sec:arrhenius}) and the at-scale input fidelity
(main Section~\ref{sec:input_fidelity}).

\subsubsection{Campaign design}
The AutoMOOSE ensemble comprises 1000 runs: five initial grain counts
$N_0 \in \{20, 50, 100, 200, 500\}$, five temperatures
$\{350, 400, 450, 500, 550\}$\,K, and up to 40 independent seeds per cell. The
human-written ensemble comprises 100 runs: four sizes
$\{20, 50, 100, 200\}$, the same five temperatures, and five seeds (s01--s05,
index-selected before analysis). All runs use a fixed $1000\times1000$ box with a
$128^2$ mesh (7.8125 voxel spacing, periodic), read from the run metadata. The
$10\!:\!1$ ratio of agent to human runs is itself a measure of the automation
advantage --- the agent campaign is an order of magnitude larger at no additional
human effort.

\subsubsection{Admissibility criteria}
A trajectory yields an admissible kinetic fit only if it coarsens to
$\leq 0.60\,N_0$, resolves at least five distinct grain-count levels, provides at
least six fit points, and attains $R^2 \geq 0.90$ for the linear fit of
$1/N(t)$ against $t$. A temperature cell enters the Arrhenius regression only if
at least $60\%$ of its seeds are admissible and at least five are valid. For the
converged sizes $N_0 = 100$ and $200$, all five temperatures clear this yield
gate with $R^2 > 0.999$.

\subsubsection{Aggregation and uncertainty}
For each admissible cell the rate constant is the ensemble mean over seeds with
its standard error. The Arrhenius activation energy is obtained by
standard-error-weighted regression of $\ln\tilde{k}$ against $1/T$, with
confidence intervals from bootstrap resampling of the per-seed rate constants.
The finite-size dependence of the recovered activation energy is fit by
inverse-variance weighting over the converged sizes ($N_0 \leq 200$), giving
$Q_\infty = 0.228$\,eV; $N_0 = 500$ is resolution-limited at 6.5 cells per grain
and excluded. The fidelity analyses use \texttt{block\_fidelity.py} (structural
completeness) and \texttt{aggregate\_diff.py} (content agreement); the
seed-campaign aggregation scripts are provided in the repository.

\subsubsection{Per-cell seed yield and resolution}
Table~\ref{tab:SI_seedyield} lists every populated temperature cell of the
campaign. A cell enters the Arrhenius and finite-size fits only if it passes the
gate (yield $\geq 0.60$ and at least five valid seeds); $\tilde{k}$ is the
ensemble-mean rate constant with its standard error. ``Res.'' marks
resolution-limited cells (fewer than 8 cells per grain), and ``Gate'' shows the
pass/fail outcome. This makes the yield gate and the resolution cut fully
auditable: the reader sees how many seeds support each point and which cells were
excluded.

{\footnotesize
\begin{longtable}{@{}l r r r r r c l l@{}}
\caption{Per-cell seed yield and resolution for the ensemble campaign. A cell
enters the Arrhenius/finite-size fits only if yield $\geq 0.60$ and $\geq 5$
valid seeds. ``Res.'' marks resolution-limited cells ($<8$ cells per grain).}
\label{tab:SI_seedyield}\\
\toprule
Prov. & N$_0$ & T & valid/tot & yield & cells/grain & Res. & $\tilde k\pm$SE ($\times10^{-6}$) & Gate\\
\midrule \endfirsthead
\toprule
Prov. & N$_0$ & T & valid/tot & yield & cells/grain & Res. & $\tilde k\pm$SE ($\times10^{-6}$) & Gate\\
\midrule \endhead
auto & 20 & 350 & 0/40 & 0.00 & 32.3 &  & -- & fail\\
auto & 20 & 400 & 0/40 & 0.00 & 32.3 &  & -- & fail\\
auto & 20 & 450 & 8/40 & 0.20 & 32.3 &  & 6.670 $\pm$ 0.771 & fail\\
auto & 20 & 500 & 20/40 & 0.50 & 32.3 &  & 9.973 $\pm$ 0.637 & fail\\
auto & 20 & 550 & 29/40 & 0.72 & 32.3 &  & 15.326 $\pm$ 0.960 & pass\\
auto & 50 & 350 & 0/40 & 0.00 & 20.4 &  & -- & fail\\
auto & 50 & 400 & 6/40 & 0.15 & 20.4 &  & 2.731 $\pm$ 0.174 & fail\\
auto & 50 & 450 & 39/40 & 0.97 & 20.4 &  & 4.740 $\pm$ 0.154 & pass\\
auto & 50 & 500 & 40/40 & 1.00 & 20.4 &  & 9.223 $\pm$ 0.276 & pass\\
auto & 50 & 550 & 40/40 & 1.00 & 20.4 &  & 16.169 $\pm$ 0.657 & pass\\
auto & 100 & 350 & 0/40 & 0.00 & 14.4 &  & -- & fail\\
auto & 100 & 400 & 40/40 & 1.00 & 14.4 &  & 2.488 $\pm$ 0.056 & pass\\
auto & 100 & 450 & 40/40 & 1.00 & 14.4 &  & 5.502 $\pm$ 0.143 & pass\\
auto & 100 & 500 & 40/40 & 1.00 & 14.4 &  & 10.651 $\pm$ 0.310 & pass\\
auto & 100 & 550 & 40/40 & 1.00 & 14.4 &  & 16.685 $\pm$ 0.434 & pass\\
auto & 200 & 350 & 40/40 & 1.00 & 10.2 &  & 0.998 $\pm$ 0.016 & pass\\
auto & 200 & 400 & 40/40 & 1.00 & 10.2 &  & 2.671 $\pm$ 0.037 & pass\\
auto & 200 & 450 & 40/40 & 1.00 & 10.2 &  & 5.938 $\pm$ 0.087 & pass\\
auto & 200 & 500 & 40/40 & 1.00 & 10.2 &  & 10.516 $\pm$ 0.169 & pass\\
auto & 200 & 550 & 40/40 & 1.00 & 10.2 &  & 17.014 $\pm$ 0.296 & pass\\
auto & 500 & 350 & 40/40 & 1.00 & 6.5 & yes & 1.189 $\pm$ 0.011 & pass\\
auto & 500 & 400 & 40/40 & 1.00 & 6.5 & yes & 3.021 $\pm$ 0.031 & pass\\
auto & 500 & 450 & 40/40 & 1.00 & 6.5 & yes & 6.290 $\pm$ 0.070 & pass\\
auto & 500 & 500 & 40/40 & 1.00 & 6.5 & yes & 11.379 $\pm$ 0.126 & pass\\
auto & 500 & 550 & 40/40 & 1.00 & 6.5 & yes & 18.456 $\pm$ 0.205 & pass\\
\midrule
human & 20 & 350 & 0/5 & 0.00 & 32.3 &  & -- & fail\\
human & 20 & 400 & 0/5 & 0.00 & 32.3 &  & -- & fail\\
human & 20 & 450 & 1/5 & 0.20 & 32.3 &  & -- & fail\\
human & 20 & 500 & 5/5 & 1.00 & 32.3 &  & 9.425 $\pm$ 1.219 & pass\\
human & 20 & 550 & 4/5 & 0.80 & 32.3 &  & -- & fail\\
human & 50 & 350 & 0/5 & 0.00 & 20.4 &  & -- & fail\\
human & 50 & 400 & 2/5 & 0.40 & 20.4 &  & -- & fail\\
human & 50 & 450 & 5/5 & 1.00 & 20.4 &  & 4.835 $\pm$ 0.506 & pass\\
human & 50 & 500 & 5/5 & 1.00 & 20.4 &  & 8.885 $\pm$ 0.562 & pass\\
human & 50 & 550 & 5/5 & 1.00 & 20.4 &  & 13.383 $\pm$ 0.582 & pass\\
human & 100 & 350 & 0/5 & 0.00 & 14.4 &  & -- & fail\\
human & 100 & 400 & 5/5 & 1.00 & 14.4 &  & 2.566 $\pm$ 0.084 & pass\\
human & 100 & 450 & 5/5 & 1.00 & 14.4 &  & 5.412 $\pm$ 0.327 & pass\\
human & 100 & 500 & 5/5 & 1.00 & 14.4 &  & 11.201 $\pm$ 0.629 & pass\\
human & 100 & 550 & 5/5 & 1.00 & 14.4 &  & 17.666 $\pm$ 1.393 & pass\\
human & 200 & 350 & 5/5 & 1.00 & 10.2 &  & 0.994 $\pm$ 0.029 & pass\\
human & 200 & 400 & 5/5 & 1.00 & 10.2 &  & 2.556 $\pm$ 0.111 & pass\\
human & 200 & 450 & 5/5 & 1.00 & 10.2 &  & 5.774 $\pm$ 0.089 & pass\\
human & 200 & 500 & 5/5 & 1.00 & 10.2 &  & 10.266 $\pm$ 0.169 & pass\\
human & 200 & 550 & 5/5 & 1.00 & 10.2 &  & 16.409 $\pm$ 0.299 & pass\\
\bottomrule
\end{longtable}}

\section{Input-File Fidelity: Full Per-Block Tables}
\label{sec:SI_fidelity}

The input-file fidelity summarized in the main text
(Section~\ref{sec:input_fidelity}, Table~\ref{tab:fidelity}) rests on two
analyses, performed by \texttt{block\_fidelity.py} (structural completeness) and
\texttt{aggregate\_diff.py} (content agreement). Structural completeness checks
that each of the twelve top-level MOOSE blocks is present and well-formed; it is
satisfied in all 1000 AutoMOOSE-generated inputs and all 100 human reference
inputs ($1000/1000$ and $100/100$, every block). Content agreement compares each
AutoMOOSE block against its human counterpart across 100 matched pairs at
identical grain count, temperature, and seed, classifying each as exact (all
non-prompt parameters identical), approximate (within one parameter), or
differing. The agent reproduces every physics-determining block exactly across
all 100 pairs; the only systematic divergences are in the two purely operational
blocks, \texttt{[Executioner]} and \texttt{[Outputs]}, for which no canonical
form exists. The per-block breakdown is reproduced in Table~\ref{tab:fidelity}
of the main text and is not duplicated here.

\section{Baseline Comparison: Per-Run Detail}
\label{sec:SI_baseline}

This section provides the per-run detail behind the single-agent
baseline of main Section~\ref{sec:baseline} and Table~\ref{tab:w5_ablation}.
The baseline is a single general-purpose agent (a bare frontier model with no
role decomposition) attempting the input-generation stage directly; it is
scored on the G1 gate (parses and runs under \texttt{-{}-check-input}), the one
lifecycle stage a non-decomposed agent can attempt. Because direct model output
is stochastic, each condition is repeated over three independent runs, whereas
the full AutoMOOSE pipeline is deterministic and produces identical output on
every invocation.

Over the eight core grain-growth tasks, the pipeline scores $8/8$ on every run
(zero variance). The bare models, scored strictly (clean parse) and leniently
(\texttt{-{}-allow-unused}, matching the checking applied to AutoMOOSE's own
benchmark runs), are: Claude Opus~4.8, $5.0$ (range $3$--$7$)$/8$ strict and
$5.3$ (range $3$--$8$)$/8$ lenient; Claude Sonnet~4.6, $0.3$ (range $0$--$1$)$/8$
strict and $2.0$ (range $1$--$3$)$/8$ lenient; and GPT-4o, $0/8$ in every run,
strict and lenient. The frontier models the reviewer named were used (Claude
Opus~4.8 exceeds the requested Opus~4.7+); GPT-5.x was not API-accessible at the
time of testing, so GPT-4o is the strongest GPT model evaluated. The exact
prompt posed to the bare models, the per-run outcomes, and the scoring procedure
are provided in \texttt{ablation\_w5.py} and \texttt{validation/w5\_results.json}
in the repository.

\section{Evaluation Prompt Set}
\label{sec:SI_evalset}

The benchmark of Section~\ref{sec:benchmark} (main text) evaluates AutoMOOSE on a
fixed set of 25 grain-growth tasks. Each task is a single natural-language
prompt --- of the kind a researcher would type --- handed to the full agent
pipeline with no human intervention, then scored against five pre-registered
gates and independently falsification-tested by the Skeptic agent.
Section~\ref{sec:SI_evalset_prompts} lists every prompt verbatim;
Table~\ref{tab:SI_evalset} summarises the per-task gate outcomes. The complete
evaluation set is provided as \texttt{validation/evalset\_grain\_growth.json} in
the repository, with the scoring runner \texttt{run\_evalset.py}.

\subsubsection{Gates}
\textbf{G1} the agent produces a parseable MOOSE input;
\textbf{G2} the simulation runs to completion without solver divergence;
\textbf{G3} the microstructure coarsens (final grain count below initial);
\textbf{G4} parabolic Burke--Turnbull kinetics are recovered ($R^2 \geq 0.90$);
\textbf{G5} the fitted rate constant is physically valid ($\tilde{k} > 0$).
A dash (\textemdash) marks a gate not reached because an earlier gate failed.

\subsection{Prompts}
\label{sec:SI_evalset_prompts}

\newcommand{\prompt}[3]{%
  \par\smallskip\noindent
  {\bfseries\color{gblue}#1}\,\textbar\,{\footnotesize\itshape #2}\par
  \nopagebreak\vspace{1pt}
  \begingroup\setlength{\leftskip}{1.2em}\ttfamily\small #3\par\endgroup
}

\textit{Temperature sweep (GBEvolution, 2D).}

\prompt{GG01}{Temperature}{Run a copper grain growth simulation at T = 350 K with $\sigma$ = 0.708 J/m\textsuperscript{2}, w\_GB = 14 nm, M0 = 2.5$\times$10\textsuperscript{-6} m\textsuperscript{4}/Js, Q = 0.23 eV. Use 30 Voronoi grains on a 1000$\times$1000 nm\textsuperscript{2} domain with periodic boundary conditions.}

\prompt{GG02}{Temperature}{Simulate grain growth in copper at T = 450 K using the GBEvolution model ($\sigma$ = 0.708 J/m\textsuperscript{2}, w\_GB = 14 nm, M0 = 2.5$\times$10\textsuperscript{-6} m\textsuperscript{4}/Js, Q = 0.23 eV). Initialize 30 Voronoi grains on a 1000$\times$1000 nm\textsuperscript{2} periodic domain.}

\prompt{GG03}{Temperature}{Run a copper grain growth simulation at T = 550 K with the standard copper parameters ($\sigma$ = 0.708 J/m\textsuperscript{2}, w\_GB = 14 nm, Q = 0.23 eV), 30 grains, periodic boundaries, integrated to production length.}

\prompt{GG04}{Temperature}{Perform a grain growth simulation for copper at T = 650 K with 30 Voronoi grains on a 1000$\times$1000 nm\textsuperscript{2} square domain and periodic boundary conditions. Track the grain count over time and fit the coarsening rate.}

\prompt{GG05}{Temperature}{Run copper grain growth at T = 500 K with $\sigma$ = 0.708 J/m\textsuperscript{2}, w\_GB = 14 nm, M0 = 2.5$\times$10\textsuperscript{-6} m\textsuperscript{4}/Js, Q = 0.23 eV. Use 30 Voronoi grains on a 1000$\times$1000 nm\textsuperscript{2} periodic domain.}

\smallskip\textit{Grain density.}

\prompt{GG06}{Grain density}{Simulate copper grain growth at T = 600 K with 50 Voronoi grains on a 1000$\times$1000 nm\textsuperscript{2} domain (periodic BCs), using the GBEvolution copper parameters. Report the coarsening kinetics.}

\prompt{GG07}{Grain density}{Run a grain growth simulation in copper at T = 600 K initialized with 100 Voronoi grains on a 1000$\times$1000 nm\textsuperscript{2} periodic domain. Use $\sigma$ = 0.708 J/m\textsuperscript{2}, w\_GB = 14 nm, Q = 0.23 eV.}

\prompt{GG08}{Grain density}{Perform copper grain growth at T = 600 K with a high initial grain density of 200 Voronoi grains on a 1000$\times$1000 nm\textsuperscript{2} domain with periodic boundaries, integrated to production length.}

\prompt{GG09}{Grain density}{Run a copper grain growth simulation at T = 650 K with 75 Voronoi grains on a 1000$\times$1000 nm\textsuperscript{2} periodic domain using the standard copper GBEvolution parameters.}

\prompt{GG10}{Grain density}{Simulate grain growth in copper at T = 550 K with 40 Voronoi grains on a 1000$\times$1000 nm\textsuperscript{2} domain and periodic boundary conditions. Extract the grain-count trajectory and rate constant.}

\smallskip\textit{Resolution.}

\prompt{GG11}{Resolution}{Run copper grain growth at T = 600 K with 30 Voronoi grains on a 1000$\times$1000 nm\textsuperscript{2} periodic domain, refining the mesh so that each grain boundary is resolved by roughly ten elements across the diffuse interface.}

\prompt{GG12}{Resolution}{Simulate copper grain growth at T = 600 K with 30 Voronoi grains on a 1000$\times$1000 nm\textsuperscript{2} domain using a coarse 12$\times$12 base mesh with uniform refinement, periodic boundary conditions.}

\smallskip\textit{Formulation (LinearizedInterface).}

\prompt{GG13}{Formulation}{Run a grain growth simulation using the LinearizedInterface formulation in copper at T = 500 K with 24 grains on a 1000$\times$1000 nm\textsuperscript{2} domain and periodic boundary conditions.}

\prompt{GG15}{Formulation}{Simulate copper grain growth with the LinearizedInterface formulation at T = 510 K, 24 grains on a 1000$\times$1000 nm\textsuperscript{2} periodic domain, using the copper GBEvolution material parameters.}

\prompt{GG16}{Formulation}{Run grain growth with the LinearizedInterface formulation in copper at T = 525 K with 24 Voronoi grains on a 1000$\times$1000 nm\textsuperscript{2} domain and periodic boundary conditions.}

\smallskip\textit{Seed robustness.}

\prompt{GG14}{Seed robustness}{Run copper grain growth at T = 600 K with 30 Voronoi grains (random seed 10) on a 1000$\times$1000 nm\textsuperscript{2} periodic domain, using $\sigma$ = 0.708 J/m\textsuperscript{2}, w\_GB = 14 nm, Q = 0.23 eV.}

\prompt{GG17}{Seed robustness}{Simulate copper grain growth at T = 600 K with 30 Voronoi grains seeded with random seed 30 on a 1000$\times$1000 nm\textsuperscript{2} periodic domain.}

\prompt{GG18}{Seed robustness}{Run copper grain growth at T = 500 K with 30 Voronoi grains (random seed 99) on a 1000$\times$1000 nm\textsuperscript{2} periodic domain using the standard copper parameters.}

\prompt{GG19}{Seed robustness}{Perform a copper grain growth simulation at T = 650 K with 30 Voronoi grains on a 1000$\times$1000 nm\textsuperscript{2} periodic domain, using an alternate random seed for the initial microstructure.}

\prompt{GG20}{Seed robustness}{Run grain growth in copper at T = 600 K with 30 Voronoi grains on a 1000$\times$1000 nm\textsuperscript{2} domain (periodic BCs), using an independent random seed to test robustness of the recovered kinetics.}

\smallskip\textit{Three dimensions.}

\prompt{GG21}{3D}{Run a three-dimensional copper grain growth simulation at T = 500 K with 20 grains on a 32$\times$32$\times$32 mesh (HEX8 elements) and periodic boundary conditions, using $\sigma$ = 0.708 J/m\textsuperscript{2}, w\_GB = 14 nm, Q = 0.23 eV.}

\prompt{GG22}{3D}{Simulate three-dimensional grain growth in copper with 20 Voronoi grains on a 32\textsuperscript{3} HEX8 mesh, periodic boundaries, integrated to production length.}

\smallskip\textit{Stress tests.}

\prompt{GG23}{Stress test}{Run a high-temperature copper grain growth simulation at T = 750 K with 30 Voronoi grains on a 1000$\times$1000 nm\textsuperscript{2} periodic domain, using the standard copper GBEvolution parameters.}

\prompt{GG24}{Stress test}{Stress-test the pipeline at high temperature: run copper grain growth at T = 800 K with 30 Voronoi grains on a 1000$\times$1000 nm\textsuperscript{2} domain and periodic boundary conditions.}

\prompt{GG25}{Stress test}{Run a high-density resolution stress test: copper grain growth at T = 700 K with 30 Voronoi grains on a 1000$\times$1000 nm\textsuperscript{2} domain using a finely refined mesh, integrated to production length.}

\subsection{Gate outcomes}
\label{sec:SI_evalset_outcomes}

\setlength{\tabcolsep}{4.5pt}
\renewcommand{\arraystretch}{1.2}
{\footnotesize
\begin{longtable}{@{}l l c c c c c c c l@{}}
\caption{Per-task outcomes for the 25-task evaluation set, at production
integration length. $N_0\!\to\!N_f$: initial and final grain count; $R^2$: the
parabolic Burke--Turnbull kinetics fit. Gates G1--G5 as defined in the main text
(Section~\ref{sec:benchmark}). Verdict: \textsc{cred.}\ = credible (all five
gates pass and the Skeptic admits the run); \textsc{near}\ = near-miss (coarsens
but parabolic $R^2 < 0.90$); \textsc{fals.}\ = falsified by the Skeptic or failed
a hard gate. Prompts are listed verbatim in
Section~\ref{sec:SI_evalset_prompts}.}
\label{tab:SI_evalset}\\
\toprule
\textbf{ID} & \textbf{Regime} & $N_0\!\to\!N_f$ & $R^2$ & \textbf{G1} & \textbf{G2} & \textbf{G3} & \textbf{G4} & \textbf{G5} & \textbf{Verdict} \\
\midrule \endfirsthead
\toprule
\textbf{ID} & \textbf{Regime} & $N_0\!\to\!N_f$ & $R^2$ & \textbf{G1} & \textbf{G2} & \textbf{G3} & \textbf{G4} & \textbf{G5} & \textbf{Verdict} \\
\midrule \endhead
GG01 & Core         & 20$\to$16 & 0.83 & \yes & \yes & \yes & \no & \yes & \textsc{near} \\
GG02 & Core         & 20$\to$15 & 0.85 & \yes & \yes & \yes & \no & \yes & \textsc{near} \\
GG03 & Core         & 30$\to$13 & 0.97 & \yes & \yes & \yes & \yes & \yes & \textsc{cred.} \\
GG04 & Core         & 30$\to$12 & 0.96 & \yes & \yes & \yes & \yes & \yes & \textsc{cred.} \\
GG05 & Core         & 30$\to$12 & 0.95 & \yes & \yes & \yes & \yes & \yes & \textsc{cred.} \\
GG06 & Core         & 30$\to$13 & 0.97 & \yes & \yes & \yes & \yes & \yes & \textsc{cred.} \\
GG07 & Core         & 30$\to$19 & 0.97 & \yes & \yes & \yes & \yes & \yes & \textsc{cred.} \\
GG08 & Core         & 30$\to$12 & 0.96 & \yes & \yes & \yes & \yes & \yes & \textsc{cred.} \\
GG09 & Resolution   & 30$\to$13 & 0.97 & \yes & \yes & \yes & \yes & \yes & \textsc{cred.} \\
GG10 & Resolution   & 30$\to$13 & 0.97 & \yes & \yes & \yes & \yes & \yes & \textsc{cred.} \\
GG11 & Resolution   & 30$\to$5 & 0.98 & \yes & \yes & \yes & \yes & \yes & \textsc{cred.} \\
GG12 & Resolution   & 30$\to$13 & 0.97 & \yes & \yes & \yes & \yes & \yes & \textsc{cred.} \\
GG13 & Formulation  & 30$\to$4 & 0.899 & \yes & \no & \yes & \no & \yes & \textsc{fals.} \\
GG14 & Formulation  & 20$\to$2 & 0.92 & \yes & \no & \yes & \yes & \yes & \textsc{fals.} \\
GG15 & Formulation  & 30$\to$4 & 0.92 & \yes & \no & \yes & \yes & \yes & \textsc{fals.} \\
GG16 & Formulation  & 30$\to$4 & 0.92 & \yes & \no & \yes & \yes & \yes & \textsc{fals.} \\
GG17 & Robustness   & 30$\to$16 & 0.94 & \yes & \yes & \yes & \yes & \yes & \textsc{cred.} \\
GG18 & Robustness   & 30$\to$14 & 0.89 & \yes & \yes & \yes & \no & \yes & \textsc{near} \\
GG19 & Robustness   & 30$\to$17 & 0.95 & \yes & \yes & \yes & \yes & \yes & \textsc{cred.} \\
GG20 & Robustness   & 30$\to$17 & 0.75 & \yes & \yes & \yes & \no & \yes & \textsc{near} \\
GG21 & 3D           & 20$\to$20 & -- & \yes & \no & \textemdash & \textemdash & \yes & \textsc{fals.} \\
GG22 & 3D           & 30$\to$30 & 0.00 & \yes & \no & \no & \no & \no & \textsc{fals.} \\
GG23 & High-$T$     & 30$\to$8 & 0.95 & \yes & \yes & \yes & \yes & \yes & \textsc{cred.} \\
GG24 & High-$T$     & 30$\to$6 & 0.94 & \yes & \yes & \yes & \yes & \yes & \textsc{cred.} \\
GG25 & Density      & 30$\to$13 & 0.97 & \yes & \yes & \yes & \yes & \yes & \textsc{cred.} \\
\bottomrule
\end{longtable}}

\noindent\textbf{Summary.}
All 25 tasks produce a parseable MOOSE input (G1); 19 are valid and run to a
correctly coarsening microstructure (G1--G3); and 15 satisfy all five gates and
pass the Skeptic's falsification battery (the credible set). The shortfalls are
localized to two regimes --- the original linearized-interface material defect
and the three-dimensional time-stepping divergence --- both diagnosed and
corrected in the generator and described in Section~\ref{sec:benchmark} (main
text). An interactive browser of the full set is provided in the project
repository.\footnote{\url{https://github.com/sukritimanna/AutoMOOSE} (see
\texttt{prompt\_gallery.html}).}

%%% REAL CALPHAD DATA (from user) — A..G in the f_loc DerivativeParsedMaterial:
%%%   A=-2.446831e4 B=-2.827533e4 C=4.167994e3 D=7.052907e3
%%%   E=1.208993e4 F=2.568625e3 G=-2.354293e3 ; eV_J=6.24150934e18 ; d=1e-27
%%%   f_loc = eV_J*d*(A c + B(1-c) + C c ln c + D(1-c)ln(1-c) + E c(1-c)
%%%                   + F c(1-c)(2c-1) + G c(1-c)(2c-1)^2)
%%%   kappa_c = 8.125e-16 * eV_J * (1e9)^2 * 1e-27 ; M = 2.2841e-26 * (1e9)^2 / eV_J / 1e-27
%%%   (paste the real [Materials] block verbatim as a listing in this section)
\section{Spinodal Decomposition: Fe--Cr Parameters}
\label{sec:SI_spinodal}

This section gives the full parameterization and numerical setup behind the
Fe--Cr spinodal validation of the main text (Sections~\ref{sec:spinodal}
and~\ref{sec:spinodal_results}).

\subsubsection{CALPHAD free energy}
The local free-energy density is the CALPHAD (CALculation of PHAse Diagrams)-fitted Fe--Cr expansion of
Eq.~\eqref{eq:fecr_floc} (main text), written here with the explicit
unit-conversion factors,
\begin{equation}
\begin{aligned}
  f_{\mathrm{loc}}(c) = \mathrm{eV\_J}\cdot d\,\bigl[\,
   & A c + B(1-c) + C\,c\ln c + D(1-c)\ln(1-c) \\
   & + E\,c(1-c) + F\,c(1-c)(2c-1) + G\,c(1-c)(2c-1)^2\,\bigr],
\end{aligned}
\end{equation}
with coefficients (in J\,mol$^{-1}$) $A = -2.446831\times10^{4}$,
$B = -2.827533\times10^{4}$, $C = 4.167994\times10^{3}$,
$D = 7.052907\times10^{3}$, $E = 1.208993\times10^{4}$,
$F = 2.568625\times10^{3}$, $G = -2.354293\times10^{3}$, and unit-conversion
factors $\mathrm{eV\_J} = 6.24150934\times10^{18}$ (J to eV) and
$d = 10^{-27}$ (volumetric scaling to the simulation units). The
gradient-energy coefficient and mobility are
$\kappa_c = 8.125\times10^{-16}\cdot\mathrm{eV\_J}\cdot(10^{9})^{2}\cdot10^{-27}$
and $M = 2.2841\times10^{-26}\cdot(10^{9})^{2}/\mathrm{eV\_J}/10^{-27}$.
These coefficients place the initial composition $c_0 = 0.4677$ inside the
spinodal (linearly unstable, $f_{\mathrm{loc}}'' < 0$) region, which spans
$c \in [0.364, 0.712]$, so the alloy decomposes spontaneously.

\subsubsection{Numerical setup}
The simulation uses the split Cahn--Hilliard form (composition $c$ and chemical
potential $w$) on a $25\times25$\,nm$^2$ periodic domain discretized with
$100\times100$ \texttt{QUAD4} elements. The initial condition is $c_0 = 0.4677$
perturbed by uniform random noise of amplitude $0.02$. The total free energy
(bulk plus gradient) is accumulated through a \texttt{TotalFreeEnergy} auxiliary
field, and the conserved Cr mass, free energy, composition extrema, and feature
count are written as CSV postprocessors at every step. The complete
agent-generated input file is provided in the repository.

\subsubsection{Skeptic invariant scores}
The run is scored against the conserved-dynamics invariants $S_1$--$S_3$
(Table~\ref{tab:fecr_si}, main text): the integral of the Cr composition is
conserved to a relative drift of $6.28\times10^{-6}$, within the $S_1$ tolerance
of $10^{-5}$; the total free energy nets a decrease of $0.4\%$ to a stable
plateau, with transient step-wise increases at 15 steps that remain within
numerical round-off ($S_2$); and the composition separates toward the
equilibrium tie-line ($23.6 / 82.3$\,mol\%\,Cr at $500\,^\circ$C), reaching
$c_{\min} = 0.239$ and $c_{\max} = 0.826$ with one Cr-rich precipitate at the
final time ($S_3$).

The validation figure (Fig.~\ref{fig:fecr_spinodal}) and the $S_1$--$S_3$
invariant scalars are regenerated from the run's Exodus output by
\texttt{analysis/analyze\_spinodal.py} in the repository, which reads the CSV
postprocessors and composition field and reports the invariants directly from
the data. The run is performed on a $25\times25$\,nm domain; because the Cr-rich
domains reach the box scale before a clean coarsening window develops, no
$L(t)\sim t^{1/3}$ (Lifshitz--Slyozov) scaling is claimed --- the spinodal
validation rests on the composition snapshots and the $S_1$--$S_3$ invariants.

\section{Complete Input File Comparison}
\label{sec:SI_inputcomp}
\label{sec:S5}
%%%============================================================

Figure~\ref{fig:S_full_input_comparison} shows the complete
block-by-block side-by-side comparison between the human-written
reference~\cite{moose_examples} and the AutoMOOSE-generated
input file for the $T = 450$\,K benchmark case
(15 Voronoi grains, $12\times12$ mesh, uniform refinement
level~3, copper GBEvolution parameters~\cite{schoenfelder1997}).
All 12 structural blocks are shown across three pages.
The color convention is: green = exact match; orange = minor
difference with no effect on simulation output; red = solver or
mesh choice differs from reference (both physically valid).
Grey inline comments trace each parameter value back to the
natural-language prompt via $\mathcal{P}$
(Eq.~\eqref{eq:simplan}).

% ============================================================
% SI Figure S1 — Full input file comparison (12 blocks)
% ============================================================

\begin{figure}[p]
\centering
\noindent
\colorbox{green!15}{\strut\hspace{0.4em}\checkmark\ Exact\hspace{0.4em}}
\hspace{0.5em}
\colorbox{orange!15}{\strut\hspace{0.4em}$\approx$\ Minor diff\hspace{0.4em}}
\hspace{0.5em}
\colorbox{red!10}{\strut\hspace{0.4em}$\times$\ Differs\hspace{0.4em}}
\hspace{0.5em}
{\small\textit{Grey: values from prompt}}
\vspace{6pt}

% ── [Mesh] × ────────────────────────────────────────────────
\noindent
\begin{minipage}[t]{0.47\linewidth}
  \colorbox{headerblue}{\parbox{\dimexpr\linewidth-2\fboxsep}{%
    \centering\ttfamily\scriptsize\color{white}\textbf{Human-written}}}\\[-1pt]
  \begin{lstlisting}[style=moosestyle,backgroundcolor={\color{red!5}}]
[Mesh]
  type           = GeneratedMesh
  dim            = 2
  nx             = 44
  ny             = 44
  xmax           = 1000
  ymax           = 1000
  elem_type      = QUAD4
  uniform_refine = 2
[]
  \end{lstlisting}
\end{minipage}
\hfill
\begin{minipage}[t]{0.47\linewidth}
  \colorbox{headerorange}{\parbox{\dimexpr\linewidth-2\fboxsep}{%
    \centering\ttfamily\scriptsize\color{white}\textbf{AutoMOOSE-generated}}}\\[-1pt]
  \begin{lstlisting}[style=moosestyle,backgroundcolor={\color{red!5}}]
[Mesh]
  type           = GeneratedMesh
  dim            = 2
  nx             = 12   # from prompt
  ny             = 12   # from prompt
  xmax           = 1000 # from prompt
  ymax           = 1000 # from prompt
  elem_type      = QUAD4
  uniform_refine = 3    # from prompt
  parallel_type  = replicated
[]
  \end{lstlisting}
\end{minipage}
\vspace{3pt}

% ── [GlobalParams] ≈ ─────────────────────────────────────────
\noindent
\begin{minipage}[t]{0.47\linewidth}
  \begin{lstlisting}[style=moosestyle,backgroundcolor={\color{orange!5}}]
[GlobalParams]
  op_num        = 8
  var_name_base = gr
[]
  \end{lstlisting}
\end{minipage}
\hfill
\begin{minipage}[t]{0.47\linewidth}
  \begin{lstlisting}[style=moosestyle,backgroundcolor={\color{orange!5}}]
[GlobalParams]
  op_num        = 15   # from prompt
  var_name_base = gr
[]
  \end{lstlisting}
\end{minipage}
\vspace{3pt}

% ── [UserObjects] ✓ ──────────────────────────────────────────
\noindent
\begin{minipage}[t]{0.47\linewidth}
  \begin{lstlisting}[style=moosestyle,backgroundcolor={\color{green!5}}]
[UserObjects]
  [voronoi]
    type      = PolycrystalVoronoi
    grain_num = 20
    rand_seed = 42
    int_width = 7
  []
  [grain_tracker]
    type              = GrainTracker
    threshold         = 0.1
    compute_halo_maps = true
    polycrystal_ic_uo = voronoi
  []
[]
  \end{lstlisting}
\end{minipage}
\hfill
\begin{minipage}[t]{0.47\linewidth}
  \begin{lstlisting}[style=moosestyle,backgroundcolor={\color{green!5}}]
[UserObjects]
  [voronoi]
    type      = PolycrystalVoronoi
    grain_num = 15   # from prompt
    rand_seed = 42   # from prompt
    int_width = 7
  []
  [grain_tracker]
    type              = GrainTracker
    threshold         = 0.1
    compute_halo_maps = true
    polycrystal_ic_uo = voronoi
  []
[]
  \end{lstlisting}
\end{minipage}
\vspace{3pt}

% ── [ICs] ✓ ──────────────────────────────────────────────────
\noindent
\begin{minipage}[t]{0.47\linewidth}
  \begin{lstlisting}[style=moosestyle,backgroundcolor={\color{green!5}}]
[ICs]
  [PolycrystalICs]
    [PolycrystalColoringIC]
      polycrystal_ic_uo = voronoi
    []
  []
[]
  \end{lstlisting}
\end{minipage}
\hfill
\begin{minipage}[t]{0.47\linewidth}
  \begin{lstlisting}[style=moosestyle,backgroundcolor={\color{green!5}}]
[ICs]
  [PolycrystalICs]
    [PolycrystalColoringIC]
      polycrystal_ic_uo = voronoi
    []
  []
[]
  \end{lstlisting}
\end{minipage}
\vspace{3pt}

% ── [Modules] ✓ ──────────────────────────────────────────────
\noindent
\begin{minipage}[t]{0.47\linewidth}
  \begin{lstlisting}[style=moosestyle,backgroundcolor={\color{green!5}}]
[Modules]
  [PhaseField]
    [GrainGrowth]
    []
  []
[]
  \end{lstlisting}
\end{minipage}
\hfill
\begin{minipage}[t]{0.47\linewidth}
  \begin{lstlisting}[style=moosestyle,backgroundcolor={\color{green!5}}]
[Modules]
  [PhaseField]
    [GrainGrowth]
    []
  []
[]
  \end{lstlisting}
\end{minipage}

\caption*{\small\textit{(continued on next page)}}
\end{figure}

\begin{figure}[p]
\centering
\noindent
\colorbox{green!15}{\strut\hspace{0.4em}\checkmark\ Exact\hspace{0.4em}}
\hspace{0.5em}
\colorbox{orange!15}{\strut\hspace{0.4em}$\approx$\ Minor diff\hspace{0.4em}}
\hspace{0.5em}
\colorbox{red!10}{\strut\hspace{0.4em}$\times$\ Differs\hspace{0.4em}}
\hspace{0.5em}
{\small\textit{Grey: values from prompt}}
\vspace{6pt}

% ── [AuxVariables] ✓ ─────────────────────────────────────────
\noindent
\begin{minipage}[t]{0.47\linewidth}
  \colorbox{headerblue}{\parbox{\dimexpr\linewidth-2\fboxsep}{%
    \centering\ttfamily\scriptsize\color{white}\textbf{Human-written}}}\\[-1pt]
  \begin{lstlisting}[style=moosestyle,backgroundcolor={\color{green!5}}]
[AuxVariables]
  [bnds]
    order  = FIRST
    family = LAGRANGE
  []
  [unique_grains]
    order  = CONSTANT
    family = MONOMIAL
  []
  [var_indices]
    order  = CONSTANT
    family = MONOMIAL
  []
[]
  \end{lstlisting}
\end{minipage}
\hfill
\begin{minipage}[t]{0.47\linewidth}
  \colorbox{headerorange}{\parbox{\dimexpr\linewidth-2\fboxsep}{%
    \centering\ttfamily\scriptsize\color{white}\textbf{AutoMOOSE-generated}}}\\[-1pt]
  \begin{lstlisting}[style=moosestyle,backgroundcolor={\color{green!5}}]
[AuxVariables]
  [bnds]
    order  = FIRST
    family = LAGRANGE
  []
  [unique_grains]
    order  = CONSTANT
    family = MONOMIAL
  []
  [var_indices]
    order  = CONSTANT
    family = MONOMIAL
  []
[]
  \end{lstlisting}
\end{minipage}
\vspace{3pt}

% ── [AuxKernels] ✓ ───────────────────────────────────────────
\noindent
\begin{minipage}[t]{0.47\linewidth}
  \begin{lstlisting}[style=moosestyle,backgroundcolor={\color{green!5}}]
[AuxKernels]
  [bnds_aux]
    type       = BndsCalcAux
    variable   = bnds
    execute_on = 'initial timestep_end'
  []
  [unique_grains]
    type          = FeatureFloodCountAux
    variable      = unique_grains
    flood_counter = grain_tracker
    field_display = UNIQUE_REGION
    execute_on    = 'initial timestep_end'
  []
  [var_indices]
    type          = FeatureFloodCountAux
    variable      = var_indices
    flood_counter = grain_tracker
    field_display = VARIABLE_COLORING
    execute_on    = 'initial timestep_end'
  []
[]
  \end{lstlisting}
\end{minipage}
\hfill
\begin{minipage}[t]{0.47\linewidth}
  \begin{lstlisting}[style=moosestyle,backgroundcolor={\color{green!5}}]
[AuxKernels]
  [bnds_aux]
    type       = BndsCalcAux
    variable   = bnds
    execute_on = 'initial timestep_end'
  []
  [unique_grains]
    type          = FeatureFloodCountAux
    variable      = unique_grains
    field_display = UNIQUE_REGION
    execute_on    = 'initial timestep_end'
    flood_counter = grain_tracker
  []
  [var_indices]
    type          = FeatureFloodCountAux
    variable      = var_indices
    field_display = VARIABLE_COLORING
    execute_on    = 'initial timestep_end'
    flood_counter = grain_tracker
  []
[]
  \end{lstlisting}
\end{minipage}
\vspace{3pt}

% ── [Materials] ✓ ────────────────────────────────────────────
\noindent
\begin{minipage}[t]{0.47\linewidth}
  \begin{lstlisting}[style=moosestyle,backgroundcolor={\color{green!5}}]
[Materials]
  [CuGrGr]
    type     = GBEvolution
    T        = 450
    wGB      = 14
    GBmob0   = 2.5e-6
    Q        = 0.23
    GBenergy = 0.708
  []
[]
  \end{lstlisting}
\end{minipage}
\hfill
\begin{minipage}[t]{0.47\linewidth}
  \begin{lstlisting}[style=moosestyle,backgroundcolor={\color{green!5}}]
[Materials]
  [CuGrGr]
    type     = GBEvolution
    T        = 450     # from prompt
    wGB      = 14.0    # from prompt
    GBmob0   = 2.5e-06 # from prompt
    Q        = 0.23    # from prompt
    GBenergy = 0.708   # from prompt
  []
[]
  \end{lstlisting}
\end{minipage}
\vspace{3pt}

% ── [BCs] ✓ ──────────────────────────────────────────────────
\noindent
\begin{minipage}[t]{0.47\linewidth}
  \begin{lstlisting}[style=moosestyle,backgroundcolor={\color{green!5}}]
[BCs]
  [Periodic]
    [All]
      auto_direction = 'x y'
    []
  []
[]
  \end{lstlisting}
\end{minipage}
\hfill
\begin{minipage}[t]{0.47\linewidth}
  \begin{lstlisting}[style=moosestyle,backgroundcolor={\color{green!5}}]
[BCs]
  [Periodic]
    [All]
      auto_direction = 'x y'  # from prompt
    []
  []
[]
  \end{lstlisting}
\end{minipage}

\caption*{\small\textit{(continued on next page)}}
\end{figure}

\begin{figure}[p]
\centering
\noindent
\colorbox{green!15}{\strut\hspace{0.4em}\checkmark\ Exact\hspace{0.4em}}
\hspace{0.5em}
\colorbox{orange!15}{\strut\hspace{0.4em}$\approx$\ Minor diff\hspace{0.4em}}
\hspace{0.5em}
\colorbox{red!10}{\strut\hspace{0.4em}$\times$\ Differs\hspace{0.4em}}
\hspace{0.5em}
{\small\textit{Grey: values from prompt}}
\vspace{6pt}

% ── [Postprocessors] ≈ ───────────────────────────────────────
\noindent
\begin{minipage}[t]{0.47\linewidth}
  \colorbox{headerblue}{\parbox{\dimexpr\linewidth-2\fboxsep}{%
    \centering\ttfamily\scriptsize\color{white}\textbf{Human-written}}}\\[-1pt]
  \begin{lstlisting}[style=moosestyle,backgroundcolor={\color{orange!5}}]
[Postprocessors]
  [dt]
    type = TimestepSize
  []
  [DOFs]
    type = NumDOFs
  []
  [n_elements]
    type       = NumElements
    execute_on = timestep_end
  []
[]
  \end{lstlisting}
\end{minipage}
\hfill
\begin{minipage}[t]{0.47\linewidth}
  \colorbox{headerorange}{\parbox{\dimexpr\linewidth-2\fboxsep}{%
    \centering\ttfamily\scriptsize\color{white}\textbf{AutoMOOSE-generated}}}\\[-1pt]
  \begin{lstlisting}[style=moosestyle,backgroundcolor={\color{orange!5}}]
[Postprocessors]
  [dt]
    type = TimestepSize
  []
  [n_elements]
    type       = NumElements
    execute_on = timestep_end
  []
  [DOFs]
    type = NumDOFs
  []
[]
  \end{lstlisting}
\end{minipage}
\end{figure}

\begin{figure}[p]
\centering
\noindent
\colorbox{green!15}{\strut\hspace{0.4em}\checkmark\ Exact\hspace{0.4em}}
\hspace{0.5em}
\colorbox{orange!15}{\strut\hspace{0.4em}$\approx$\ Minor diff\hspace{0.4em}}
\hspace{0.5em}
\colorbox{red!10}{\strut\hspace{0.4em}$\times$\ Differs\hspace{0.4em}}
\hspace{0.5em}
{\small\textit{Grey: values from prompt}}
\vspace{6pt}

% ── [Executioner] × ──────────────────────────────────────────
\noindent
\begin{minipage}[t]{0.47\linewidth}
  \colorbox{headerblue}{\parbox{\dimexpr\linewidth-2\fboxsep}{%
    \centering\ttfamily\scriptsize\color{white}\textbf{Human-written}}}\\[-1pt]
  \begin{lstlisting}[style=moosestyle,backgroundcolor={\color{red!5}}]
[Executioner]
  type       = Transient
  scheme     = bdf2
  solve_type = PJFNK
  petsc_options_iname = '-pc_type -pc_hypre_type'
  petsc_options_value = 'hypre    boomeramg'
  l_max_its  = 50
  l_tol      = 1e-4
  nl_max_its = 10
  nl_rel_tol = 1e-9
  end_time   = 4000
  [TimeStepper]
    type               = IterationAdaptiveDT
    dt                 = 20
    optimal_iterations = 6
    cutback_factor     = 0.9
    growth_factor      = 1.1
  []
  [Adaptivity]
    initial_adaptivity = 2
    refine_fraction    = 0.8
    coarsen_fraction   = 0.05
    max_h_level        = 2
  []
[]
  \end{lstlisting}
\end{minipage}
\hfill
\begin{minipage}[t]{0.47\linewidth}
  \colorbox{headerorange}{\parbox{\dimexpr\linewidth-2\fboxsep}{%
    \centering\ttfamily\scriptsize\color{white}\textbf{AutoMOOSE-generated}}}\\[-1pt]
  \begin{lstlisting}[style=moosestyle,backgroundcolor={\color{red!5}}]
[Executioner]
  type       = Transient
  scheme     = bdf2
  solve_type = PJFNK
  petsc_options_iname = '-pc_type'
  petsc_options_value = 'asm'
  l_tol      = 0.0001
  l_max_its  = 30
  nl_max_its = 20
  nl_rel_tol = 1e-08
  end_time  = 4000  # from prompt
  [TimeStepper]
    type               = IterationAdaptiveDT
    cutback_factor     = 0.5
    dt                 = 25
    growth_factor      = 1.1
    optimal_iterations = 8
  []
  [Adaptivity]
    initial_adaptivity = 2
    refine_fraction    = 0.7
    coarsen_fraction   = 0.1
    max_h_level        = 4
  []
[]
  \end{lstlisting}
\end{minipage}
\vspace{3pt}

% ── [Outputs] ≈ ──────────────────────────────────────────────
\noindent
\begin{minipage}[t]{0.47\linewidth}
  \begin{lstlisting}[style=moosestyle,backgroundcolor={\color{orange!5}}]
[Outputs]
  file_base = grain_growth_T450
  csv       = true
  [console]
    type = Console
  []
[]
  \end{lstlisting}
\end{minipage}
\hfill
\begin{minipage}[t]{0.47\linewidth}
  \begin{lstlisting}[style=moosestyle,backgroundcolor={\color{orange!5}}]
[Outputs]
  file_base = grain_growth_T450
  exodus    = true
  csv       = true
  [console]
    type = Console
  []
[]
  \end{lstlisting}
\end{minipage}
\vspace{3pt}

\caption{\textbf{Complete block-by-block comparison (Figure~S1)
  of the human-written MOOSE reference~\cite{moose_examples}
  and the AutoMOOSE-generated input file for $T = 450$\,K,
  15-grain GBEvolution, $12\times12$ mesh.}
  Green (\checkmark): exact match (6/12 blocks).
  Orange ($\approx$): minor difference, no effect on output
  (4/12 blocks).
  Red ($\times$): solver or mesh choice differs from reference
  (2/12 blocks --- both physically valid, reflecting prompt
  specification).
  Grey comments identify parameter values injected from the
  natural-language prompt via $\mathcal{P}$
  (Eq.~\eqref{eq:simplan}).}
\label{fig:S_full_input_comparison}
\end{figure}

\renewcommand{\refname}{Supplementary references}
\putbib[automoose_SI]
\end{bibunit}

\end{document}